\documentclass[letterpaper, 10 pt, conference]{ieeeconf}  

\IEEEoverridecommandlockouts                              

\usepackage{color}
\usepackage{mathtools}
\usepackage{fancyvrb}
\usepackage{hyperref}

\usepackage{booktabs}
\usepackage{bigstrut}
\usepackage{makecell}
\usepackage[table]{xcolor}
\usepackage{rotating}
\usepackage{bbding}
\usepackage[font=small]{caption}

\usepackage{soul}
\usepackage{cite}
\usepackage{dblfloatfix}
\usepackage{placeins}

\usepackage{times}
\usepackage{epsfig}
\usepackage{amsmath}
\usepackage{amssymb}
\usepackage{multirow}
\usepackage{graphics} 
\usepackage{mathptmx} 
\usepackage{subcaption}
\usepackage{url}

\usepackage{algorithm}
\usepackage{algpseudocode}

\title{Sketch2Scene: Automatic Generation of Interactive 3D Game Scenes from User's Casual Sketches
}

\author{Yongzhi Xu$^{1*}$, Yonhon Ng$^{1*}$, Yifu Wang$^{1*}$, Inkyu Sa, Yunfei Duan$^{1}$, \\Zhenhong Sun$^{1}$, Yang Li$^{1}$, Pan Ji$^{1}$, Hongdong Li$^{2}$
\thanks{$^{*}$ Equal contribution.}
\thanks{$^{1}$ XR Vision Labs, Tencent.}%
\thanks{$^{2}$ Australian National University}
\thanks{$\dagger$ Enhanced 2D conceptual image generation method is detailed in \cite{sun2026tss}.}
}

\let\oldtwocolumn\twocolumn
\renewcommand\twocolumn[1][]{
     \oldtwocolumn[{#1}{
     \begin{center}
     \centering
     \includegraphics[width=1\textwidth]{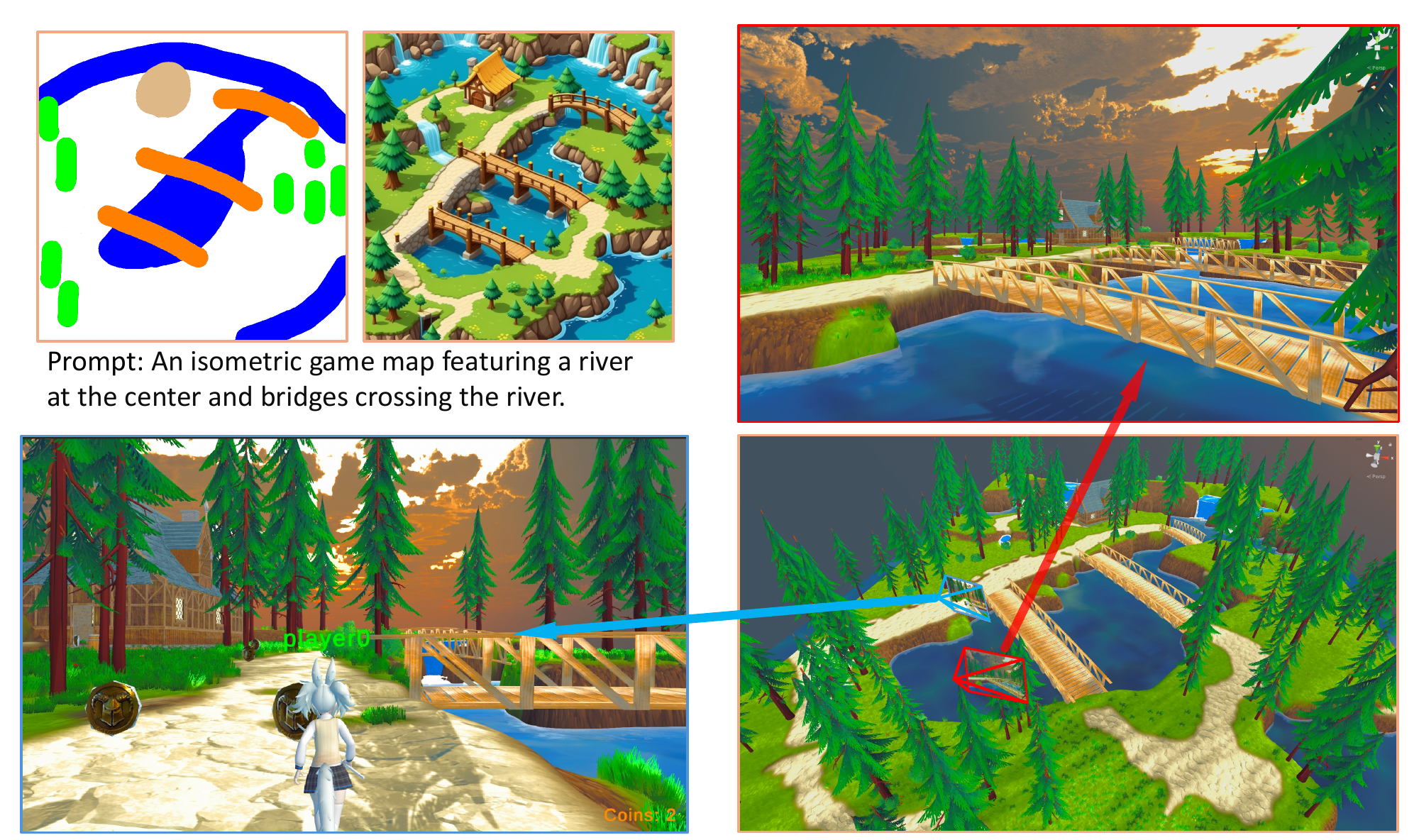}
     \captionof{figure}{We propose an efficient pipeline for automatically generating interactive 3D scenes from user's natural input prompt (e.g., hand-drawn sketch and text description of the scene). 
        An interactive and playable 3D game scene can be created instantly with a hand-drawn sketch of the scene.}
    \end{center}
    \label{fig:teaser}
     }]
}

\begin{document}

\maketitle
\thispagestyle{empty}
\pagestyle{empty}


\begin{abstract}
3D Content Generation is at the heart of many computer graphics applications, including video gaming, film-making, virtual and augmented reality, etc.  This paper proposes a novel deep-learning based approach for automatically generating interactive and playable 3D game scenes, all from the user's casual prompts such as a hand-drawn sketch. Sketch-based input offers a natural, and convenient way to convey the user's design intention in the content creation process.  To circumvent the data-deficient challenge in learning (i.e. the lack of large training data of 3D scenes), our method leverages a pre-trained 2D denoising diffusion model to generate a 2D image of the scene as the conceptual guidance.  In this process, we adopt the isometric projection mode to factor out unknown camera poses while obtaining the scene layout.  From the generated isometric image, we use a pre-trained image understanding method to segment the image into meaningful parts, such as off-ground objects, trees, and buildings, and extract the 2D scene layout. These segments and layouts are subsequently fed into a procedural content generation (PCG) engine, such as a 3D video game engine like Unity or Unreal, to create the 3D scene. The resulting 3D scene can be seamlessly integrated into a game development environment and is readily playable.  Extensive tests demonstrate that our method can efficiently generate high-quality and interactive 3D game scenes with layouts that closely follow the user's intention. 

\end{abstract} 

\maketitle
\section*{Multimedia Material}
\noindent Project Page: {\scriptsize\url{https://xrvisionlabs.github.io/Sketch2Scene/}}\\
\noindent Code$\dagger$: 
{\scriptsize\url{https://github.com/Tencent/Triplet_Tuning}}\\

\section{Introduction}
Generative AI models are taking the world with storm, by enabling the automatic creation of new contents of versatile modalities (e.g. text, image, video, audio and music, etc.), simply from user's natural prompt input. 
AI-generated images, music, and videos can reach a level of quality close to those created by professional artists. This success has already ventured into the realm of 3D object-level asset modeling (such as LRM~\cite{hong2023lrm}, CRM~\cite{wang2024crm}, and MeshLRM~\cite{wei2024meshlrm}), thanks to the growing size of massive 3D object datasets, such as Objaverse-XL~\cite{deitke2024objaverse}. Existing methods published so far have been focusing on the AI generation of small 3D assets of single object level, e.g. ~\cite{hong2023lrm,wang2024crm,wei2024meshlrm}.  

In contrast, the generation of high-quality 3D scenes, such as an open-world game scene, largely remains an under-explored problem. The main reason for this stems from the data efficiency issue for deep learning, namely, due to the lack of a large amount of high-quality 3D scenes to permit large scale training of powerful machine learning models.   For example, so far, there is virtually no publicly available large-scale game scene dataset, other than some city-scale urban driving/street-view scenes captured mainly for autonomous driving research.  

In this paper, we introduce Sketch2Scene, a novel pipeline for 3D scene generation. This method automatically creates realistic and interactive virtual environments using a user-controlled diffusion model, with input provided by a user-drawn sketch and optionally a text prompt. By leveraging casual user sketches, our approach effectively addresses the above-mentioned limitations in generating large-scale, open-world outdoor scenes. To overcome the lack of 3D scene training data, we design a method that leverages a pre-trained 2D denoising diffusion model (e.g.~\cite{stable_diffusion}) for 2D isometric image generation. 

Our method first generates an illustrative 2D image (in isometric projection) depicting the intended concept of the 3D game scene. 
Then, a visual scene understanding module is designed to interpret the image, forming a background terrain (basemap) and foreground object layout map.
This layout map, used as a blueprint, is fed into a procedural content generation pipeline to create 3D game scenes that are compatible hence readily playable in an existing game or rendering engine, such as Unity or Blender. 

To ensure precise and adaptable sketch control, we train the ControlNet~\cite{contolnet} using a semantic-constraint diffusion loss. 
Furthermore, we employ a newly developed basemap inpainting model to generate the scene's basemap. 
To facilitate this process, we have curated a unique gaming isometric dataset for training both the ControlNet and the basemap inpainting networks. For achieving game-ready quality, we use high-resolution texture tiles composed with generated splat maps from the reference Bird's Eye View (BEV) image. Our method significantly surpasses existing scene creation techniques in terms of shape quality, diversity, and controllability.

Our key contributions can be summarised as:  
\begin{itemize}
    \item a controllable, sketch-guided 2D isometric image generation pipeline.
    \item a basemap inpainting model, trained via step-unrolled denoising diffusion on a new dataset.
    \item a learning-based compositional 3D scene understanding module.
    \item a procedural generation pipeline to render an interactive 3D scene using the scene parameters obtained from the above scene understanding module.
\end{itemize}
\section{Related Works} 
\subsection{Diffusion-based 3D scene generation}
The success of diffusion models like Stable Diffusion~\cite{stable_diffusion}, DALLE~\cite{ramesh2021zero}, and Midjourney has significantly boosted interest in developing 3D content generation tools. However, generating high-fidelity 3D scenes from text prompts or images remains challenging due to the complexity and variability in shapes and appearances. Text2Room~\cite{hollein2023text2room} uses 2D text-to-image models and monocular depth estimation for iterative scene generation. Similar indoor-focused approaches include SceneScape~\cite{SceneScape}, which renders videos of diverse scenes, and RealmDreamer~\cite{shriram2024realmdreamer}, which uses a 3D Gaussian Splatting model~\cite{kerbl20233d} for wide-baseline rendering. CC3D~\cite{po2023compositional} generates compositional scenes by optimizing multiple NeRFs with SDS loss~\cite{poole2022dreamfusion}. Unlike CC3D, \cite{epstein2024disentangled} jointly optimizes relative transformations between NeRFs during the SDS process for unsupervised scene decomposition. ControlRoom3D~\cite{schult23controlroom3d} and CTRL-ROOM~\cite{fang2023ctrl} create panorama-view-based text-to-3D room generation models, using 3D room layouts and a fine-tuned ControlNet~\cite{zhang2023controlnet} to edit generated rooms. SceneWiz3D synthesizes high-fidelity 3D scenes from text by using a hybrid scene representation, employing DMTets~\cite{shen2021deep} for objects of interest and NeRF~\cite{nerf} for the environment.
For large-scale, nature or city scene generation, Citygen~\cite{deng2023citygen} generate infinite and controllable 3D layouts by representing the 3D city layout with a semantic field and a height field. 
WonderJourney~\cite{yu2023wonderjourney} employs ChatGPT-generated text prompts to guide the image generation process, resulting in diverse and automated scene generation.
Besides generating 3D scenes from single or multi-view 2D images, another direction involves directly generating 3D scenes through text prompts or image guidance. XCube~\cite{ren2023xcube} uses a multi-resolution coarse-to-fine shape generator with sparse voxel grid representation to generate high-resolution scenes such as streets. BlockFusion~\cite{wu2024blockfusion} leverages Tri-plane diffusion to create 3D scenes as cubic blocks, enabling large-scale unbounded scene generation with a novel tri-plane extrapolation mechanism. Frankenstein~\cite{yan2024frankenstein} extends Tri-plane diffusion for building a compositional scene generation tool.

\begin{figure*}[h]  
  \centering
  \includegraphics[width=0.98\linewidth]{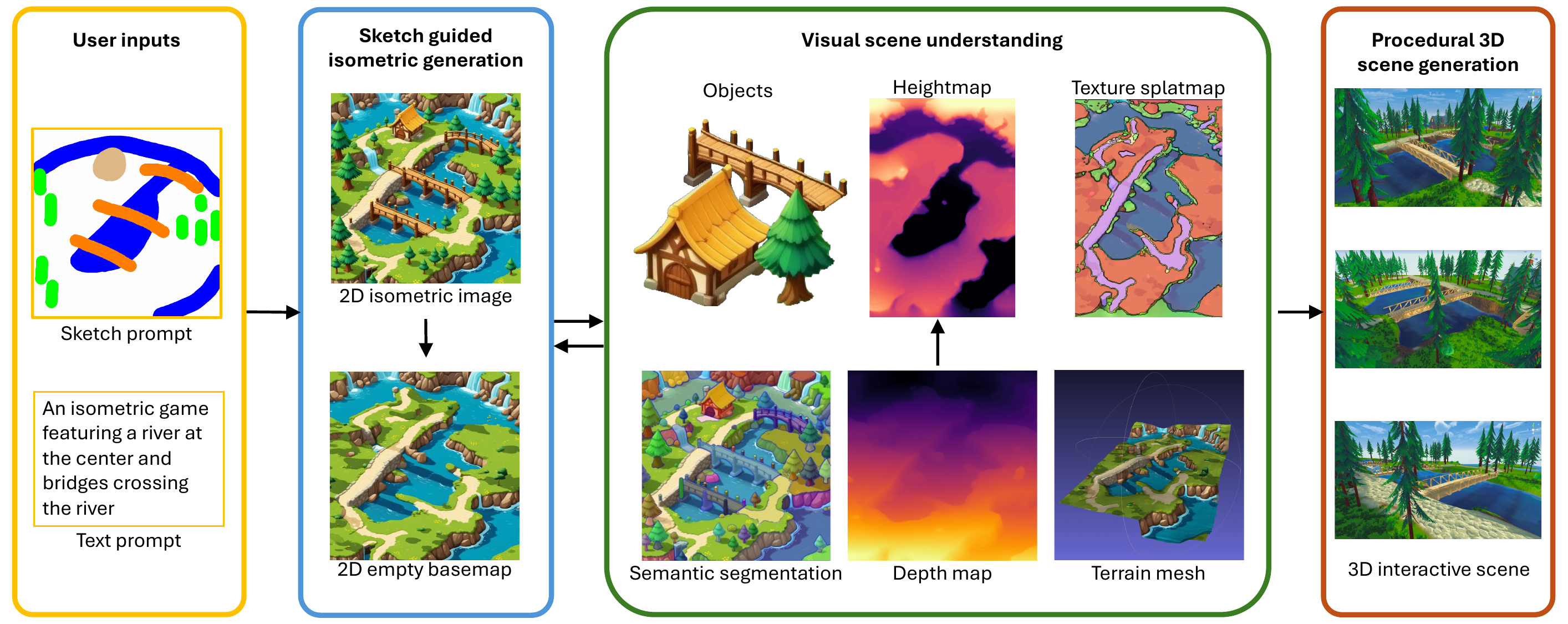} 
  \caption{\small {\bf Overview of the pipeline of the proposed method.} The input user sketch and text prompt are fed into our pre-trained ControlNet that generates a 2D isometric reference image.
  Our Scene-Understanding module then extracts the foreground object masks.
  The masks are fed to a pre-trained inpainting model which generates the isometric empty basemap (i.e., the background terrain with no objects). 
  The scene understanding module also computes the heightmap, texture splatmap and object instance pose. Finally, a procedural 3D scene generation module is employed to generate and render the 3D game scene.  {\label{fig:pipeline}}}
\end{figure*}

\subsection{Procedural generation}

Past solutions for 3D scene generation primarily focused on procedural generation methods using modifiable parameters and rule-based systems. Here we focus on combined solutions using large language models (LLMs) or diffusion models for controllable 3D scene generation, as a comprehensive listing of all works would exceed the scope of this paper. 3D-GPT~\cite{sun20233dgpt} introduced a framework using LLMs to generate Python codes for 3D modeling, enhancing real-world flexibility of~\cite{raistrick2023infinite}. SceneX~\cite{zhou2024scenex} improves LLM-guided scene generation by automating high-quality scene creation from textual descriptions using a large 3D asset database and a planner for task planning, asset retrieval, and action execution.

\section{Method}
Figure~\ref{fig:pipeline} provides an overview of our pipeline, which comprises three key modules: Sketch Guided Isometric Generation, Visual Scene Understanding, and Procedural 3D Scene Generation. The following subsections will describe each module in detail.
\subsection{Sketch Guided Isometric Generation}
\subsubsection{2D Isometric Image Generation}
Starting from a casual user sketch, our first task is to generate a 2D conceptual illustration of the 3D scene.  To this end, we propose to use a pre-trained 2D image (denoising) diffusion model to generate an oblique view of the 3D scene using the isometric projection model.  Isometric projection is a special orthographic camera projection where the coordinate axes with the same dimension have equal length, and the angle between each pair of axes is $120^\circ$.  We use this type of projection mainly for its simplicity in handling occlusions.  

We employ ControlNet~\cite{contolnet} to provide the user with precise control in the layout of generated scene. ControlNet allows a pre-trained text-to-image diffusion model to have additional spatial conditioning during the denoising steps.  We train our sketch-based conditioning with one-hot encoding with $N$ channels, where each channel corresponds to a unique sketch category (e.g. building, road, water, bridge, etc.). Compared to the more commonly used RGB pixel-domain conditioning, one-hot representation possesses the benefit of simpler training complexity and allows category overlap. 

\begin{figure} 
  \centering
  \includegraphics[width=0.9\linewidth]{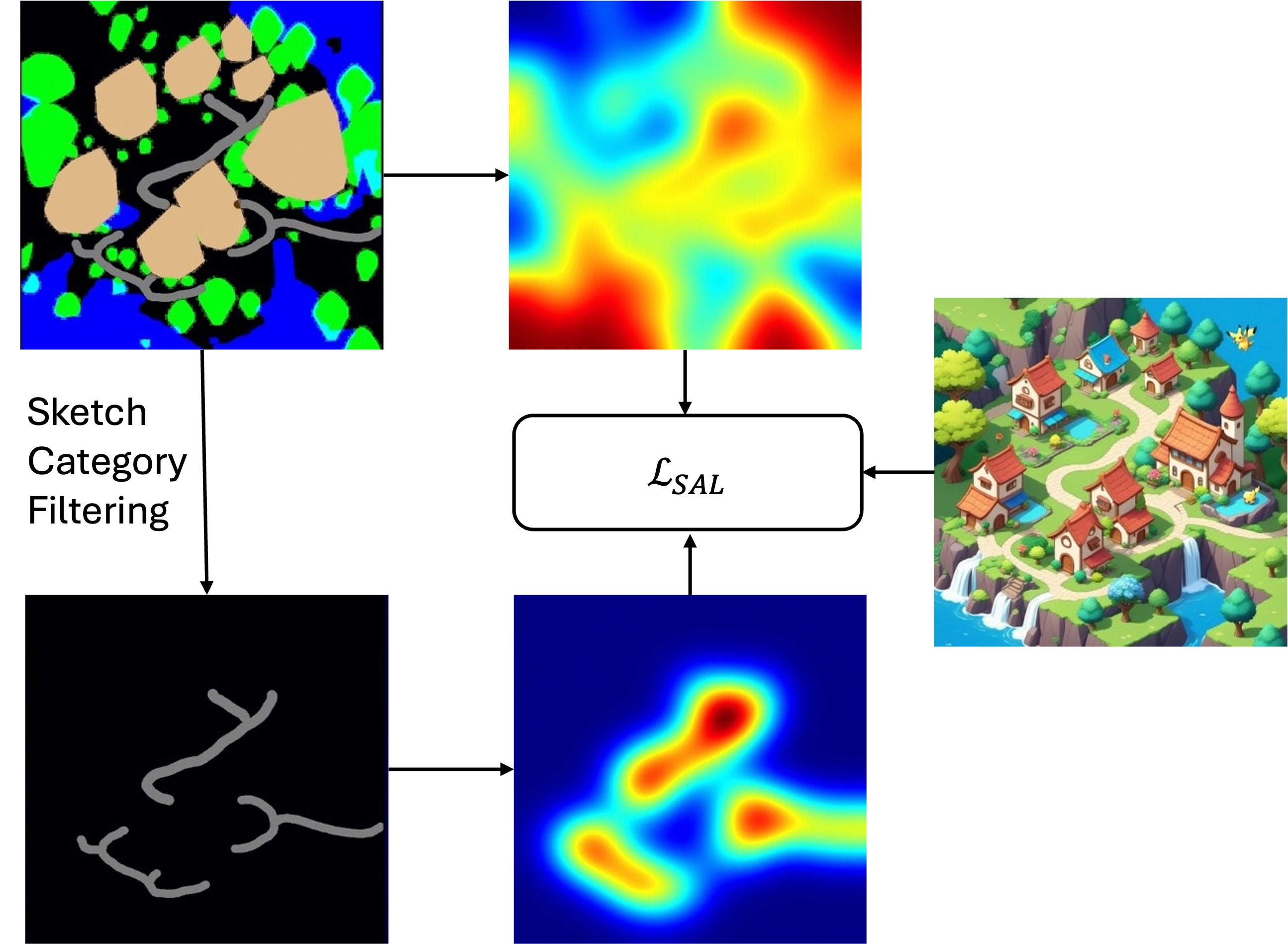} 
  \caption{The Sketch-Aware Loss (SAL) facilitates ControlNet's training with a single ground truth image associated with diverse sketches generated through random category filtering, thereby enhancing its performance on flexible sketches. }
  \label{fig:controlnet_rcf}
\end{figure}

Our method only requires the user to provide a casual guidance via a hand-drawn sketch with arbitrary number of categories. Once the sketch is provided,  our method should be able to fill in the blank regions with plausible and compatible contents.  For instance, if the user draws a few houses, the model should be able to generate a road network and trees that are naturally align well  with the houses, leading to a harmonic scene. To enable this flexibility in the input sketch, the model should be trained using sketches with a diverse combination. For example, the same water map associates with different roads, or the same roads combine with different buildings. Thus, we conducts sketch category filtering that augments the sketch by randomly dropping out each category. As shown in Fig.~\ref{fig:controlnet_rcf}, the sketch of a reference image is augmented to a new one by removing other categories but road.

The training of above augmented data does not work directly since all augmented sketches correspond to the same ground truth as illustrated in Fig.~\ref{fig:controlnet_rcf}. To address this issue, we introduce a new loss function, namely, the Sketch-Aware Loss (SAL). A soft-mask is created for each sketch and is applied as the loss weight matrix to encourage the supervision of ControlNet to focus on valid regions in the sketch. The weight is obtained by convolving the sketch mask using a Gaussian kernel, as depicted in the middle column of Fig.~\ref{fig:controlnet_rcf}. This means a higher weight is applied close to the user's sketch and vice versa. Let $\omega = \text{max}(0.1,  \mathcal{G} (f(S)))$, the resulting mask is incorporated into the following loss:
\begin{equation}
 \mathcal{L}_{SAL} = \mathbb{E}_{x_0,t,c_t,c_s,\epsilon \sim \mathcal{N} (0,1)}[ \| (\epsilon - \epsilon_{\theta} (x_t, t, c_t, c_s)) \cdot \omega \|_2^2 ],
\end{equation} 
where $S$ is the one-hot sketch, $f$ computes the maximum along the sketch channels (equivalent to \emph{any} operator in boolean array), $\mathcal{G}$ is a standard Gaussian convolution with 11$\times$11 kernel, $c_t$ is text prompt, $c_s$ is sketch condition.

\subsubsection{2D Empty Terrain Extraction}
A clean reference image of the empty terrain (aka. the ``basemap") is needed to recover the corresponding 3D terrain of the scene.  In the generated 2D isometric image, there are still some occluded regions of the terrain due to the presence of foreground objects. For instance, the ground on the far side of a building is not visible. Unlike general inpainting tasks, this is challenging due to the requirement that the inpainted region must not contain any foreground object.  Existing context-based inpainting methods struggle with filling such large masks due to a lack of prior knowledge. While diffusion-based generative inpainting methods show its potential, current state-of-the-art (SOTA) methods such as RePaint~\cite{lugmayr2022repaint}, EditBench~\cite{wang2023imagen}, and Stable Diffusion XL Inpaint (SDXL-Inpaint) ~\cite{sdxlinpaint} do not produce satisfactory results, even with carefully designed prompts. (cf. Fig~\ref{fig:inpaint_on_control_out2})

To solve this, we fine-tune the LoRA \cite{hu2021lora} on SDXL-Inpaint to learn the distribution of the basemap and foreground masks. To overcome the obstacle of the lack of isometric-basemap datasets for training, we collected a training dataset from three types of data sources: isometric images with foreground objects, perspective images of empty terrain, and terrain texture images. When using isometric images with foreground objects for training, the inpainting mask is designed to have no overlap with the foreground object. On the other hand, the other two types of training data use foreground masks that are randomly extracted from other isometric images intersected with random shapes.  
\paragraph{Training objective}
The original SDXL-Inpaint is constructed from a 9-channel input UNet, with the loss function defined as:
\begin{equation}
 \mathcal{L}_{inp} = \mathbb{E}_{t,x_0,m,c_t,\epsilon \sim \mathcal{N} (0,1)}[ \| \epsilon - \epsilon_{\theta} (y_t, t, c_t) \|_2^2 ] \;.
\end{equation}
Here,  
\begin{equation}
 y_t = \text{cat}(x_t, m, \text{E}( (1-m) \cdot x_0)) \;,
\end{equation}
\begin{equation}
 x_t = \sqrt{\bar{\alpha_t}}x_0 +  \sqrt{1-\bar{\alpha_t}}\epsilon \;.
\end{equation}
In these equations, $x_0$ represents the ground truth image, $c_t$ denotes text prompts, $m$ is the binary foreground mask, $\epsilon$ is the random Gaussian noise, and $\text{E}$ is an image encoder.
Compared to the standard text-based diffusion model \cite{ho2020denoising}, this inpainting model retains the same forward diffusion strategy, but it concatenates the mask and inverse-masked latent image into the denoising input. We employ $\mathcal{L}_{inp}$ for our curated ``ideal'' ground truth images. 
We increase the size of the training dataset by also incorporating isometric images with foreground objects (full isometric), where only a partial background region can be used as training ground truth. In this case, we simply add noise and learn the denoising of the background area:
\begin{equation}
 \hat{x_t} = m \cdot (\sqrt{\bar{\alpha_t}}x_0 +  \sqrt{1-\bar{\alpha_t}}\epsilon ) + (1-m) \cdot x_0 \;,
\end{equation}
\begin{equation}
 \hat{y_t} = \text{cat}(\hat{x_t}, m, \text{E}(m\cdot x_0)) \;,
\end{equation}
\begin{equation}
 \mathcal{L}_{inp}^{part} = \mathbb{E}_{t,x_0,m,c_t,\epsilon \sim \mathcal{N} (0,1)}[ \| m \cdot (\epsilon - \epsilon_{\theta} (\hat{y_t}, t, c_t)) \|_2^2 ] \;.
\end{equation}
During the training phase of the inpainting model, all three types of training data are thoroughly shuffled and randomly sampled.

Another obstacle that hinders the inpainting performance is caused by a distribution shift of denoising between training and inference. This shift occurs in two ways: masked regions are background during training, while masked regions are foreground during inference. Additionally, despite our efforts to mimic real foreground masks by intersecting pseudo foreground masks with random shapes, a slight discrepancy remains. 
Step-Unrolled Denoising (SUD) diffusion technique~\cite{saxena2023monocular} is designed to tackle this issue. We adapted it in our inpainting process, as detailed in Algorithm~\ref{alg:inpaint_sud}. Note that the SUD step is applied only in the later stages of the training, as it is only effective when the prediction can produce plausible results. 
\begin{algorithm}
\caption{Inpainting training step with SUD} \label{alg:inpaint_sud}

\algnewcommand\algorithmicinput{\textbf{Input:}}
\algnewcommand\algorithmicoutput{\textbf{Output:}}
\algnewcommand\algorithmictrainloop{\textbf{Training loop:}}
\algnewcommand\INPUT{\item[\algorithmicinput]} 
\algnewcommand\OUTPUT{\item[\algorithmicoutput]}
\algnewcommand\TRAINLOOP{\item[\algorithmictrainloop]}

\begin{algorithmic}

\INPUT $rgb\ image\ x_0,\ background\ mask\ m_{bg},\ text\ prompt\ c_t $ 
\TRAINLOOP 
\State $t \gets U(0,1)$
\State $\epsilon \gets \mathcal{N}(0,1)$
\State $ m_{pfg} \gets pseudo\ foreground\ mask\ library $
\State $ m=m_{pfg} \cdot m_{random} \cdot m_{bg} $
\State $ x_t = \sqrt{\alpha_t} x_0 + \sqrt{1-\alpha_t}\epsilon $
\If {$unroll\_step$}  
    \State $ \hat{m} \gets mask\_foreground $ 
    \State $ No\ gradients: \hat{\epsilon_{pred}} = f_{\theta}([x_t,\hat{m},(1-\hat{m})\cdot x_0], t) $
    \State $ \hat{x_{pred}} = (x_t-\sqrt{1-\alpha_t} \hat{\epsilon_{pred}}) / \sqrt{\alpha_t} $
    \State $ \hat{x_t} = \sqrt{\alpha_t}\hat{x_{pred}} + \sqrt{1-\alpha_t} \epsilon $
    \State $ \bar{\epsilon} = (\hat{x_t} - \sqrt{\alpha_t} x_0) / \sqrt{1-\alpha_t} $
    \State $ \bar{\epsilon_{pred}} = \epsilon_{\theta}([\hat{x_t},m,(1-m)\cdot x_0], t, c_t) $
    \State $ \mathcal{L} = \| \bar{\epsilon} - \bar{\epsilon_{pred}} \|_2^2 $
\Else
    \State $ \epsilon_{pred} = \epsilon_{\theta}([x_t,m,(1-m)\cdot x_0], t, c_t) $
    \State $ \mathcal{L} = \| \epsilon - \epsilon_{pred} \|_2^2 $
\EndIf
 
\end{algorithmic}
\end{algorithm}

\subsection{Visual Scene Understanding}
We decompose the 3D scene into three main components, namely: terrain heightmap, texture splatmap, and foreground objects. The heightmap controls the terrain's shape. The texture splatmap along with its corresponding texture tiles determines the terrain's texture and color. Splatmaps are commonly used in game engines that act as an alpha composition of tiled texture to obtain a textured terrain. Foreground objects' instance and pose establish the type, location and direction of 3D objects being placed into the scene. 

\subsubsection{Terrain HeightMap}
After the basemap inpainting, there are still regions of the scene that are partially occluded, for example, the backside of a mountain. 
We reconstruct a coarse, but watertight 3D terrain mesh from the inpainted 2D terrain map. This mesh will be the foundation for parsing game terrain parameters, enabling high-fidelity scene generation within the game environment. Unlike previous approaches which rely on incremental scene reconstruction, our method takes advantage of the isometric perspective, which offers a comprehensive overview of the environment, with minimum occlusions.  This allows us to recover the majority of colour and depth information of the scene using just a single image.  To infer the scene depth, we adopt the Depth-Anything method~\cite{depthanything}, followed by reprojecting the RGB-D image into space to obtain a colored point cloud. Then, we reconstruct the complete mesh using the Poisson reconstruction technique.

Given the coarse terrain mesh in the isometric viewpoint, one can easily rotate the view to obtain a bird-eye's view (BEV) of the terrain. This provides the depth, $d$ of the terrain from a camera looking directly down along the gravity, and the heightmap, $h$ is simply the reverse of the depth. Specifically, $ h = d_{max} - d$. 

The rough color reference also includes water regions, which are segmented out as previously described. For the water category, we not only add a water asset to the scene but also lower the terrain height in those areas to ensure the terrain is positioned below the water level.

\subsubsection{Texture SplatMap}
The rough terrain mesh provides a rough colour reference when rotated into BEV. However, using this directly for the terrain will result in blurry, low-quality visuals in-game. Popular game engines (e.g. Unity, UE) handle terrain texturing using $N$ texture tiles and $N$ channels splatmap, where the splatmap acts as an alpha composition for the corresponding texture tile. Specifically, we obtain the texture splatmap by performing segmentation using Segment Everything~\cite{sam} on the rendered RGB image of the terrain mesh in BEV, and use Osprey~\cite{Osprey} to obtain the semantic category for each segmentation mask (e.g. grass, rock, road). Then, we automatically pick from a list of texture tiles from the corresponding category and assign them to the terrain. This ensures that the terrain texture remains sharp even when viewed from a close distance. 

\subsubsection{Foreground Objects}
For above-ground objects like buildings or other landmarks, we apply the instance segmentation function of the Sam model ~\cite{sam} to obtain the 2D masks for each of the foreground objects. 

\begin{figure}
  \centering
  \includegraphics[width=0.5\linewidth]{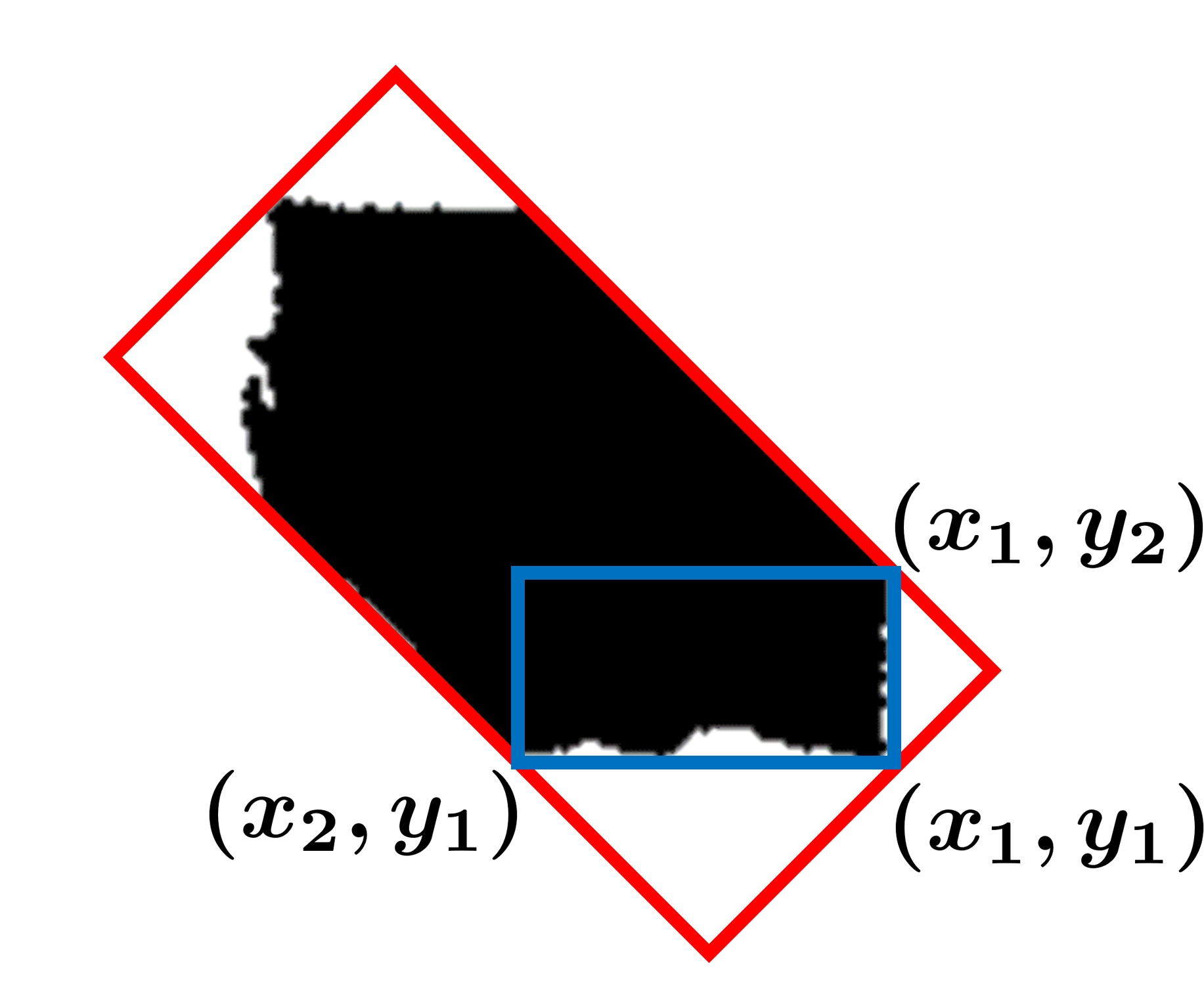}  
  \includegraphics[width=0.35\linewidth]{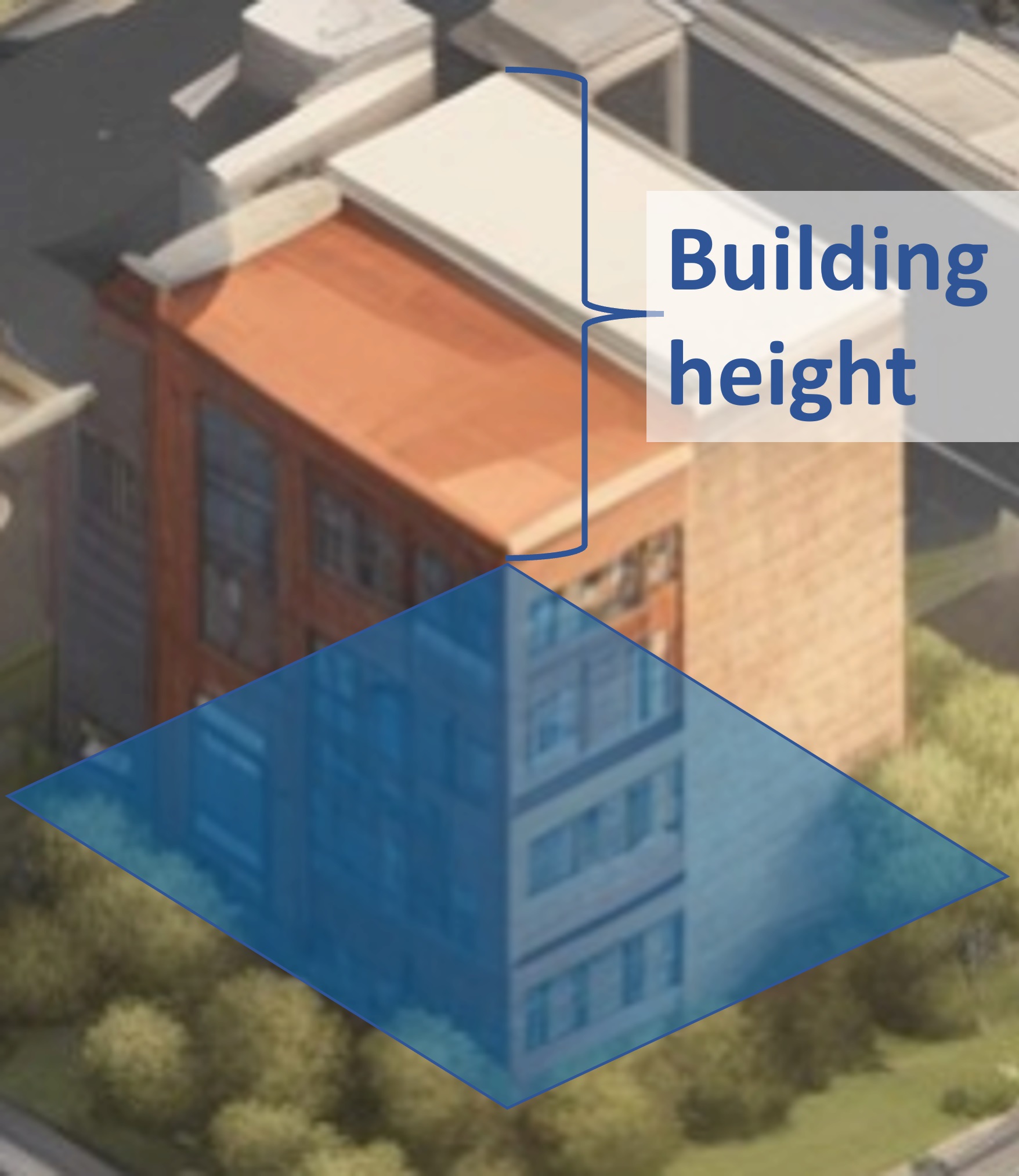} 
  \caption{Object footprint estimation, showing an illustrative example of obtaining a building footprint and height. On the left: Black region is the instance mask of a building, red box shows the homography-warped 2D object bounding box, blue box shows the estimated object footprint. On the right: Blue filled box shows the inverse-homography-warped object footprint, which can also be used to estimate the object height. } 
  \label{fig:footprint}
\end{figure}

The obtained instance segmentation mask of each object helps estimate their pose within the 3D scene. Using the characteristic of isometric images, where objects are typically at a $45^\circ$ angle from the camera, we design a method to estimate their footprints. Exploiting the specific viewpoint of an isometric projection, we warp the instance segmentation image using a homography. 
Then, using the homography-warped 2D object bounding box and instance segmentation from Grounded Segment Anything~\cite{ren2024groundedsam}, we can estimate the object footprint in the rotated view as shown on the left of Fig.~\ref{fig:footprint}. The coordinates $(x_1, y_1)$ is the maximum $x$ and $y$ coordinates of the instance mask. $(x_2, y_1)$ and $(x_1, y_2)$ are the intersection points of $y=y_1$ and $x=x_1$ with the two sides of the warped 2D object bounding box (red box) as shown in the left image of Fig.~\ref{fig:footprint}. 
Thanks to the advantages of the isometric projection, we warp the estimated object footprint back into the isometric image, and estimate the object height as shown in the right image of Fig.~\ref{fig:footprint}. 
Then, using the estimated depth, we transform the object's footprint into its corresponding 3D location. 

\begin{figure}[ht]
    \centering
    \begin{subfigure}{1\linewidth}
    \includegraphics[width=0.32\linewidth]{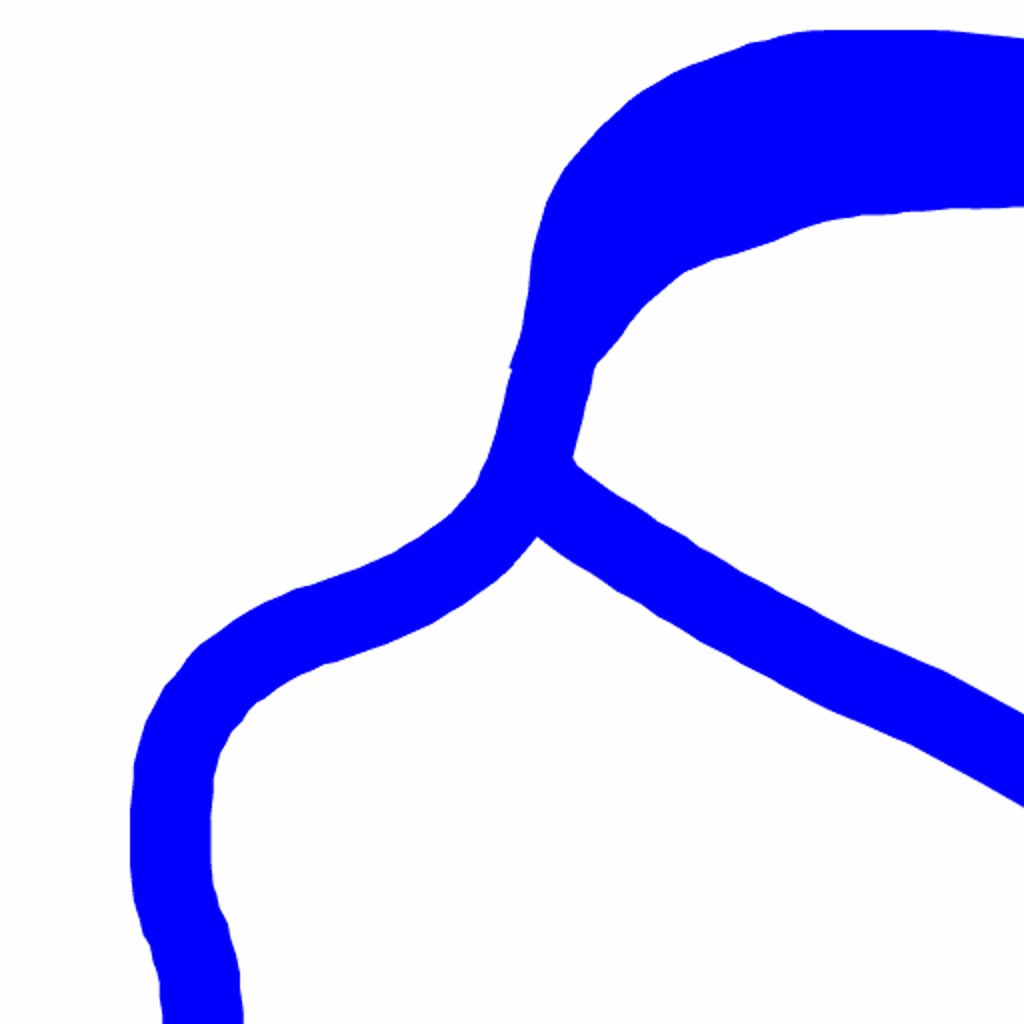}
    \includegraphics[width=0.32\linewidth]{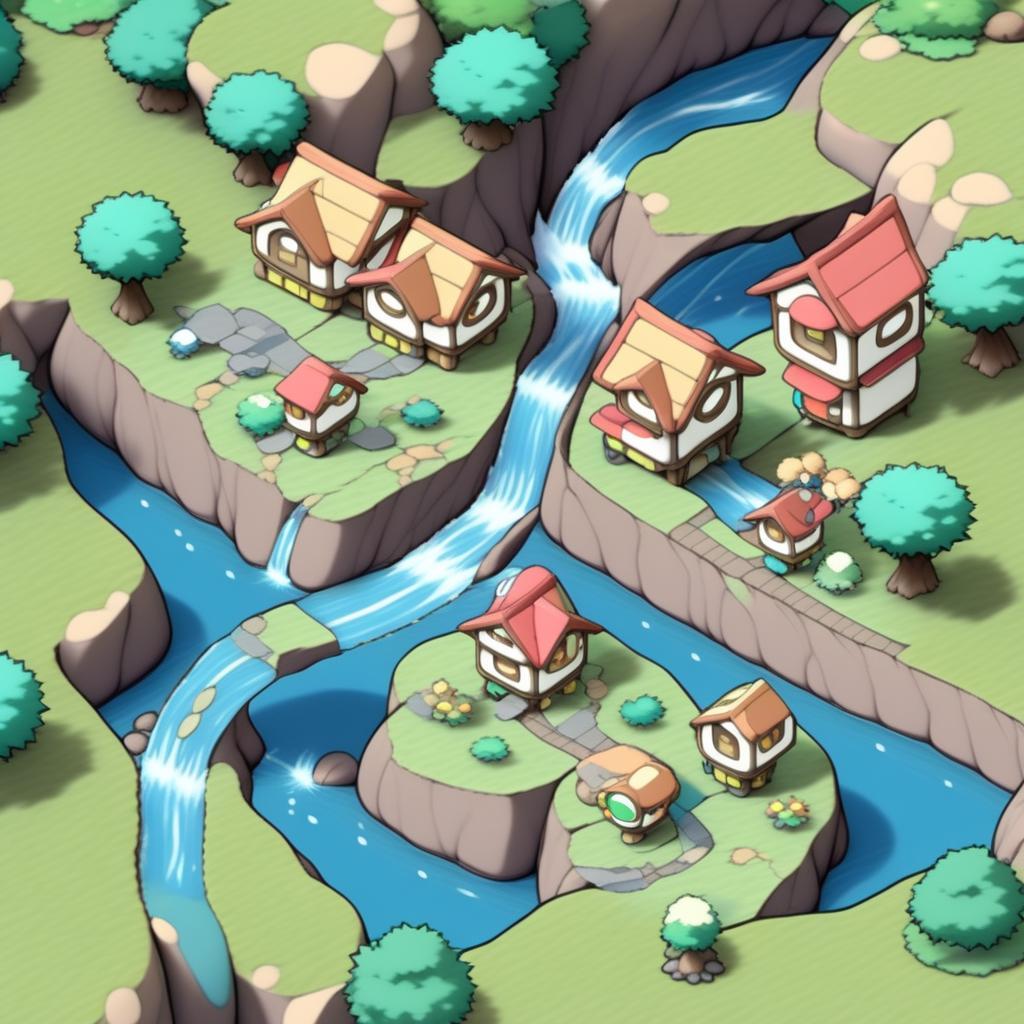} 
    \includegraphics[width=0.32\linewidth]{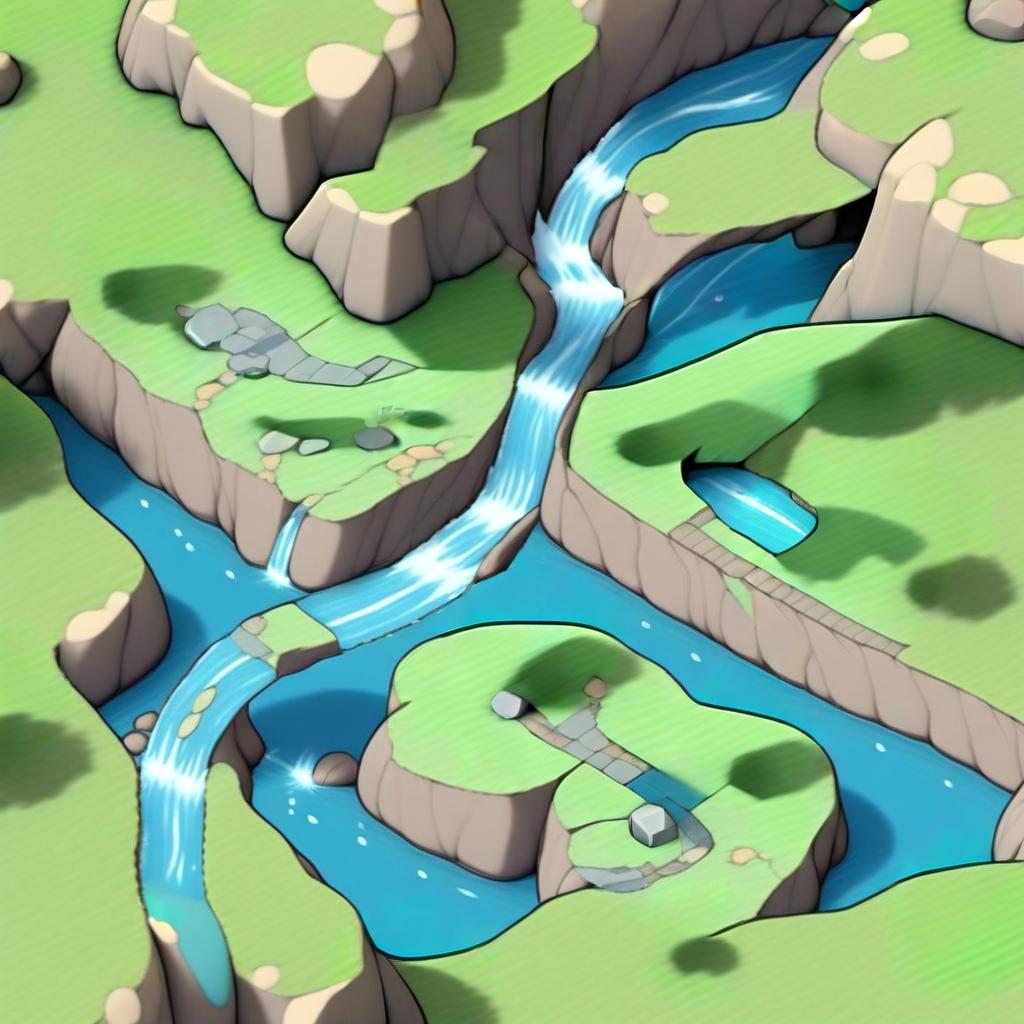} 
    \caption{\em A Pokemon-style isometric town around a crag with a river.}
    \label{subfig:sketch1}
    \end{subfigure}  
    
    \begin{subfigure}{1\linewidth}
    \includegraphics[width=0.32\linewidth]{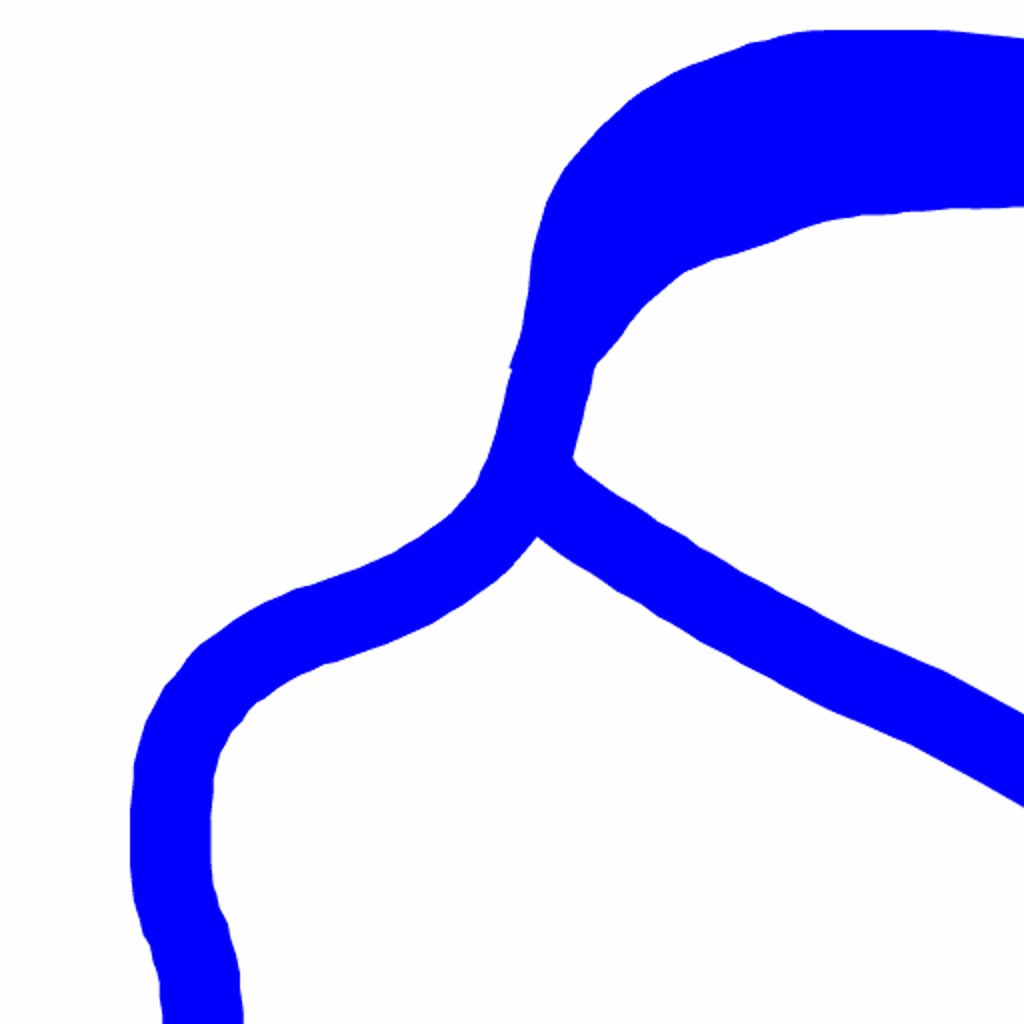}
    \includegraphics[width=0.32\linewidth]{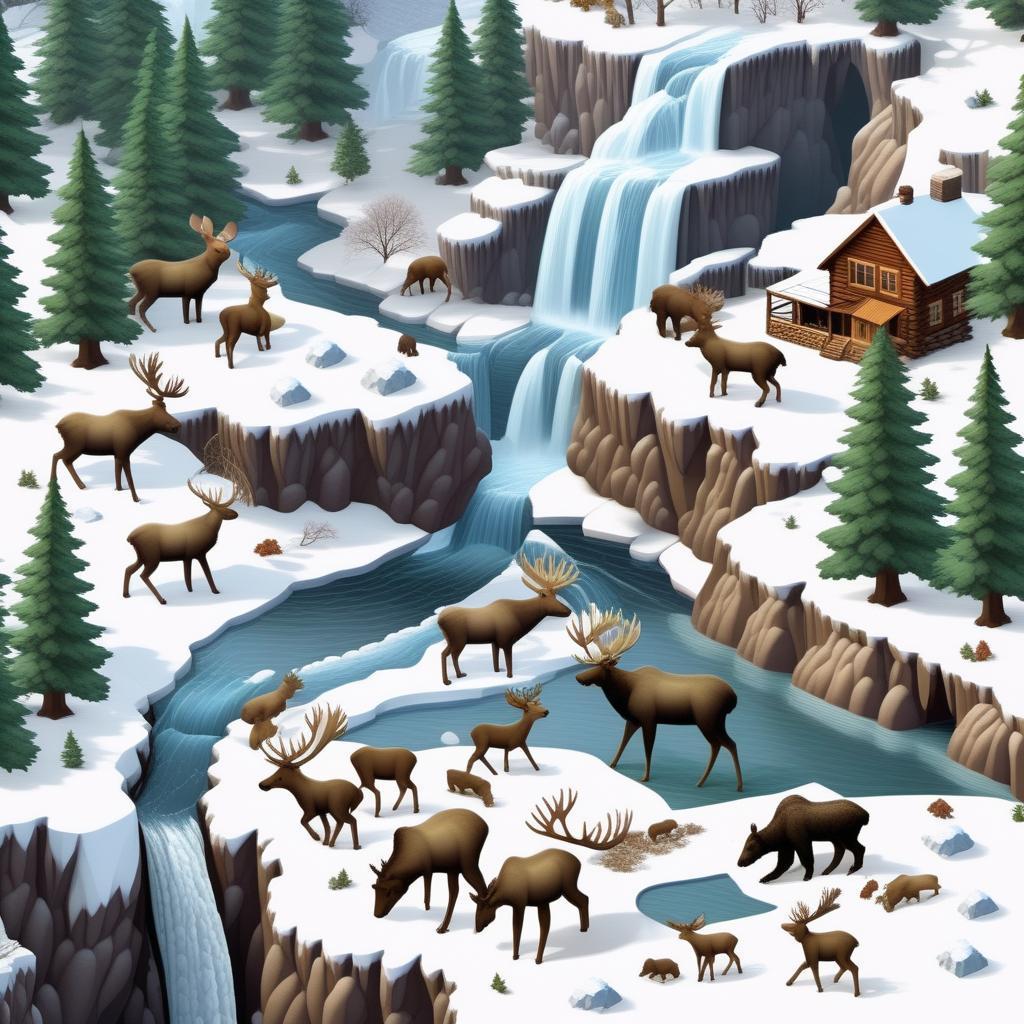} 
    \includegraphics[width=0.32\linewidth]{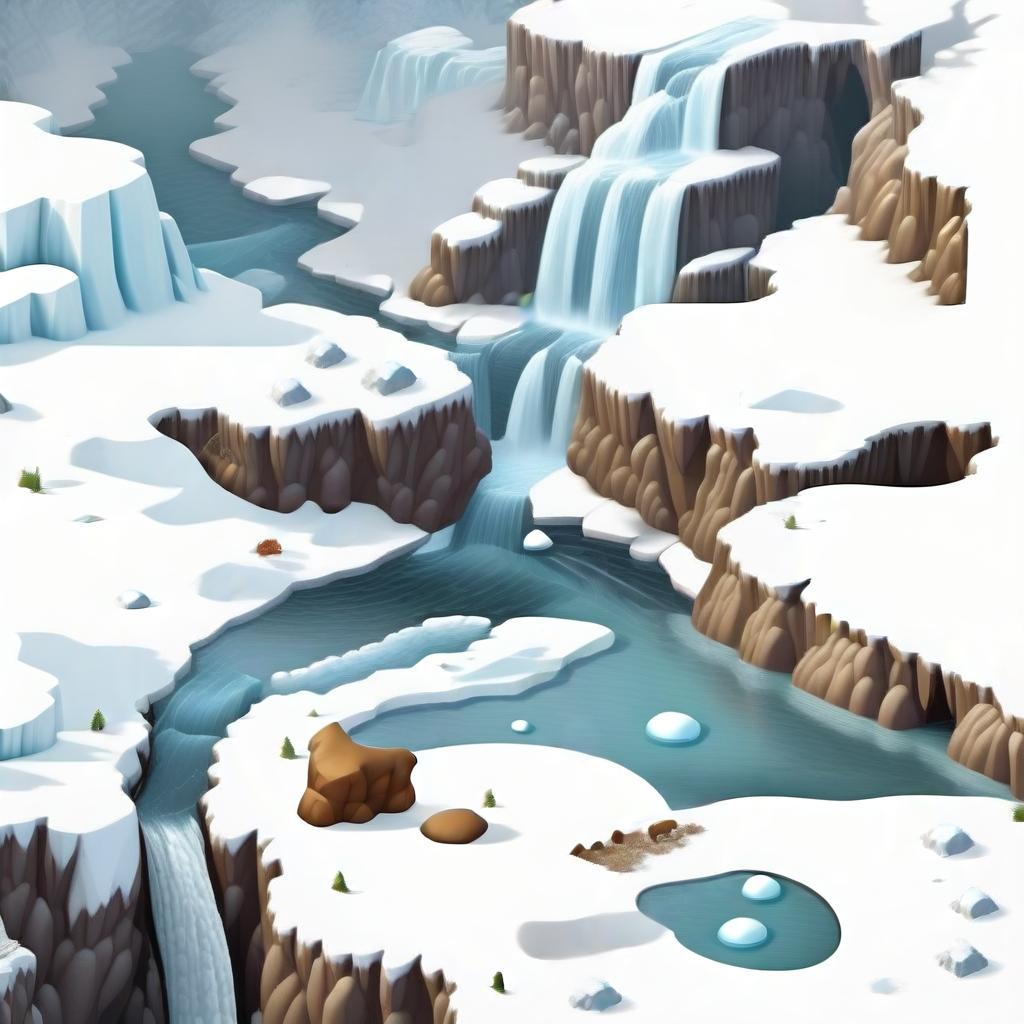 } 
    \caption{\em An isometric view of a snowy landscape with a river, a waterfall, some trees and several animals, such as deers, mooses, and bears. }
    \label{subfig:sketch2}
    \end{subfigure}  
    
    \begin{subfigure}{1\linewidth}
    \includegraphics[width=0.32\linewidth]{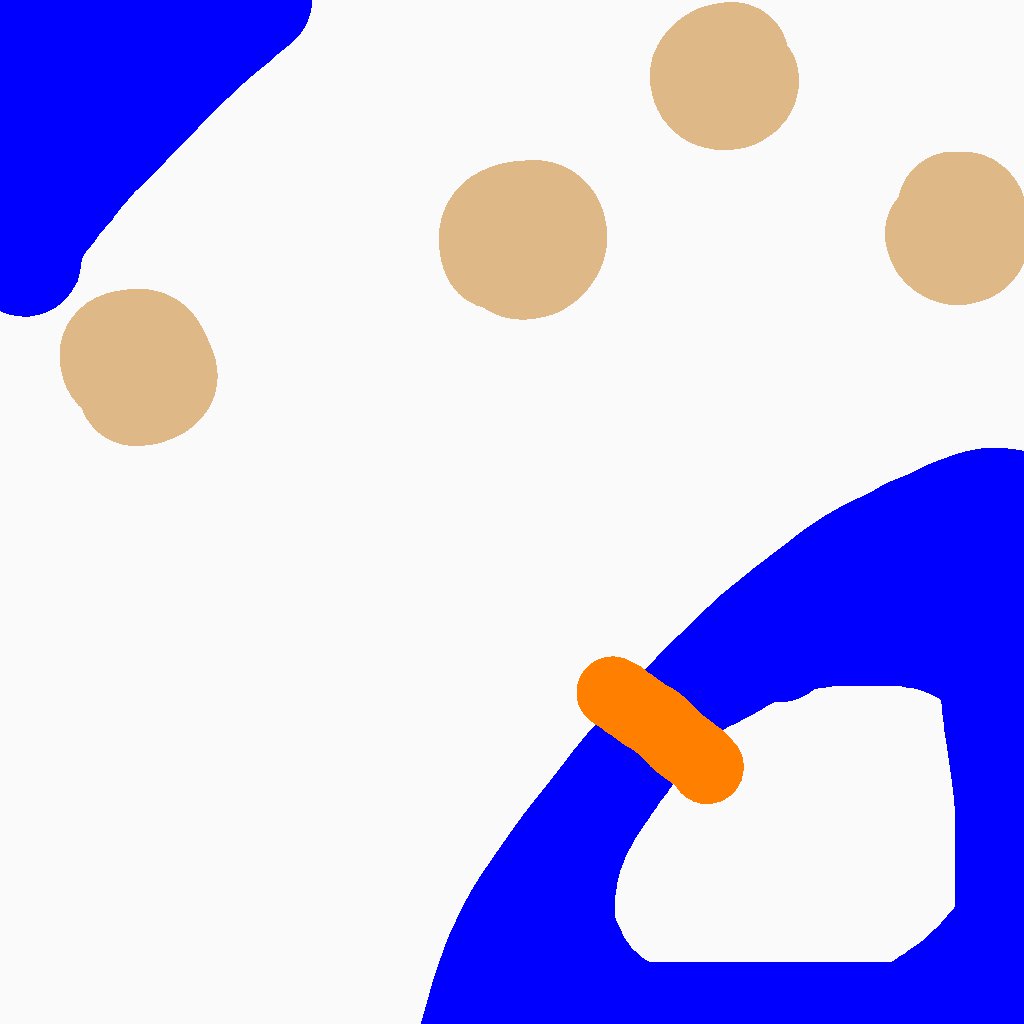}
    \includegraphics[width=0.32\linewidth]{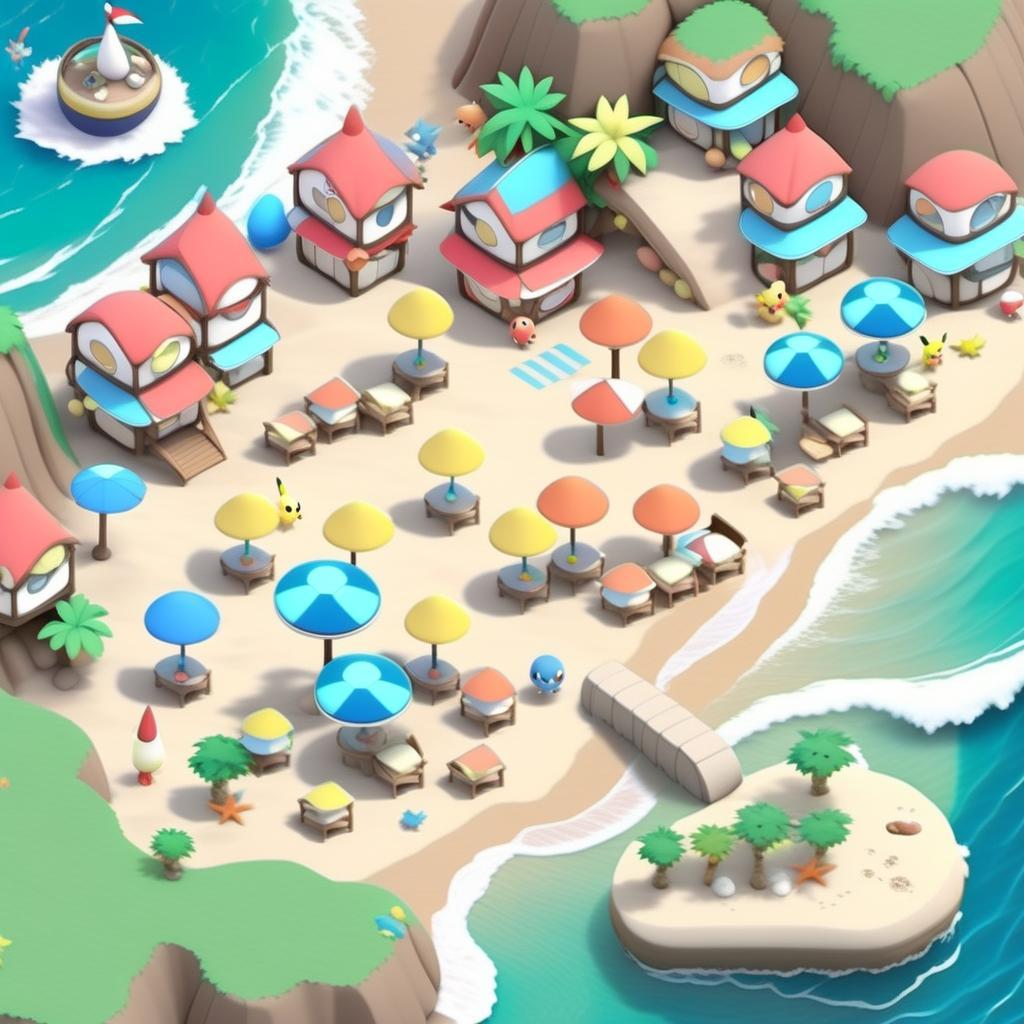} 
    \includegraphics[width=0.32\linewidth]{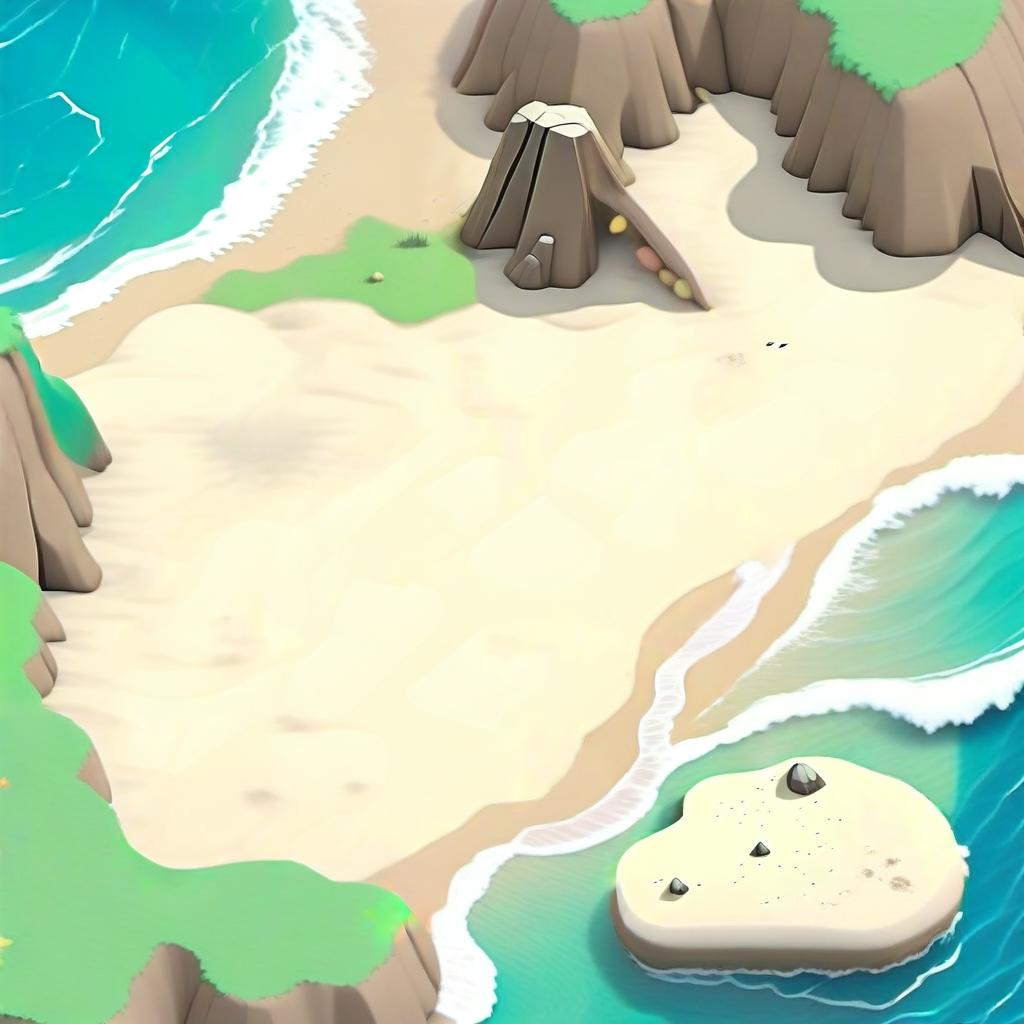} 
    \caption{\em A Pokemon-style isometric town on a beach with buildings and umbrellas.}
    \label{subfig:sketch4}
    \end{subfigure}  

    \caption{\small Results showing the generated isometric reference images (column-2), along with the inpainted basemaps (column-3). Sketch color codes: blue=water, yellow=building, orange=bridge, gray=roads, and green=trees.}
    \label{fig:inpaint_on_control_out}
\end{figure}

\subsection{Procedural 3D Scene Generation}
By leveraging the semantic and geometric understanding obtained in the previous module, we can either use 3D asset retrieval or generation, in combination with procedural generation technique for scene creation. Finally, the 3D scene is composed and rendered within the off-the-shelf 3D game engines (such as Unity or Unreal Engine).
In this work, we use the Unity game engine for building our 3D interactive environment, for Unity offers valuable optimization features for terrain, vegetation, and animation, ensuring optimized runtime performance.  Other game engines or 3D platforms (such as Blender) can be easily used as well. 

Given the heightmap, splatmap and chosen texture tiles, it is straightforward to apply them to a Unity terrain asset. This provides us with a basic 3D terrain featuring high-resolution textures. Depending on the texture type, we can designate the vegetation and small objects that can be placed or grown on them. For instance, a grass texture may include assets like grass, flowers, and rocks, which are placed across the terrain using established procedural content generation techniques. 

For larger objects, we use the segmented instances of the foreground objects (\emph{e.g.} building, bridge) to perform either object retrieval or 3D object generation.  For the former, we search the most similar  instance of 3D object from the Objaverse dataset, by comparing their CLIP scores. 
For the latter, the 3D asset s are generated using recent 2D-to-3D asset generation AI models such as the LRM ~\cite{hong2023lrm} or else \cite{wei2024meshlrm, hyperhuman2024, Yang_2024_CVPR}.  These generated 3D objects are then placed into the scene following the foreground object pose estimated in the previous steps, completing the 3D scene.

\begin{figure}[ht!]
    \centering
    \begin{subfigure}{1\linewidth}
    \includegraphics[width=0.31\linewidth]{fig/Control_Inpaint/ours/3.jpg}
    \includegraphics[width=0.31\linewidth]{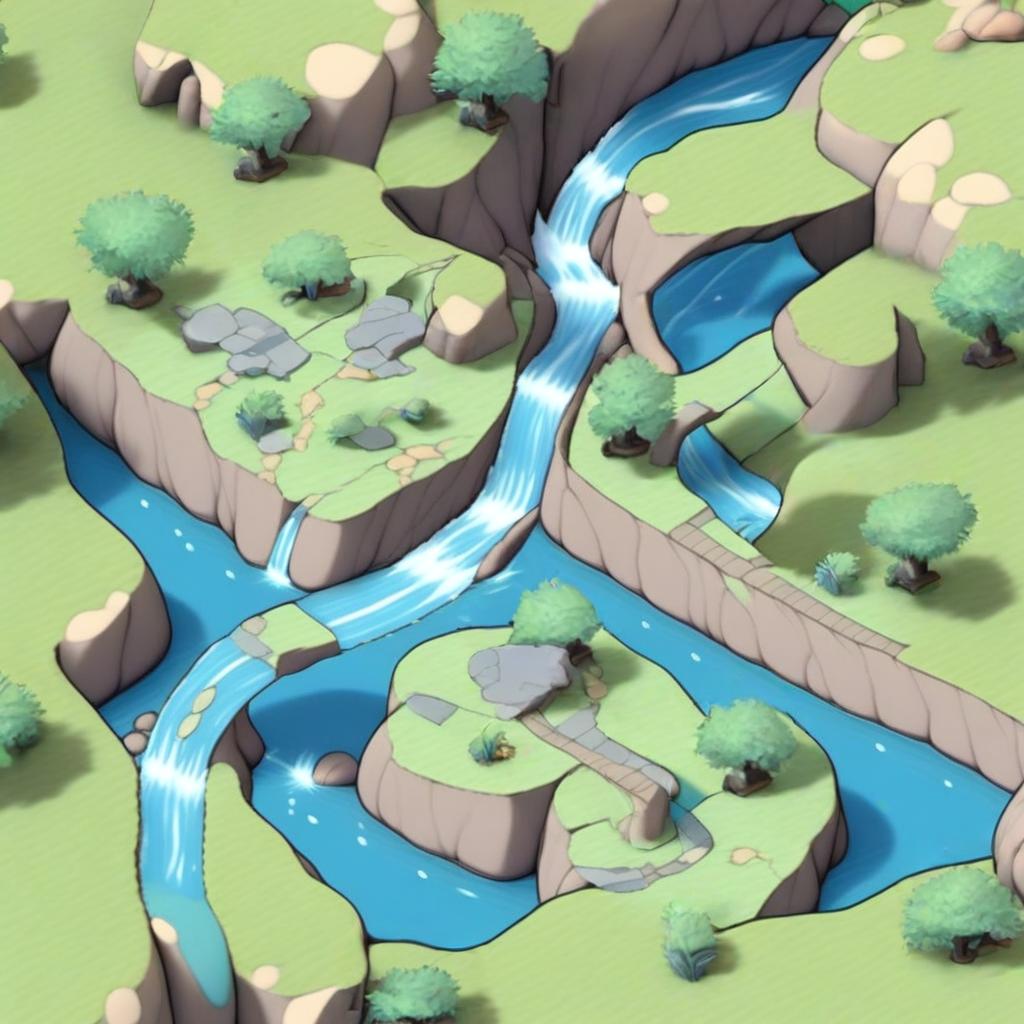}
    \includegraphics[width=0.31\linewidth]{fig/Control_Inpaint/ours/3_whole_image_G75_Str099_Step100_Pro_adap_pro.jpg} 
    \end{subfigure}  
    
    \begin{subfigure}{1\linewidth}
    \includegraphics[width=0.31\linewidth]{fig/Control_Inpaint/ours/15.jpg}
    \includegraphics[width=0.31\linewidth]{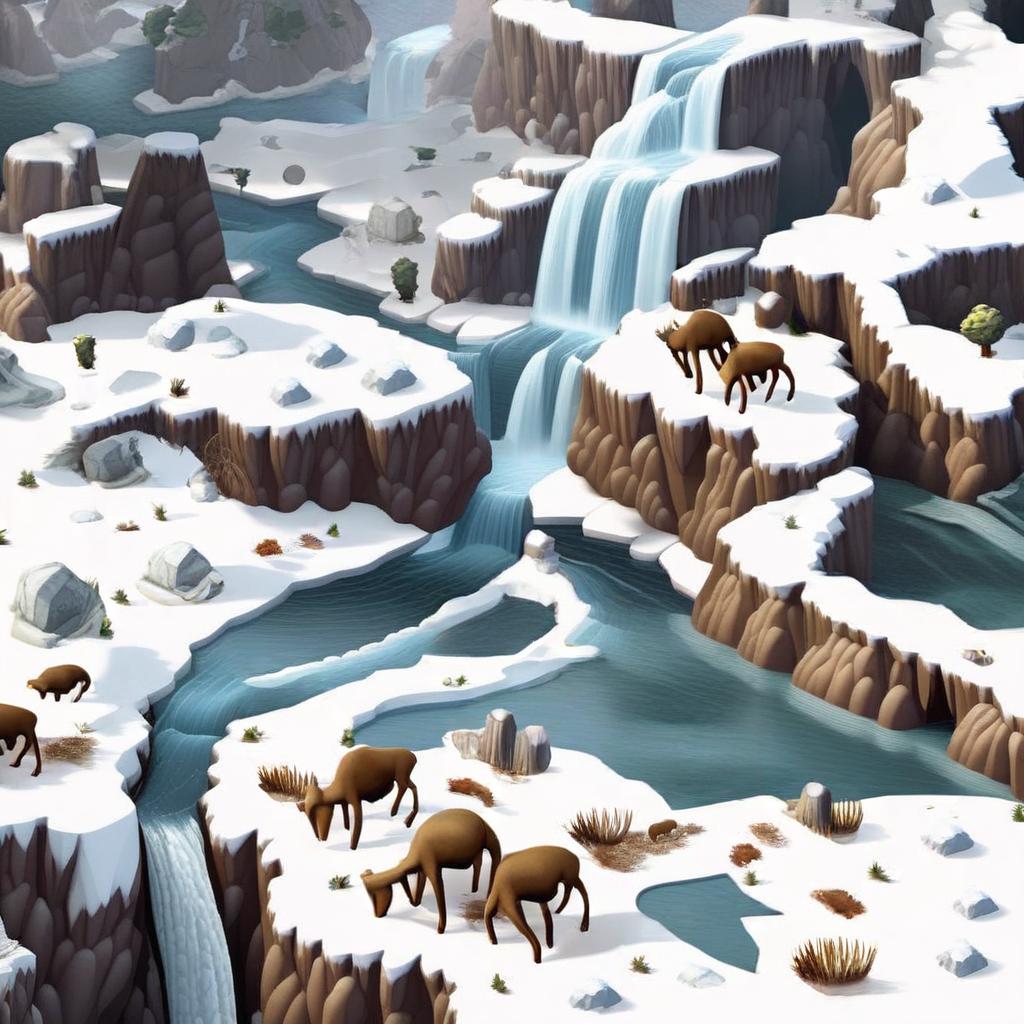}
    \includegraphics[width=0.31\linewidth]{fig/Control_Inpaint/ours/15_cvinpaint_G75_Str099_Step100_Pro_adap_pro.jpg } 
    \end{subfigure}  

    \begin{subfigure}{1\linewidth}
    \includegraphics[width=0.31\linewidth]{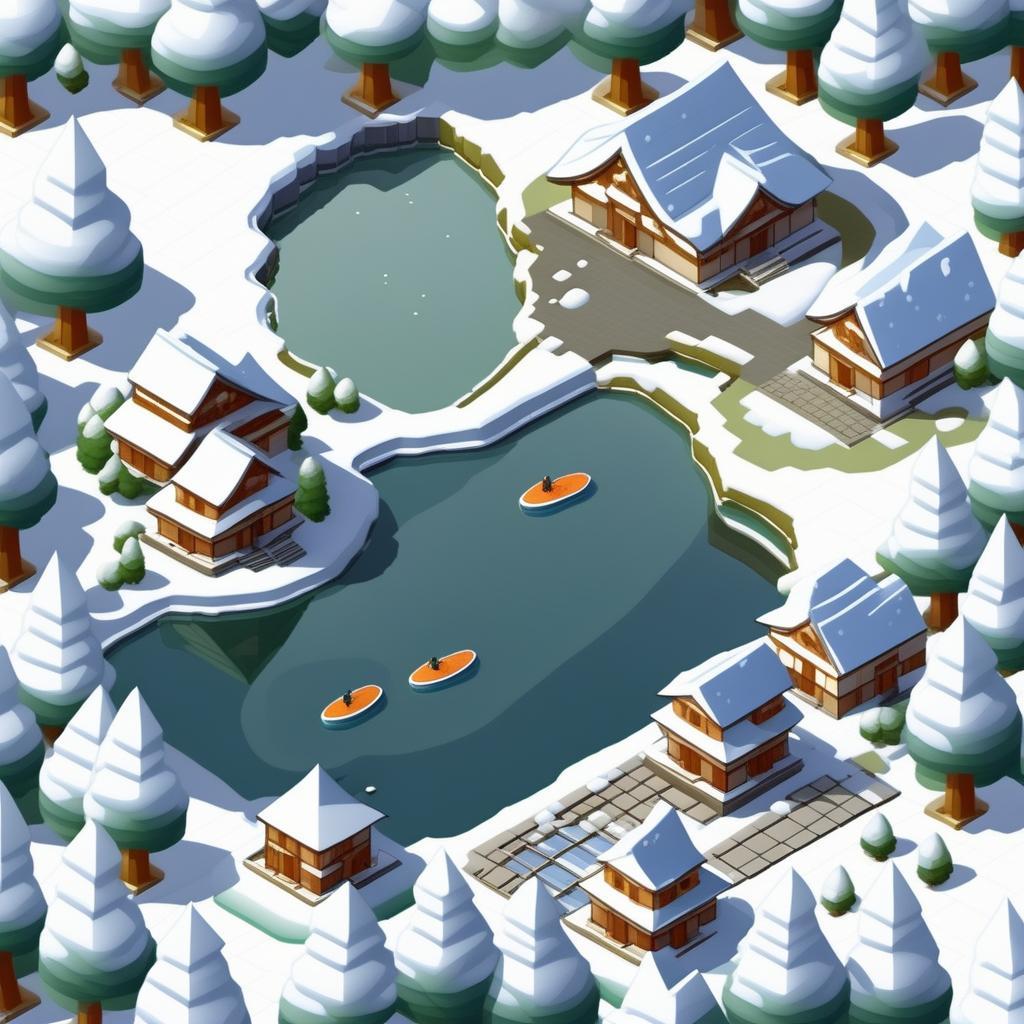} 
    \includegraphics[width=0.31\linewidth]{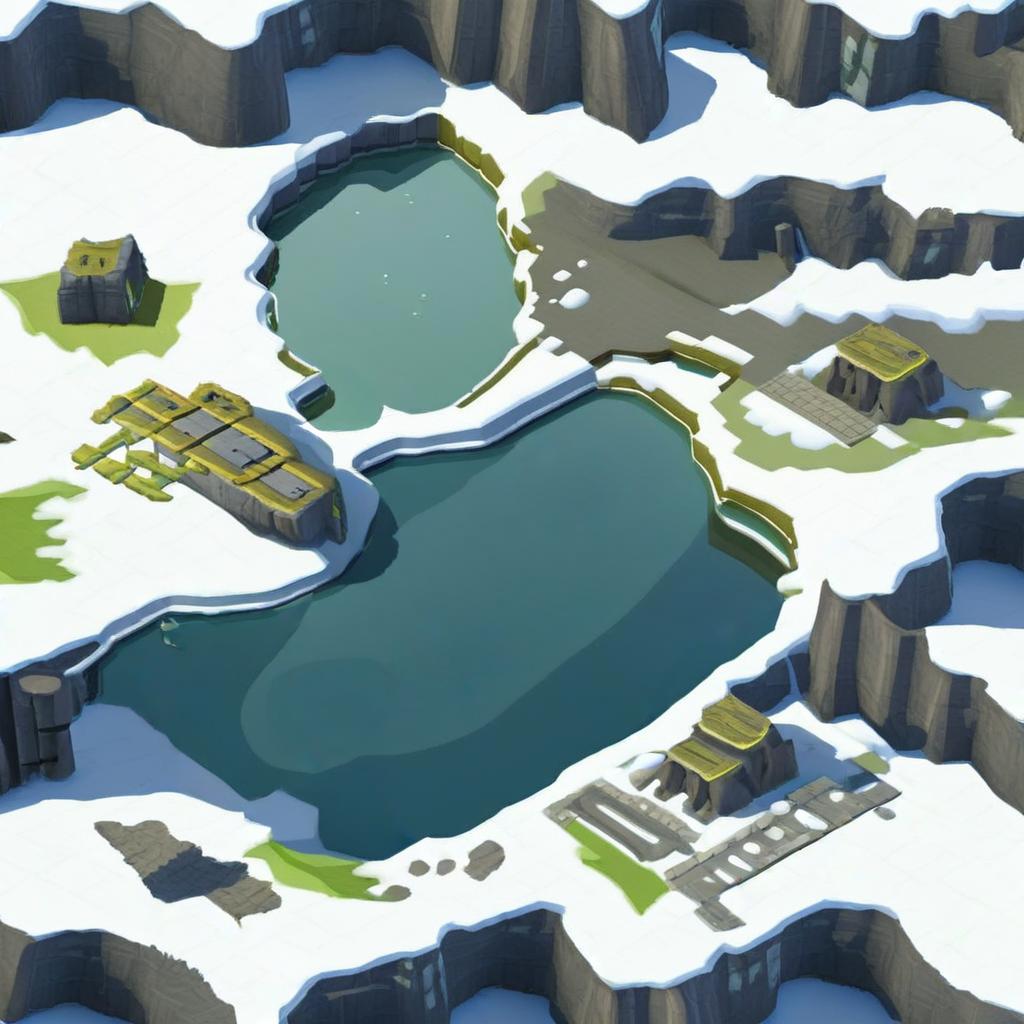}
    \includegraphics[width=0.31\linewidth]{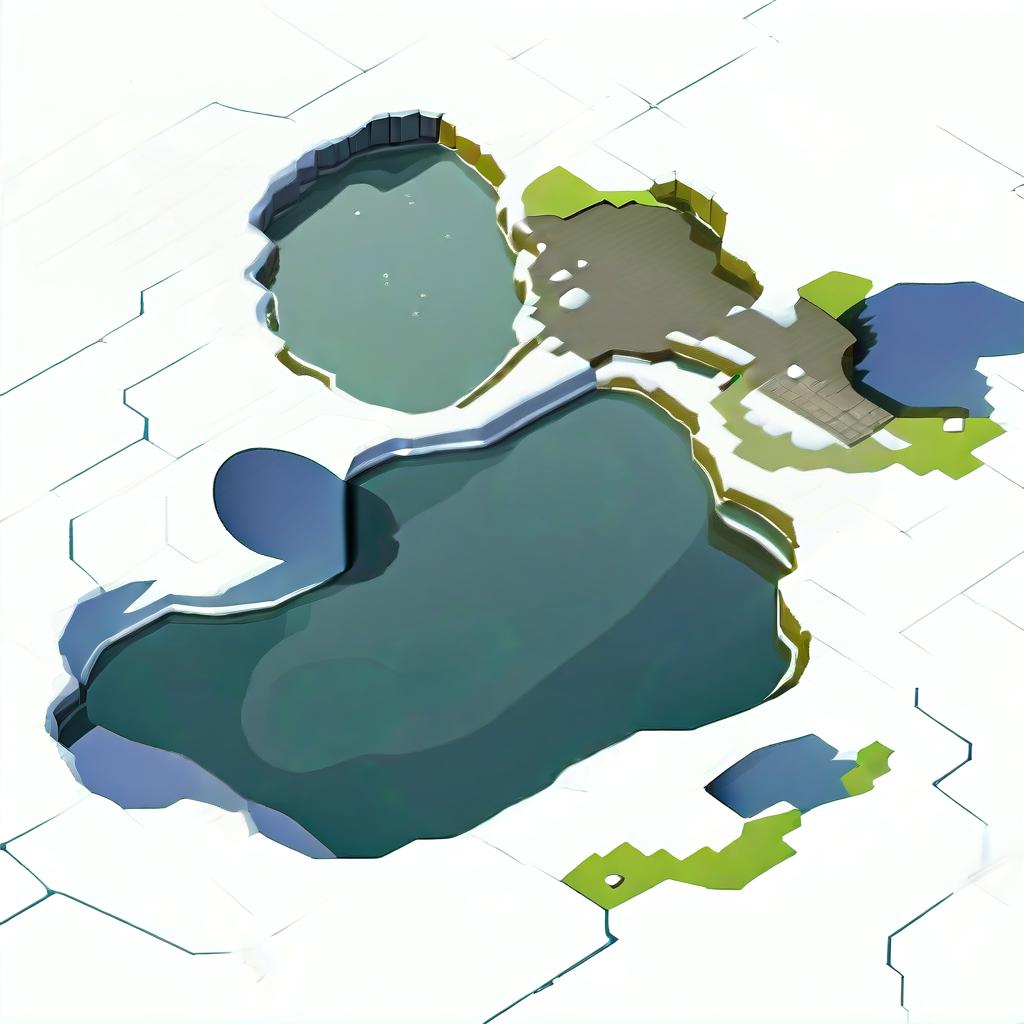 } 
    \end{subfigure}  
    
    \begin{subfigure}{1\linewidth}
    \includegraphics[width=0.31\linewidth]{fig/Control_Inpaint/ours/7.jpg} 
    \includegraphics[width=0.31\linewidth]{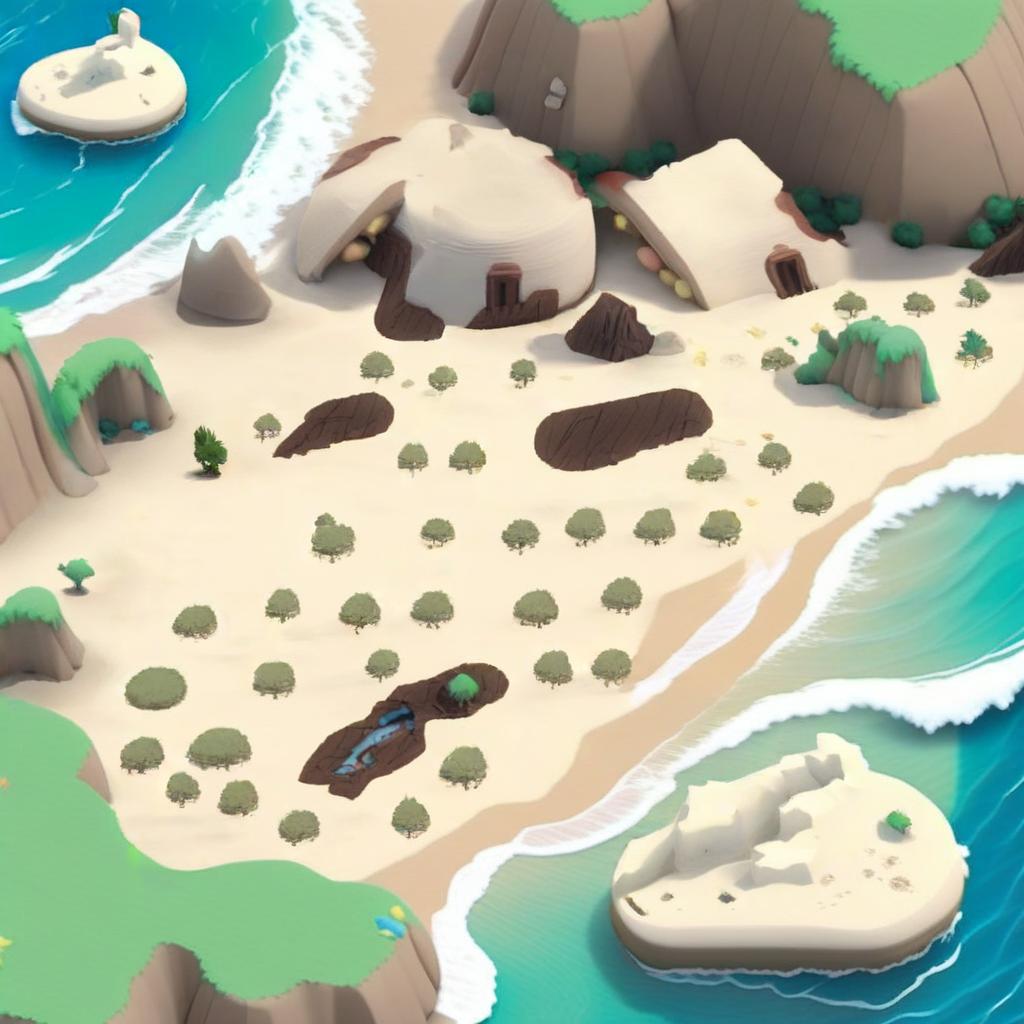}
    \includegraphics[width=0.31\linewidth]{fig/Control_Inpaint/ours/7_cvinpaint_G75_Str099_Step100_Pro_cons_pro.jpg} 
    \end{subfigure}  

    \begin{subfigure}{1\linewidth}
    \includegraphics[width=0.31\linewidth]{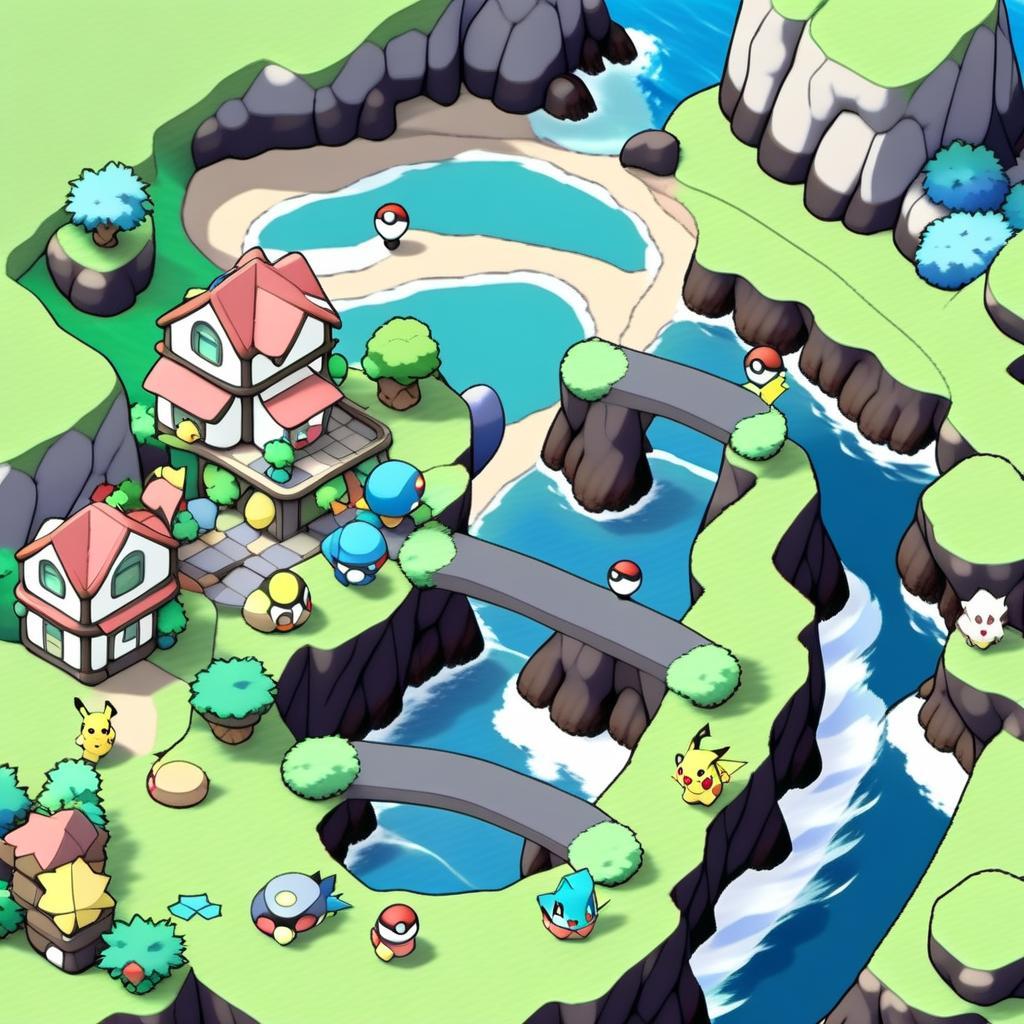}
    \includegraphics[width=0.31\linewidth]{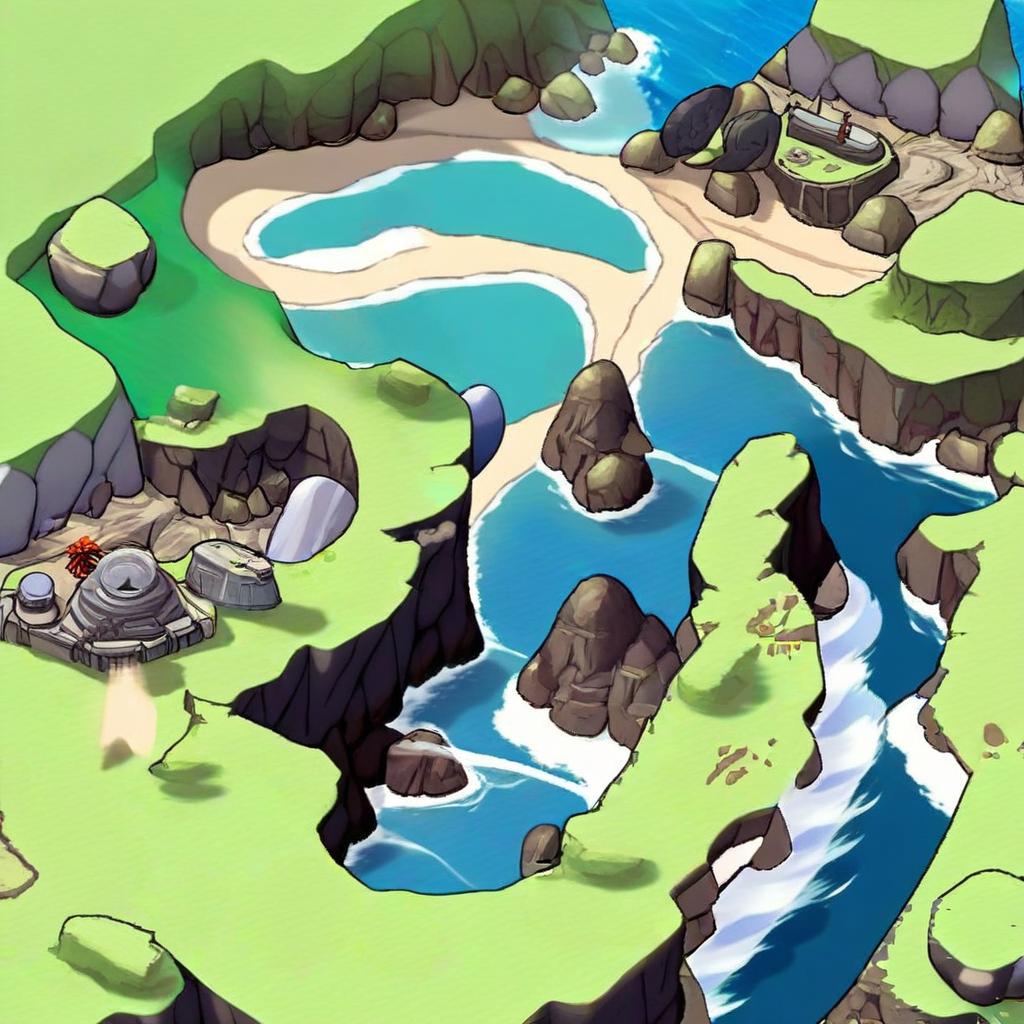}
    \includegraphics[width=0.31\linewidth]{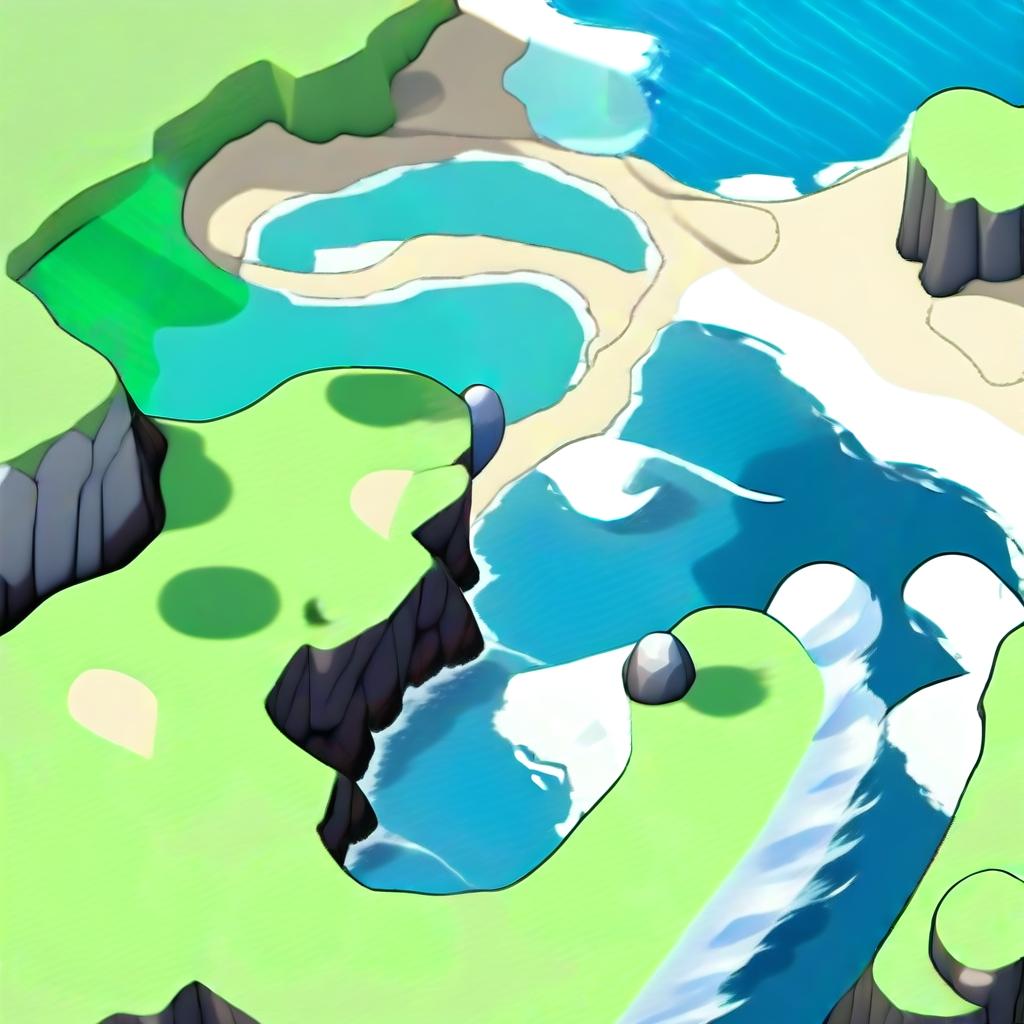} 
    \end{subfigure}  

    \caption{Basemap inpainting results of SDXL-Inpaint (middle) and ours (right) on the isometric images (left)}
    \label{fig:inpaint_on_control_out2}
\end{figure}

\section{Results}
\subsection{Training and Inference Details.}  We used AdamW optimizer with a learning rate of 1e-5 for training/fine-tuning both the ControlNet and inpainting models.  The pre-trained diffusion models adopted in our experiments were SDXL-base model\cite{podell2023sdxl} for ControlNet, and  SDXL-Inpaint model \cite{sdxlinpaint} for inpainting.  We set the rank parameter of 64 in LoRA for inpainting. The ControlNet was fine-tuned on a single NVIDIA A100 GPU, completing 50K steps in around 10 hours.  The inpainting model was trained on 4x V100 GPUs for 100K steps in about 60 hours. The total inference time of the entire pipeline is about 3 minutes using a single V100 GPU.

We collected datasets to train both the ControlNet and the Inpainting model respectively. The ControlNet dataset comprises 10,000 isometric view game scene images generated by SDXL~\cite{podell2023sdxl}, paired with corresponding text prompts from InstructBlip~\cite{dai2024instructblip} and associated sketches. These sketches were generated by combining results obtained by several StoA foudnation models, inlcuding Grounding DINO~\cite{liu2023grounding}, Segment Anything~\cite{sam}, and Osprey~\cite{Osprey}.
Since we did not have any isometric basemap as the ground truth, we curated an inpainting dataset from three sources: 5,000 isometric images with foreground objects, 4,000 manually filtered perspective images of empty terrains inpainted using ~\cite{LaMa2021}, and 1000 pure texture images.

\begin{figure}[ht!]
    \centering
    \includegraphics[width=0.35\linewidth]{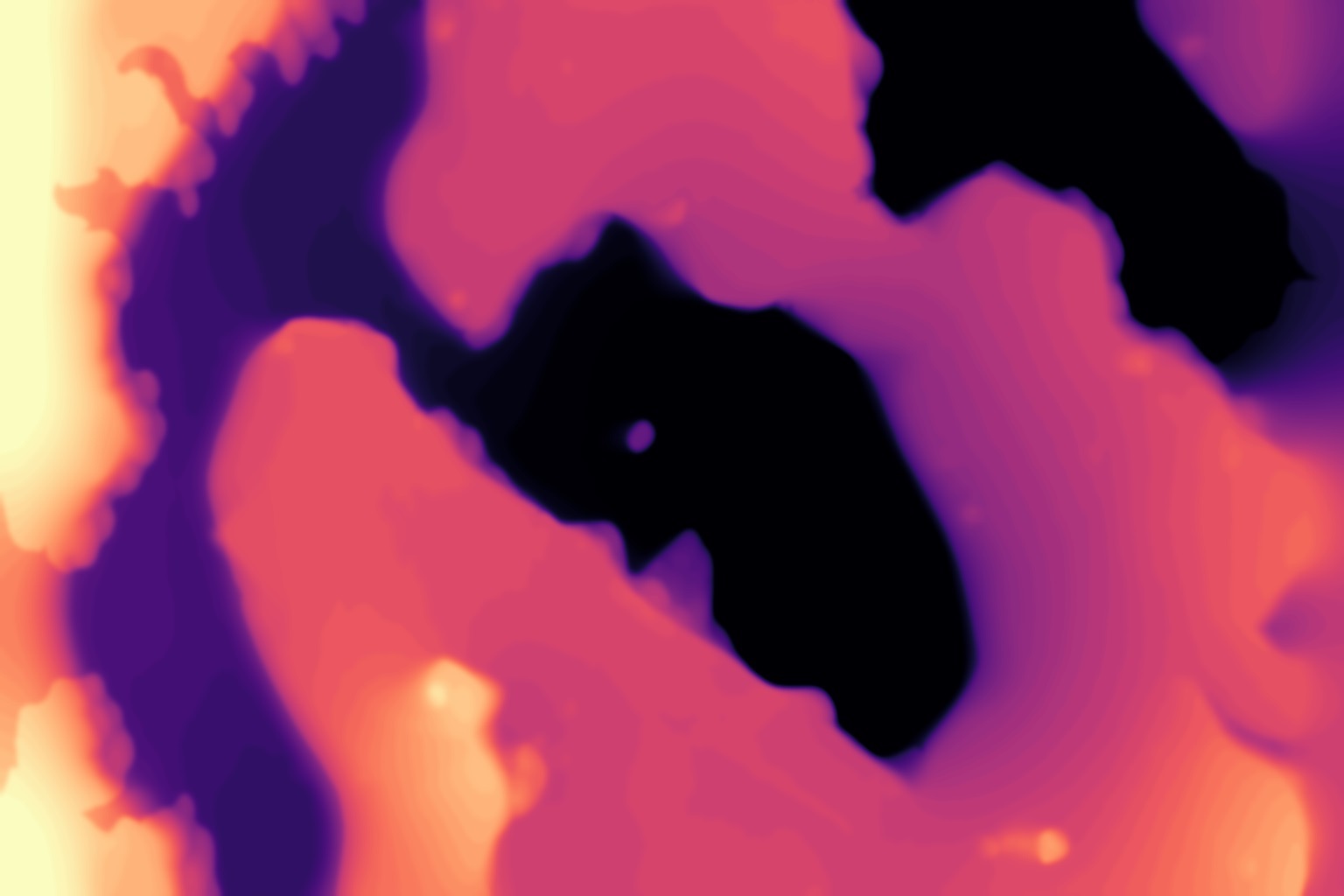}
    \includegraphics[width=0.35\linewidth]{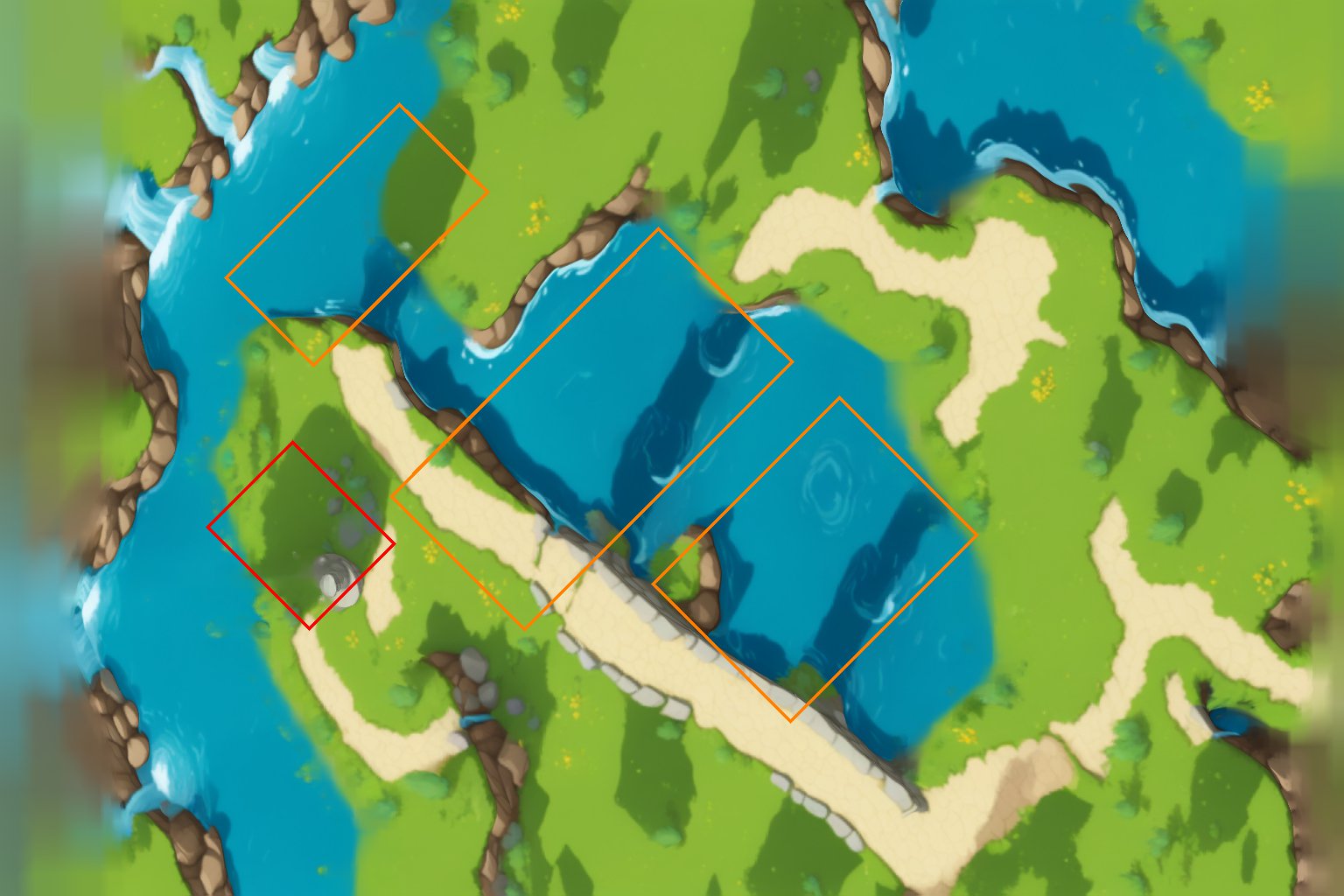}
    \includegraphics[width=0.25\linewidth]{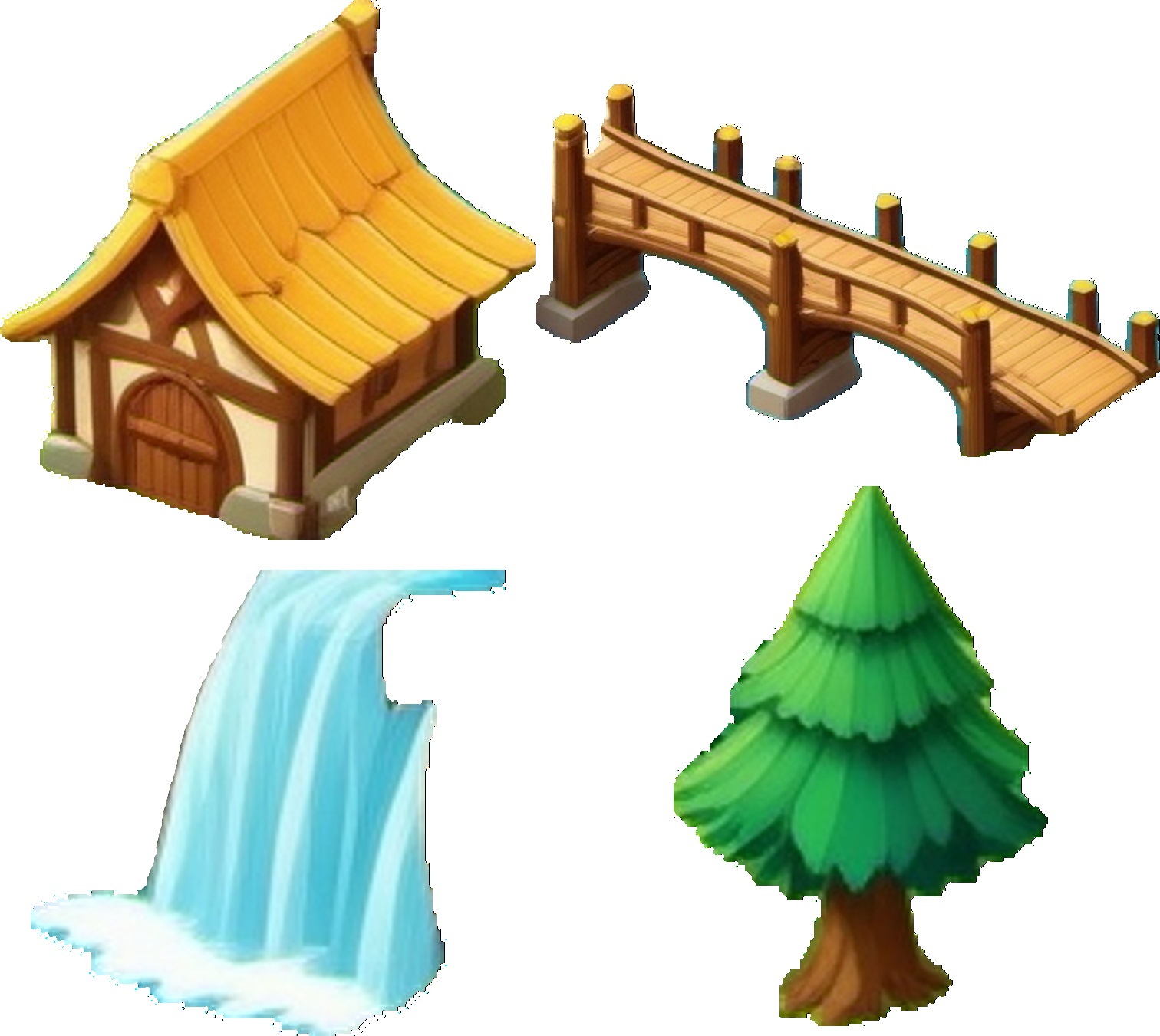}
    \includegraphics[width=0.35\linewidth]{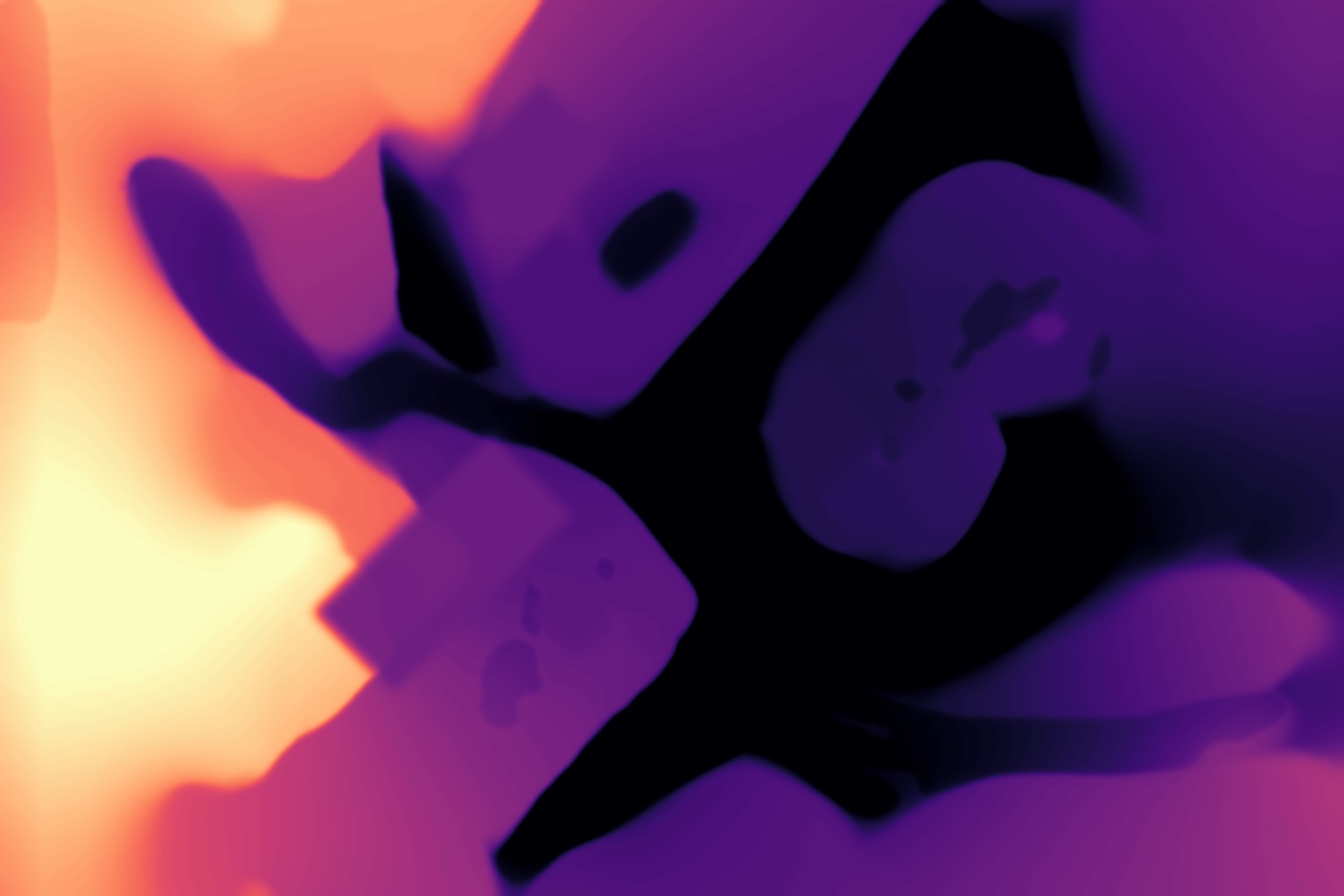}
    \includegraphics[width=0.35\linewidth]{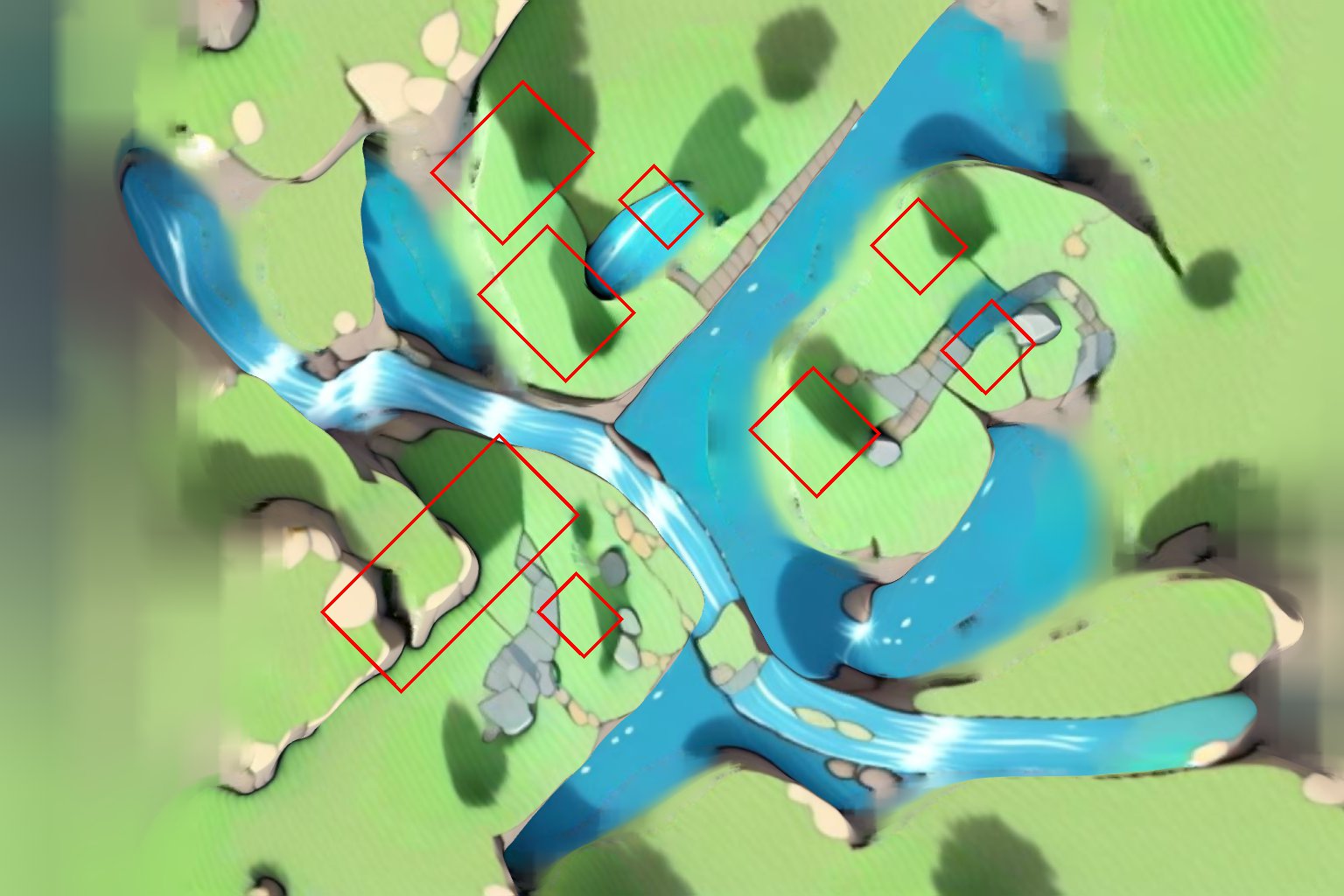}
    \includegraphics[width=0.255\linewidth]{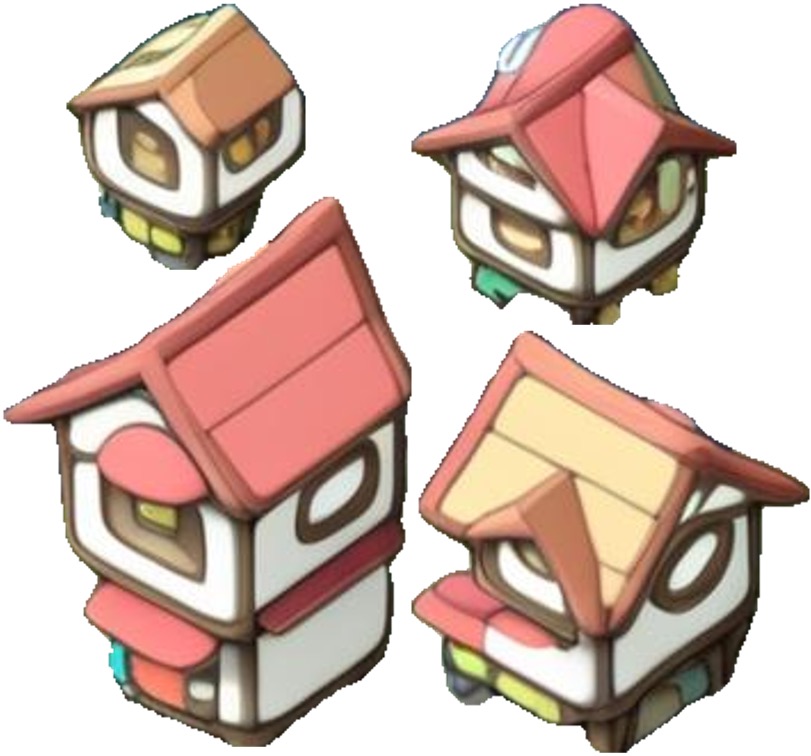}
    \includegraphics[width=0.35\linewidth]{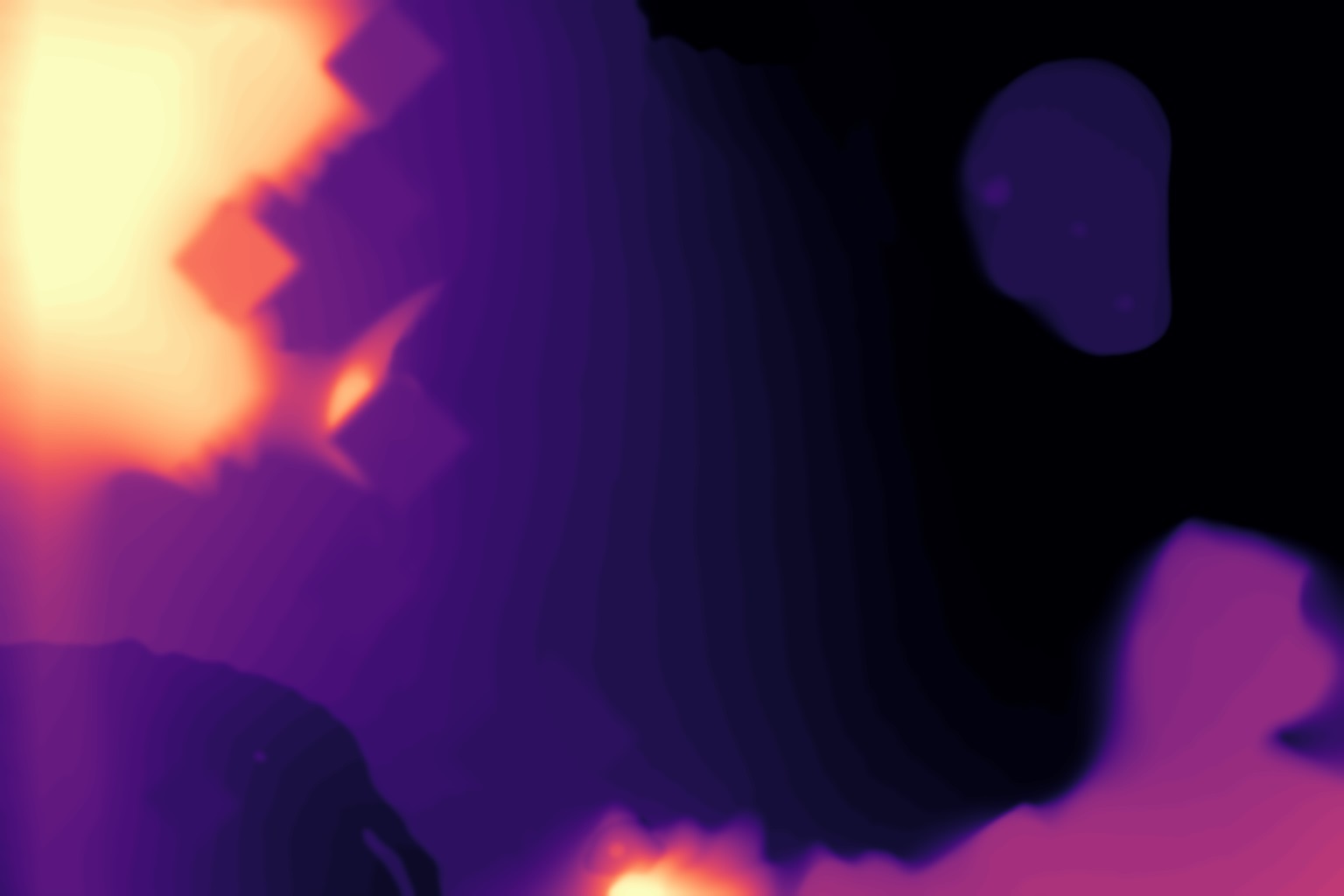}
    \includegraphics[width=0.35\linewidth]{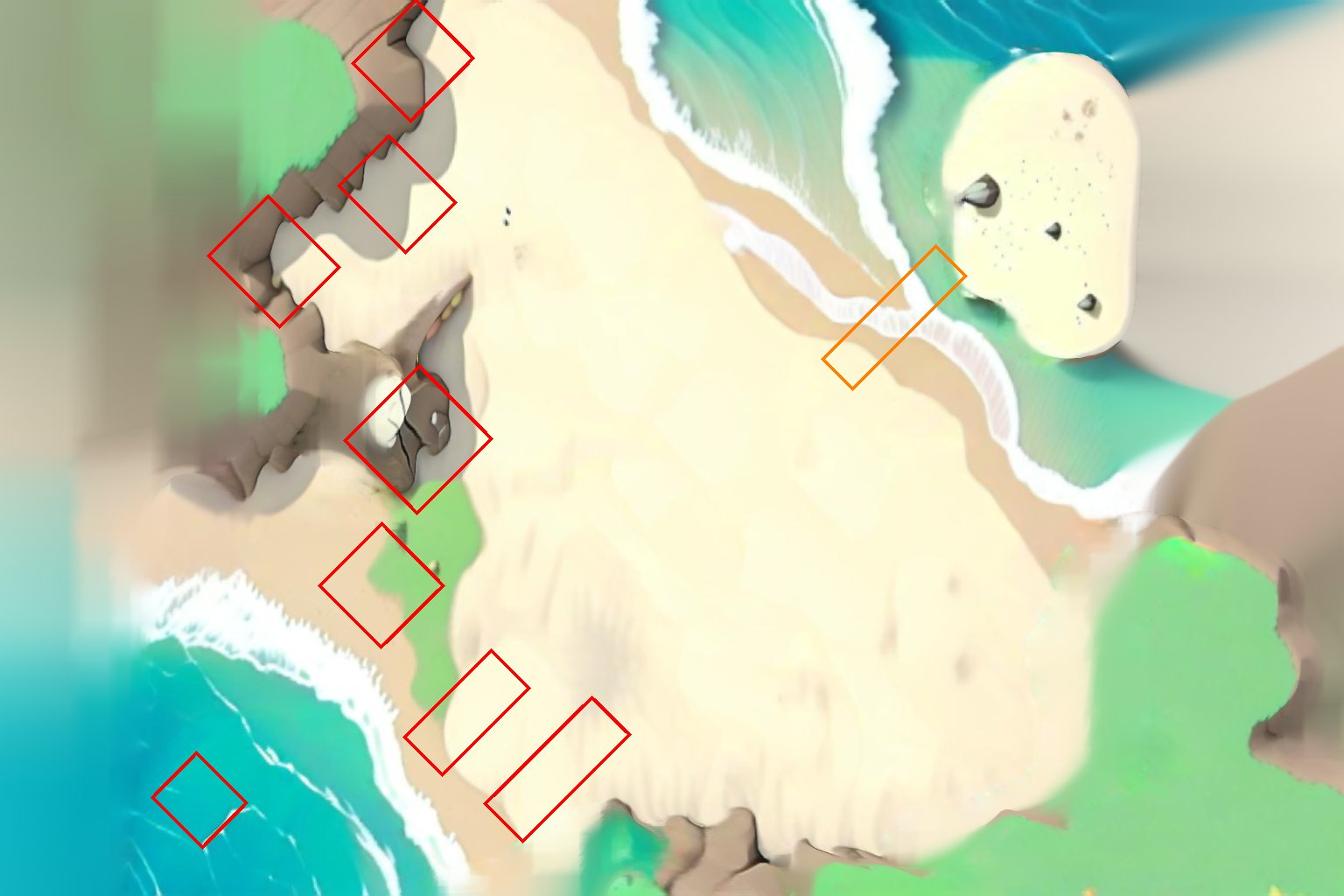}
    \includegraphics[width=0.25\linewidth]{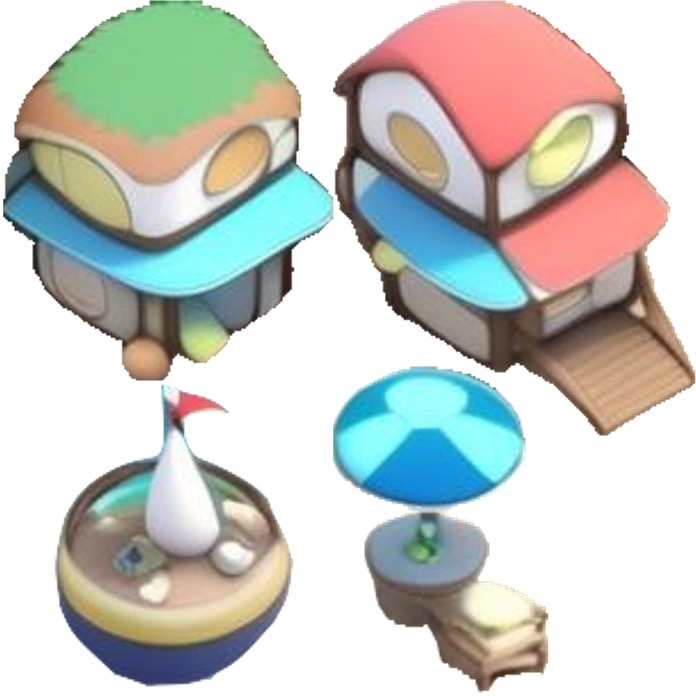}
    \caption{Scene understanding results showing heightmap, object placement bounding boxes and object reference images for Fig. 1, \ref{subfig:sketch1} and \ref{subfig:sketch4}. }
    \label{fig:example_scene_understanding}
\end{figure}

\subsection{Isometric 2D Image Generation.}
\label{sec:2d_result}
Fig. \ref{fig:inpaint_on_control_out} shows representative results using our ControlNet and inpainting models with a diverse set of user sketches and prompts. These results demonstrate ControlNet's ability to accurately follow sketch layouts and apply the scene style dictated by the prompt.  The inpainting model generates clean basemaps that consistently align with the full isometric images, even when the foreground masks cover a significant portion of the image.

As shown in Fig. \ref{fig:inpaint_on_control_out}, the ControlNet offers flexibility to the user's sketch, accommodating various categories like water-only in Fig.~\ref{subfig:sketch1} and \ref{subfig:sketch2}, and three categories in Fig.~\ref{subfig:sketch4} respectively. With the same sketch, Fig.~\ref{subfig:sketch1} and \ref{subfig:sketch2} produce distinct scenes by applying different styles from the texts. 

How to balance the influences of the sketch condition and the textual prompt guidance is the key.  Our SAL-enhanced ControlNet simplifies this balancing process by allowing casual (not precise) user sketches, occasionally adding extra objects or expanding patch areas to implement the user's design intention.  For example, in Fig.~\ref{subfig:sketch2}, the river and waterfall blend coherently to meet both text and sketch requirements.
In Fig.~\ref{subfig:sketch4}, eight buildings are added to match the phrase ``town with many buildings" while respecting the original user-drawn sketch.

\begin{figure*}[hb]
    \centering
    \includegraphics[width=1.0\linewidth]{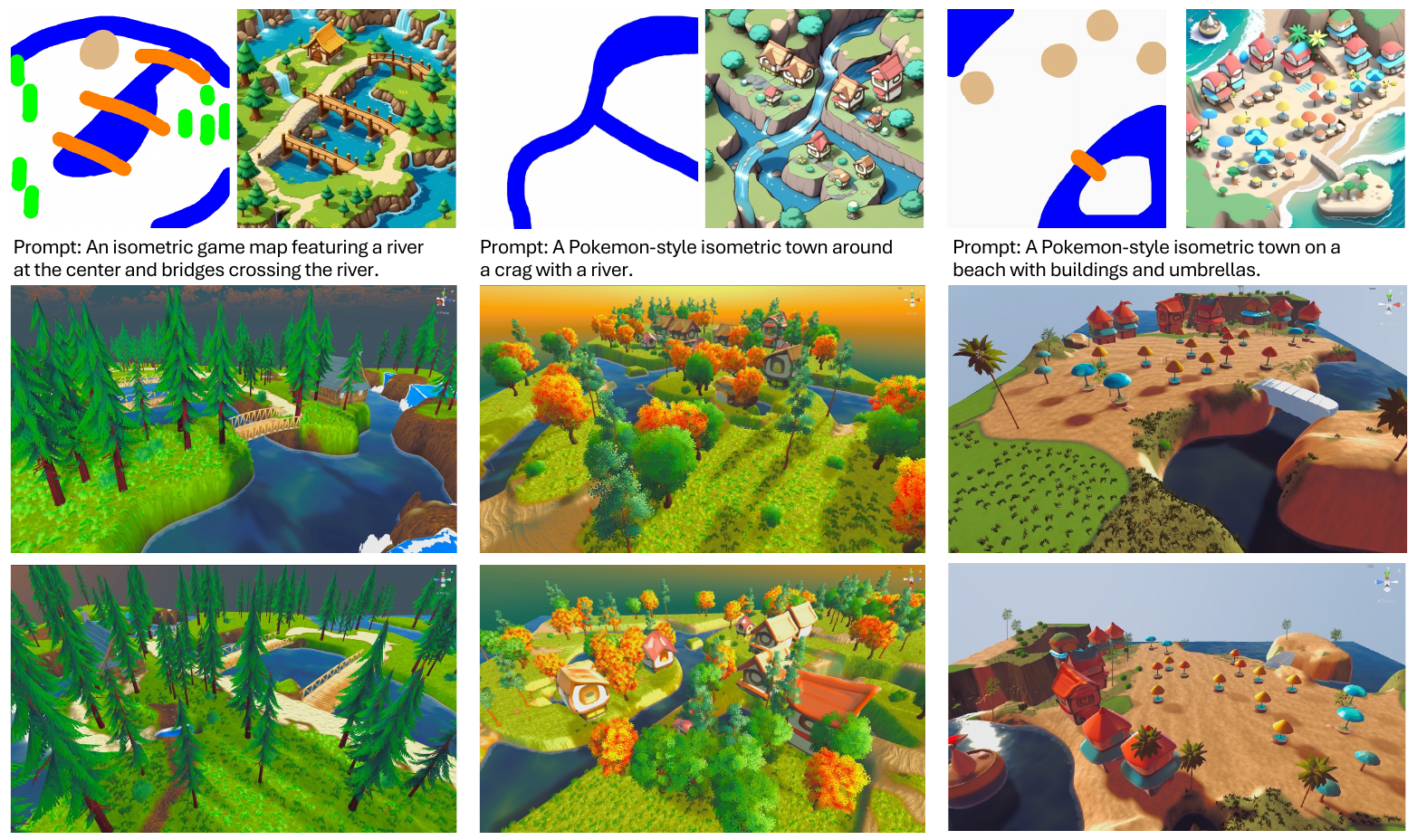}
    \caption{ {\bf More example 3D scene generation results.} From left to right: Different views of the generated 3D scenes from the isometric images of Fig. 1, Fig.~\ref{subfig:sketch1} and \ref{subfig:sketch4}. }
    \label{fig:example_scenes}
\end{figure*}


\subsection{Inpainting Comparisons}
\label{sec:inpaint_result}
We compare inpainting results with SDXL-Inpaint~\cite{sdxlinpaint} on the isometric images in Fig.~\ref{fig:inpaint_on_control_out2}.  
We use a positive prompt of ``{\textit{an empty terrain map with nothing rising above the surface. This is a landscape without any buildings, vegetation, or bridges.}}" and a negative prompt of {\textit{``buildings, vegetation, trees, bridges, artifacts, low-quality"}}.  
Our model successfully produced clean and consistent basemaps, whereas SDXL-Inpaint tended to substitute buildings and trees with artifacts.

\subsection{Visual Scene Understanding}
Given the 2D isometric and empty basemap, our visual scene understanding module recovers the instance-level semantic segmentation of the foreground objects, estimates the isometric depth, recovers the rough terrain mesh, renders the BEV heightmap and color image, segments the splatmap and recovers the foreground object placement. Figure~\ref{fig:example_scene_understanding} shows examples of the generated heightmap, BEV object placement and the extracted object reference images.

\subsection{Procedural 3D Scene Generation}
\label{sec:3d_result}
Figure~\ref{fig:example_scenes} shows three 3D scenes generated from the isometric images from Fig. 1, ~\ref{subfig:sketch1} and \ref{subfig:sketch4}. It show that the layout and texture style of the 3D scenes are well aligned with the associated sketch and isometric image. 

\begin{figure*}
\begin{center}   

\includegraphics[width=1.0\linewidth]{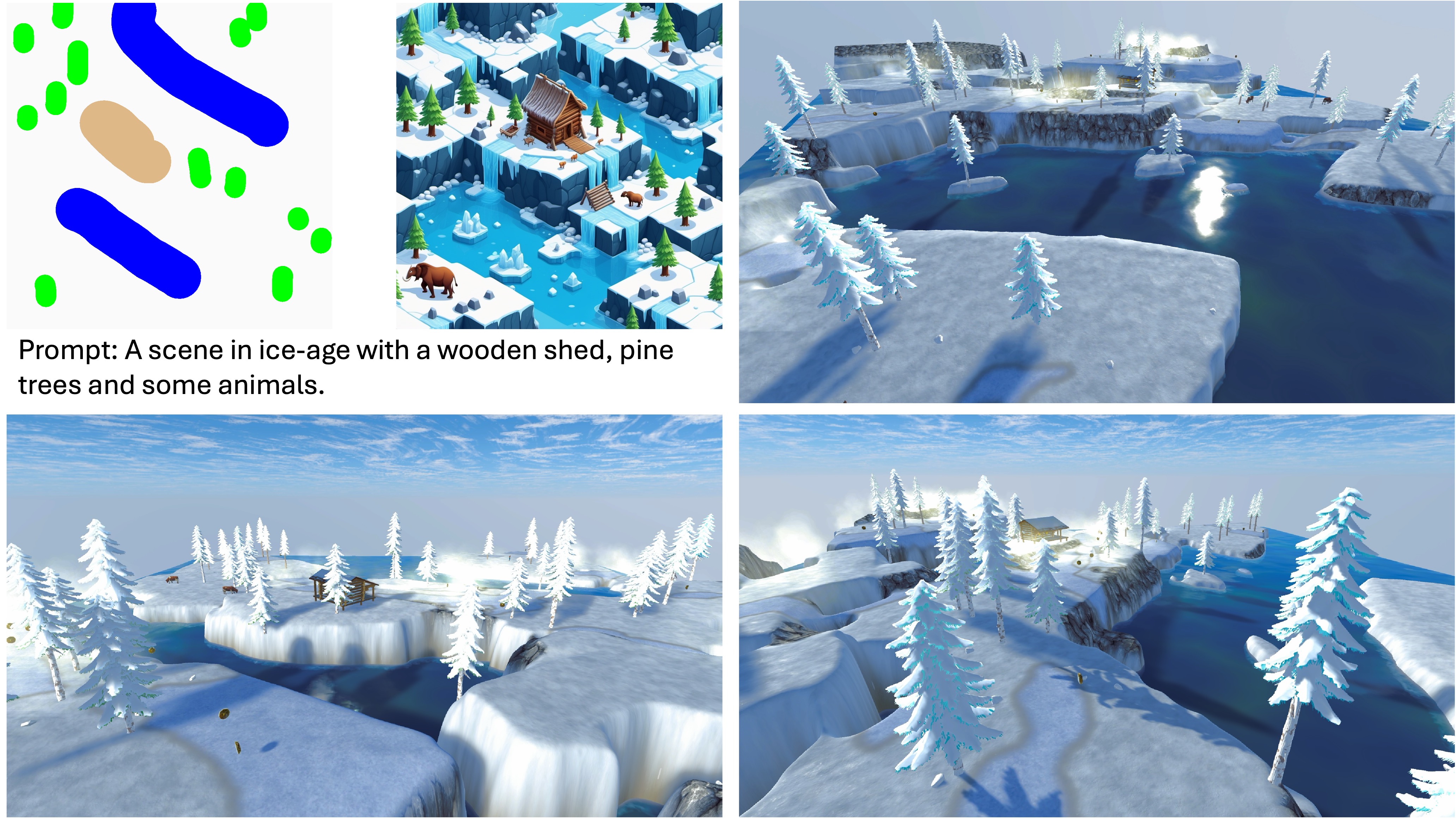}
    \caption{An additional example of 3D scene generation from text and sketch. The top left displays the input sketch and text, along with the generated isometric image. The other three images are rendered from
    the 3D scene with different viewpoints. The prompt is: ``A scene in ice-age with a wooden shed, pine trees and some animals."}
    \label{fig:scene_3D_ice}
\end{center}
\end{figure*}

\begin{figure*}
  \centering
  \includegraphics[width=1.0\linewidth]{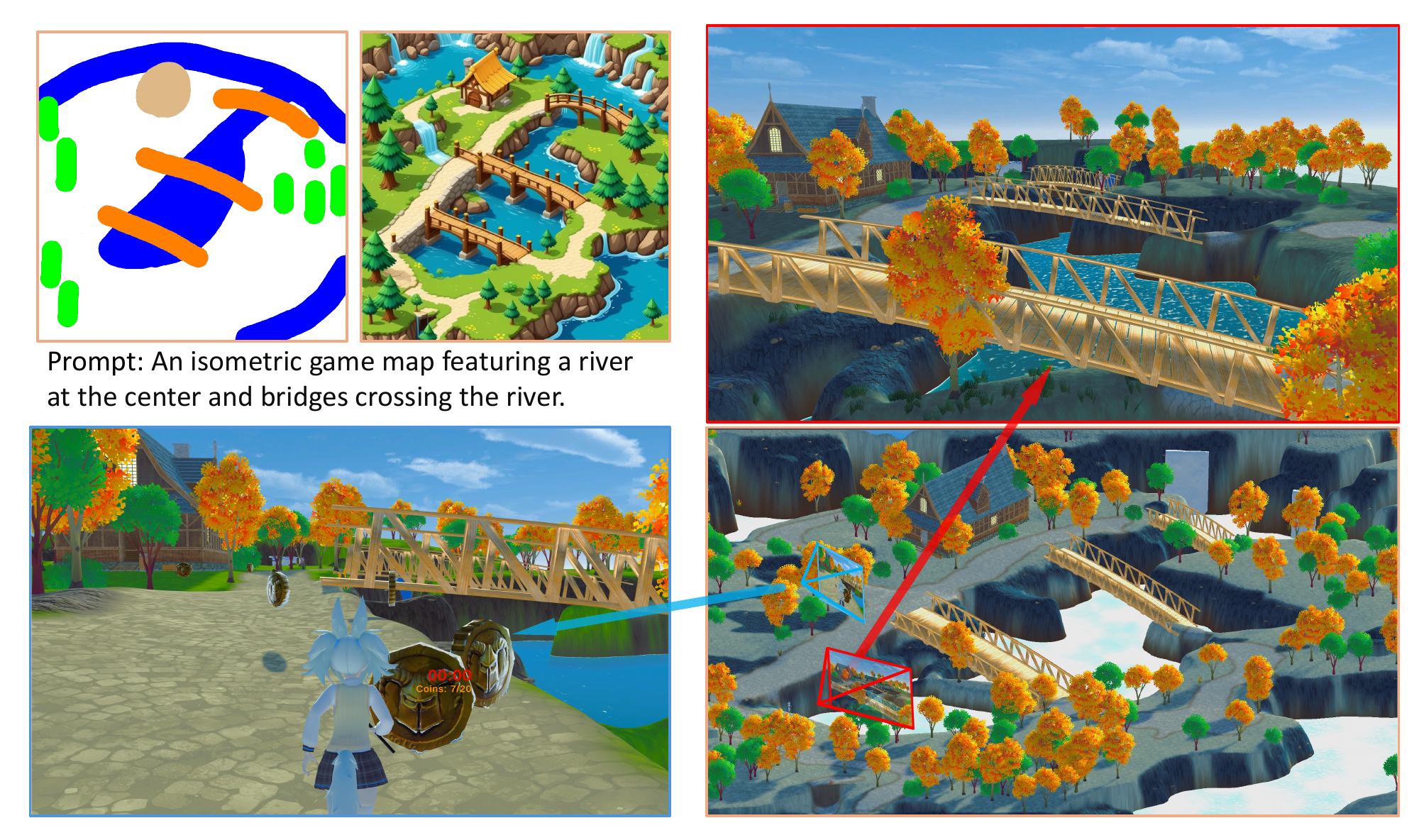} 
  \caption{ {\bf 3D Scene Editing}:  By varying the parameters of the 3D assets, e.g. the type and color of the trees, our method can facilitate content-editing and style transfer for the same 3D game scene. }
  \label{fig:scene-editing}
\end{figure*}

The objects in the first scene is retrieved from the Objaverse, while objects in the second and third scenes are generated by ~\cite{hyperhuman2024} using the object instance images extracted from the isometric image. These objects not only harmonize with the scene's texture style but are also automatically and accurately scaled, oriented, and positioned in the 3D scene according to the BEV footprint.  Note that variations in material composition and lighting dynamics have led to a slight discrepancy in color between the rendered images of 3D scenes and the reference image. More example results are shown in Fig.~\ref{fig:scene_3D_ice} and~\ref{fig:scene-editing}.  

\subsection{Limitations}
Our current implementation adopts a multi-stage pipeline involving many intermediate stages. Errors can be easily accumulated, which sometimes requires the user to restart from a different noise seed. One potential remedy is to concurrently generate multiple modalities at the same time, such as RGB, semantic, depth, surface material, and object footprint, and fuse these intermediate results until a coherent final result is obtained. Concurrently generating foreground and background layers is also a possible solution, by for exmaple applying the newly proposed LayerDiffusion method ~\cite{zhang2024transparent} . Currently, in our pipeline, terrain texture and terrain materials are obtained solely by retrieving a terrain database,  which limits the diversity of terrain textures. In the future, we plan to develop diffusion based texture-generation models similar to \cite{wang2024textureinfinite,texturedreamer}.

\section{Conclusion}  
We have proposed a novel approach called Sketch2Scene for generating 3D interactive scenes from users' casual sketches and text prompts. To address the main challenge of insufficient large-scale training data for 3D scenes, we leverage and improve pre-trained large-scale 2D diffusion models for the task. We provides two innovations to  existing diffusion models: (1) SAL-enhanced ControlNet, and (2) step-unrolled diffusion inpainting.  In contrast to other recent generative techniques for 3D scene generation (e.g., using SDS loss \cite{hollein2023text2room}, or direct triplane regression \cite{wu2024blockfusion}), our approach generates high quality and interactive 3D scenes with vivid 3D assets that can be seamlessly integrated into existing game engines, ready for many downstream applications.  We also discussed limitations and possible remedies in the paper. The reader is invited to watch our companion video on our project page (https://xrvisionlabs.github.io/Sketch2Scene/). 

{ \small 
\bibliographystyle{IEEEtran}
\bibliography{ref}
}


\clearpage 
\section*{Appendix}
\subsection{Dataset}
\paragraph{ControlNet-Dataset} While several successful text/sketch-to-image works have already been presented, none of them focus specifically on isometric view game scenes. Since collecting a large number of isometric view game scene images for training is challenging, we created a dataset by generating these images using the SDXL model. The dataset is designed to validate the effectiveness of our method and reduce the domain gap with the original model. 

We first used text prompts as input and employed the SDXL model to generate 10,000 isometric view game scene images. To label these images for training, we utilize Grounding DINO and Segment Anything to detect and segment semantic masks for elements such as buildings, trees, and boats. Additionally, we used Segment Anything along with Osprey to generate masks for irregularly shaped semantic elements such as water bodies, bridges, and roads. We manually annotated road element masks in 2,000 images for better accuracy. All images were then captioned using InstructBlip~\cite{dai2024instructblip} to obtain detailed text prompts.

\begin{figure}[h]
  \centering
  \includegraphics[width=0.95\linewidth]{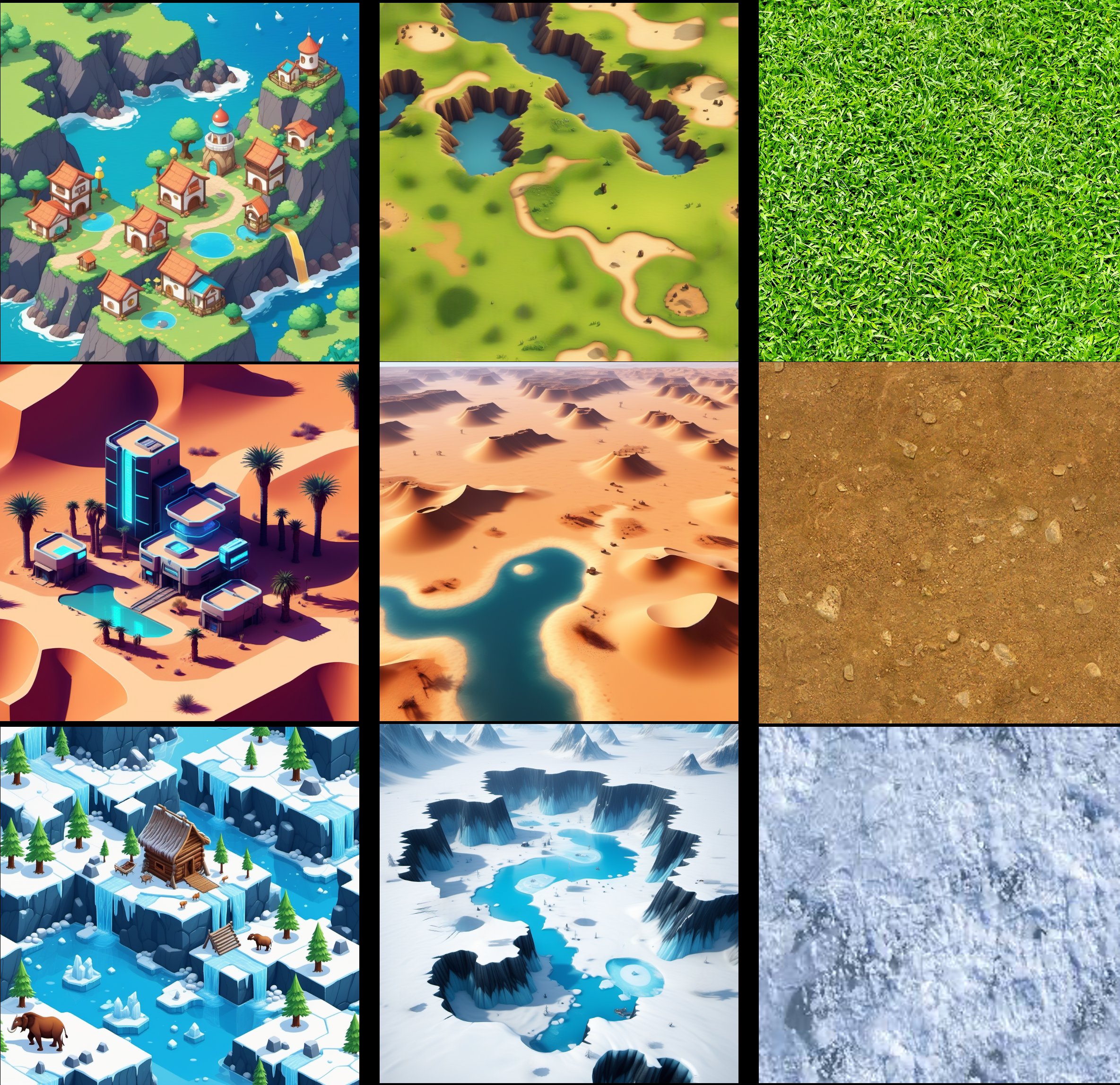}  
  \caption{Examples of inpainting training data. From left to right columns: full isometric, inpainted from perspective semi-empty images, pure texture maps.} 
  \label{fig:inpaint_data}
\end{figure}

\begin{figure}[h]
  \centering  
    \begin{subfigure}{1\linewidth}
        \includegraphics[width=0.31\linewidth]{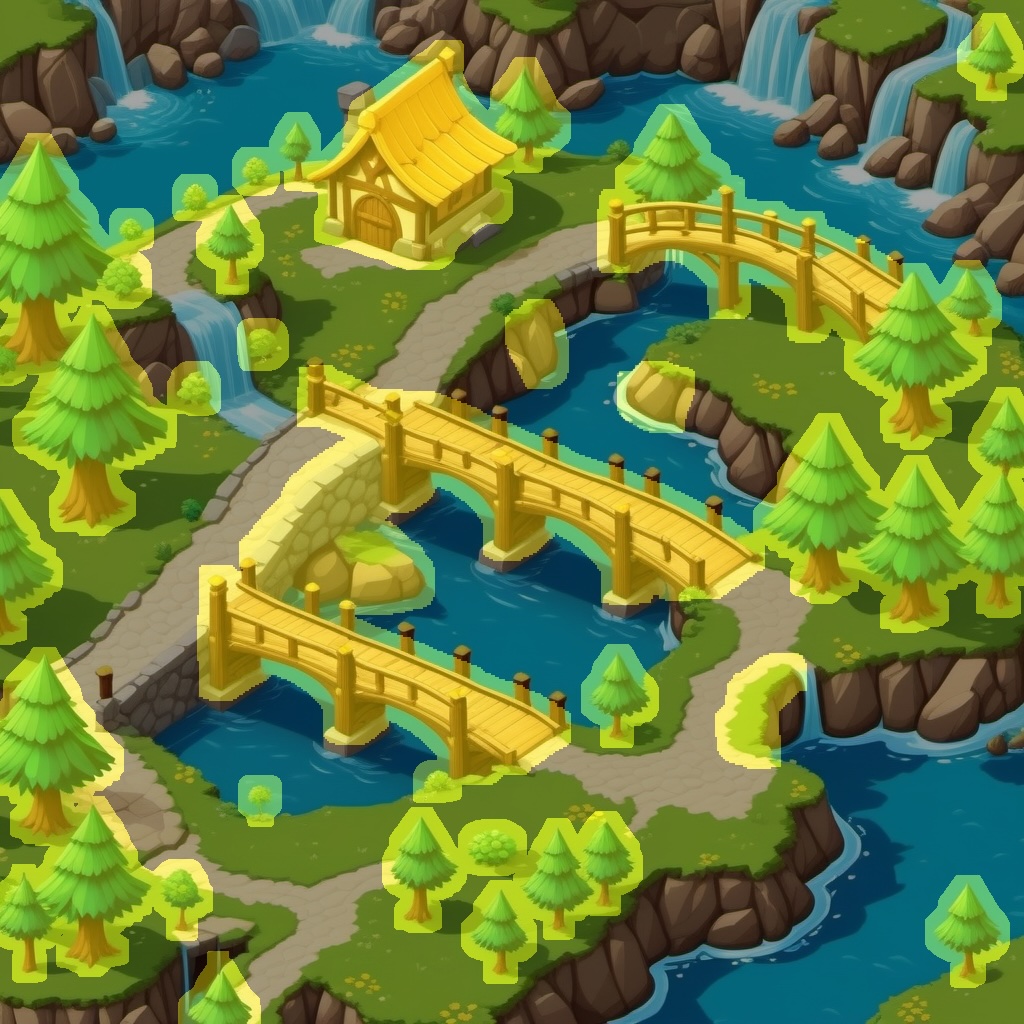} 
        \includegraphics[width=0.31\linewidth]{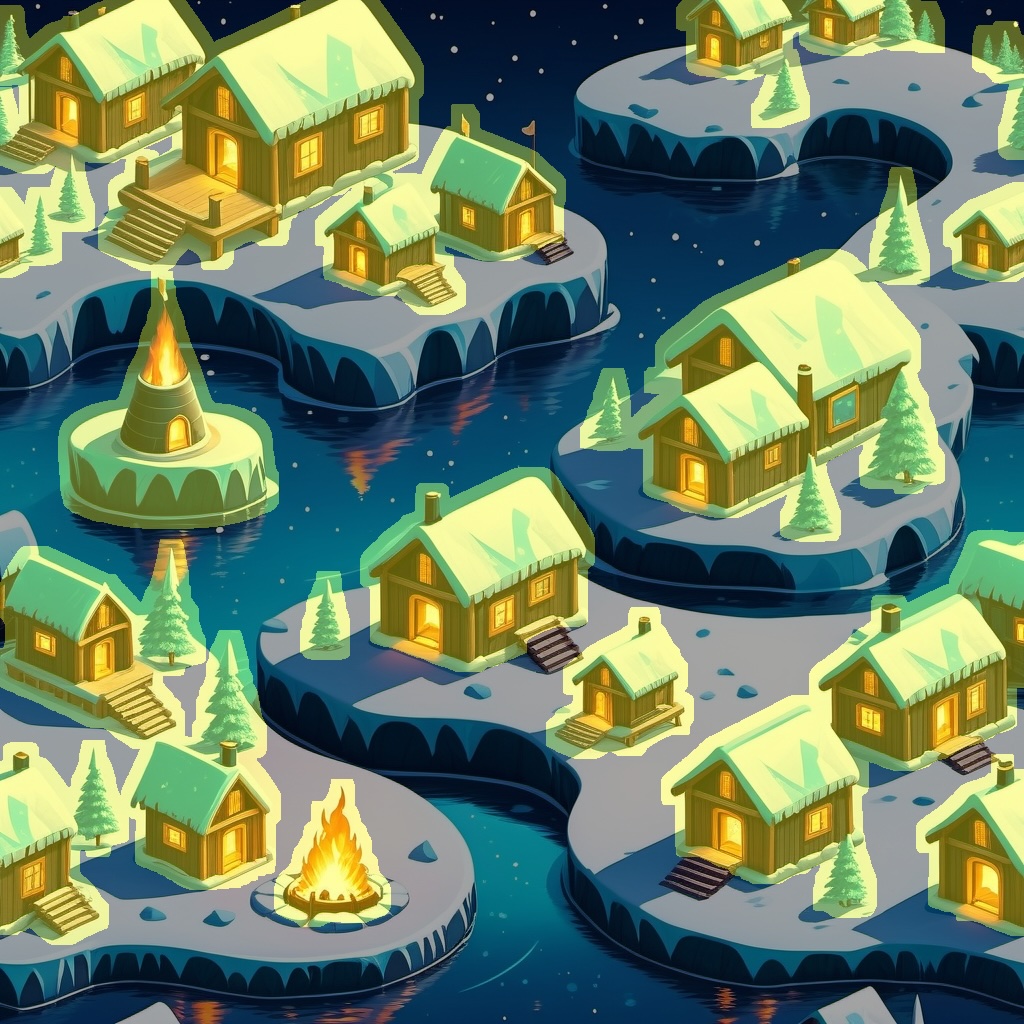} 
        \includegraphics[width=0.31\linewidth]{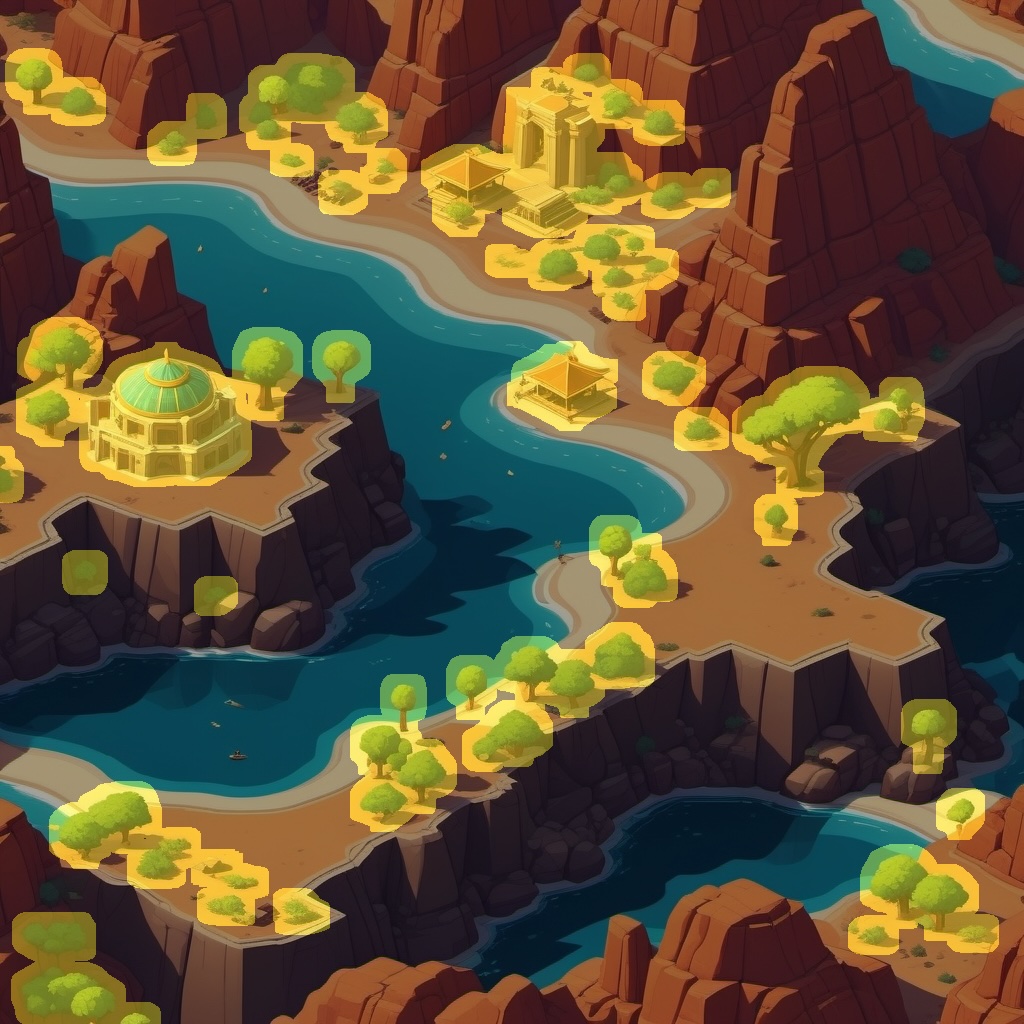} 
    \end{subfigure}
  \caption{Examples of inpainting inference masks, where the mask regions are foreground objects.}
  \label{fig:inp_inf_mask}
\end{figure}

\paragraph{Inpainting-Dataset} 
Ideal training data for our inpainting model would be large-scale, pure isometric basemap images, but collecting or generating these is challenging. We found a viable alternative by combining three types of readily available data as mentioned in the main paper. Figure~\ref{fig:inpaint_data} shows supplementary examples of this data. The masks during the inference phase of inpainting are foreground masks of full isometric images. as illustrated in Fig.~\ref{fig:inp_inf_mask}. Supplementary examples of masks in the training of inpainting are displayed in ~\ref{fig:inpaint_data}.

\subsection{Examples of inpainting training data}
\begin{figure}[h]
  \centering 
    \begin{subfigure}{1\linewidth}
    \includegraphics[width=0.31\linewidth]{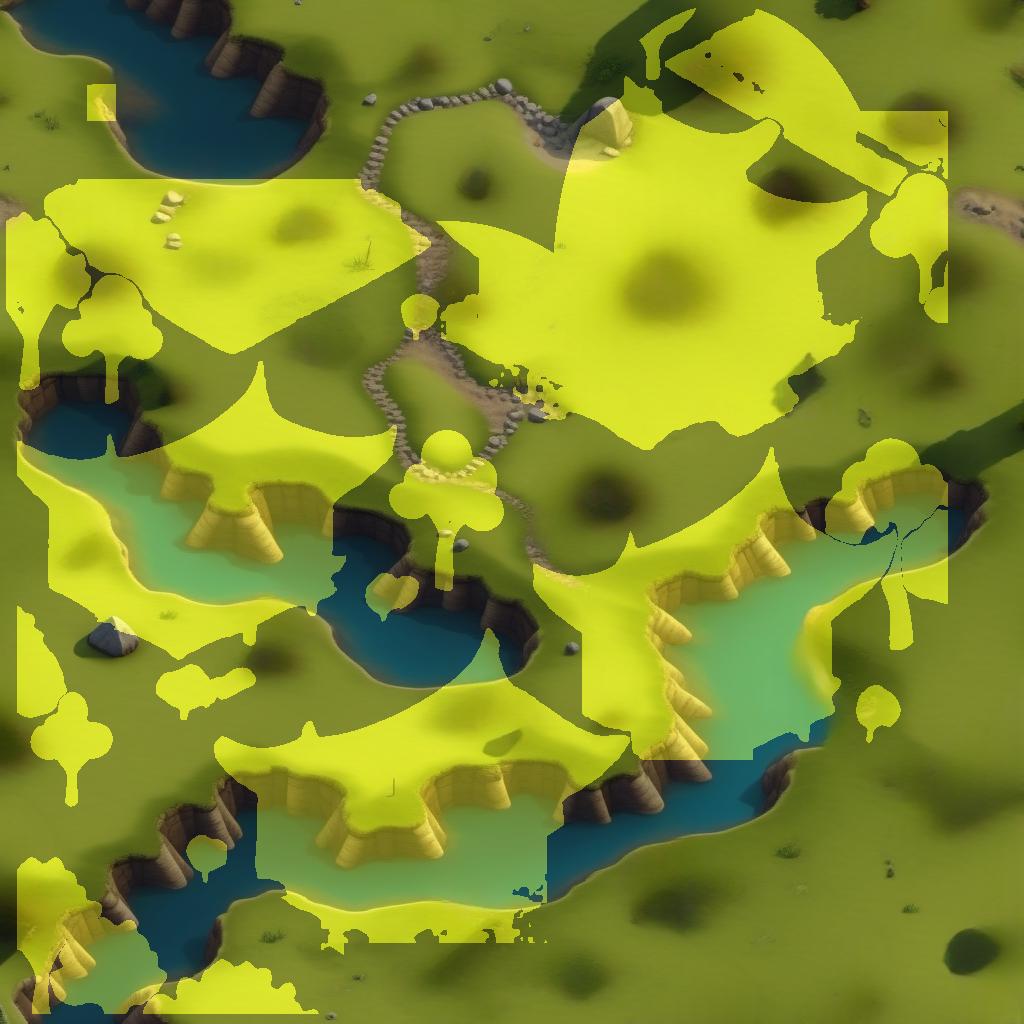} 
    \includegraphics[width=0.31\linewidth]{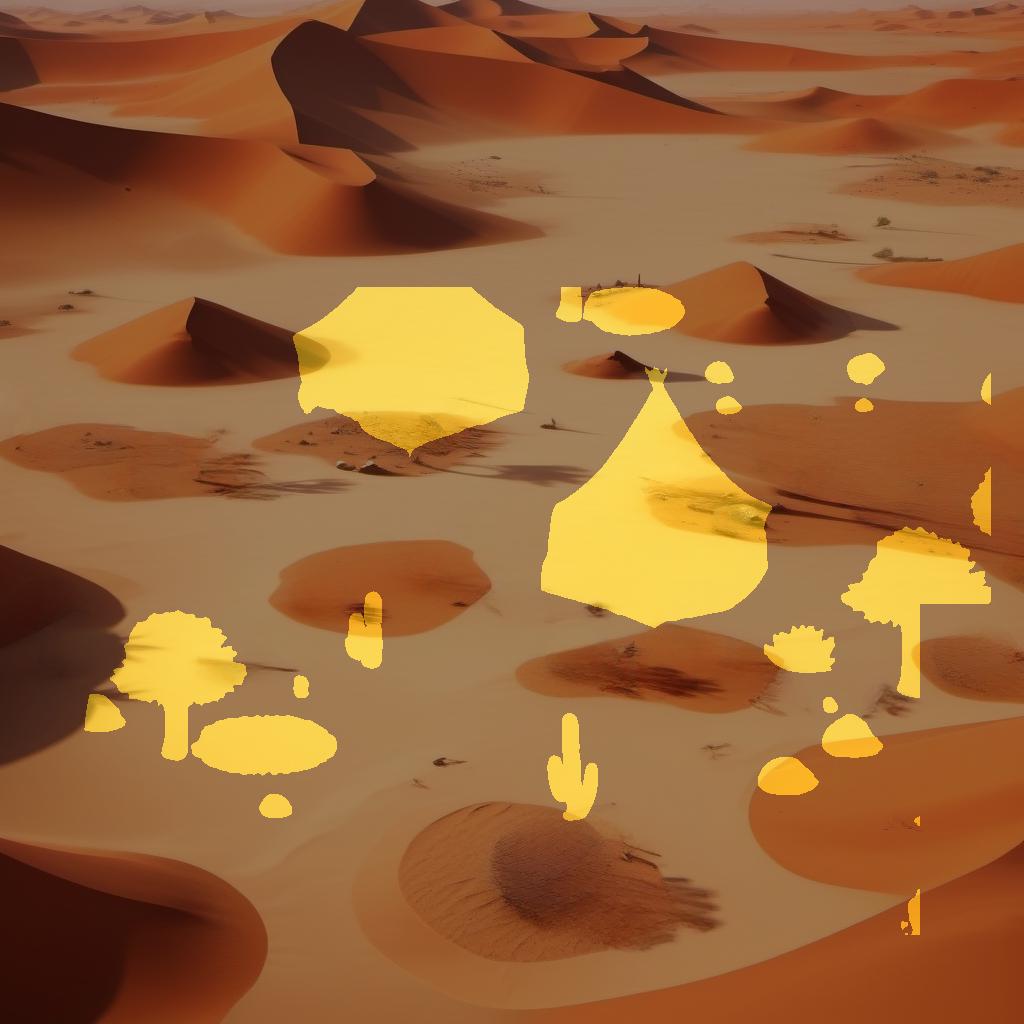}
    \includegraphics[width=0.31\linewidth]{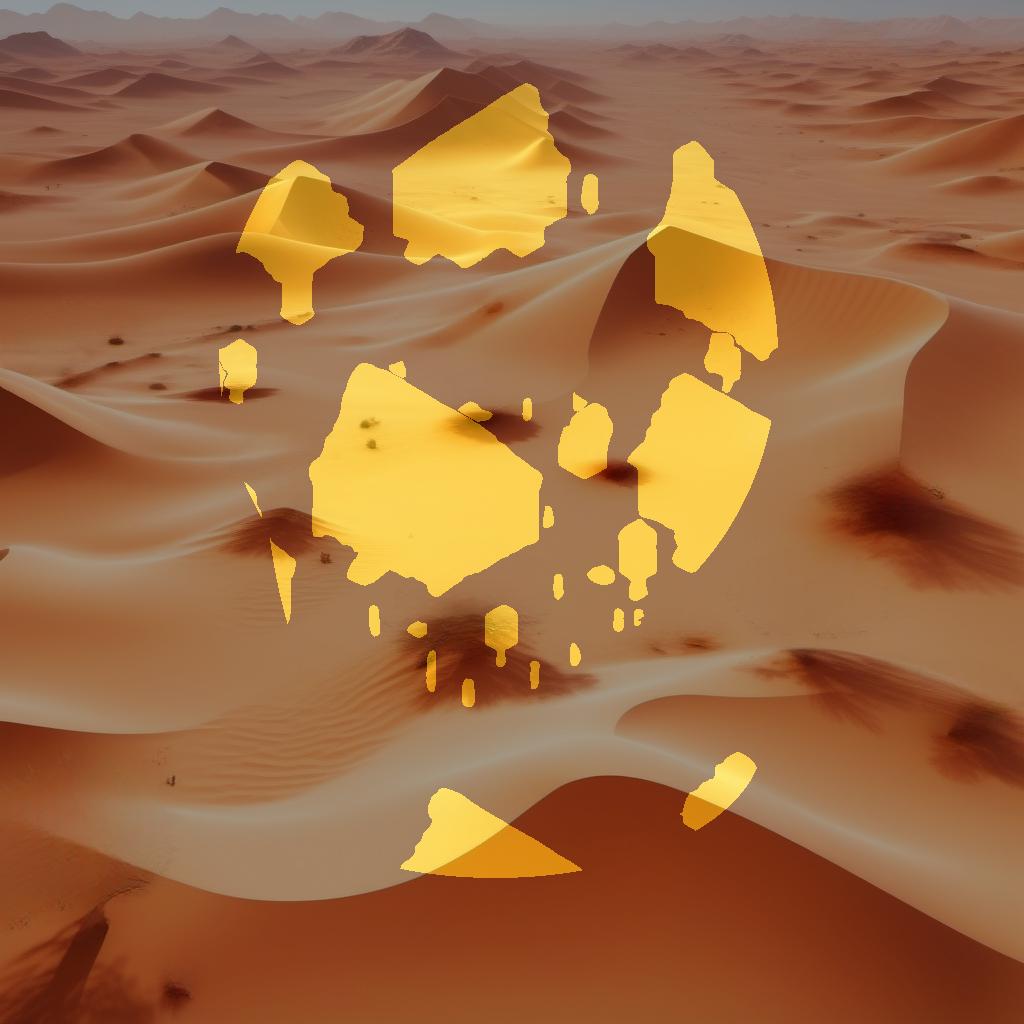}
    \end{subfigure}

    \begin{subfigure}{1\linewidth}
    \includegraphics[width=0.31\linewidth]{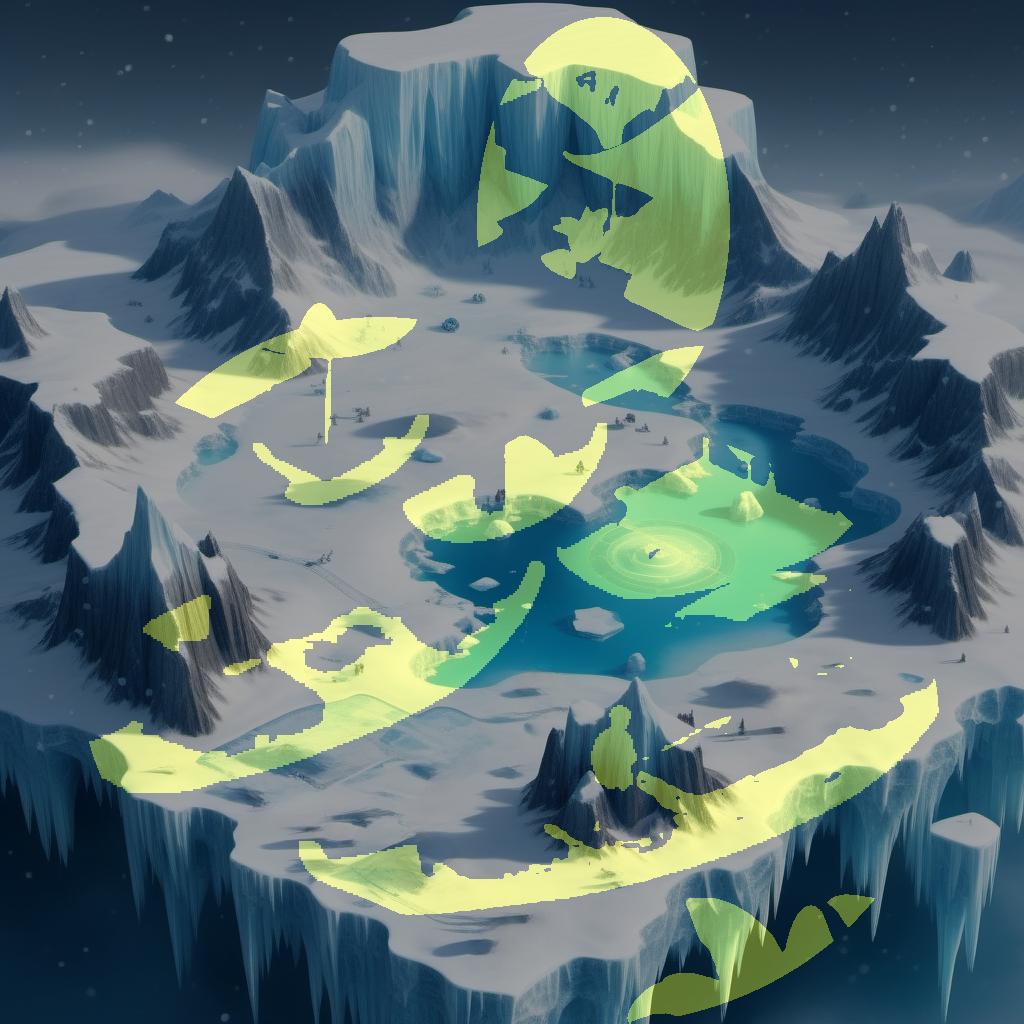} 
    \includegraphics[width=0.31\linewidth]{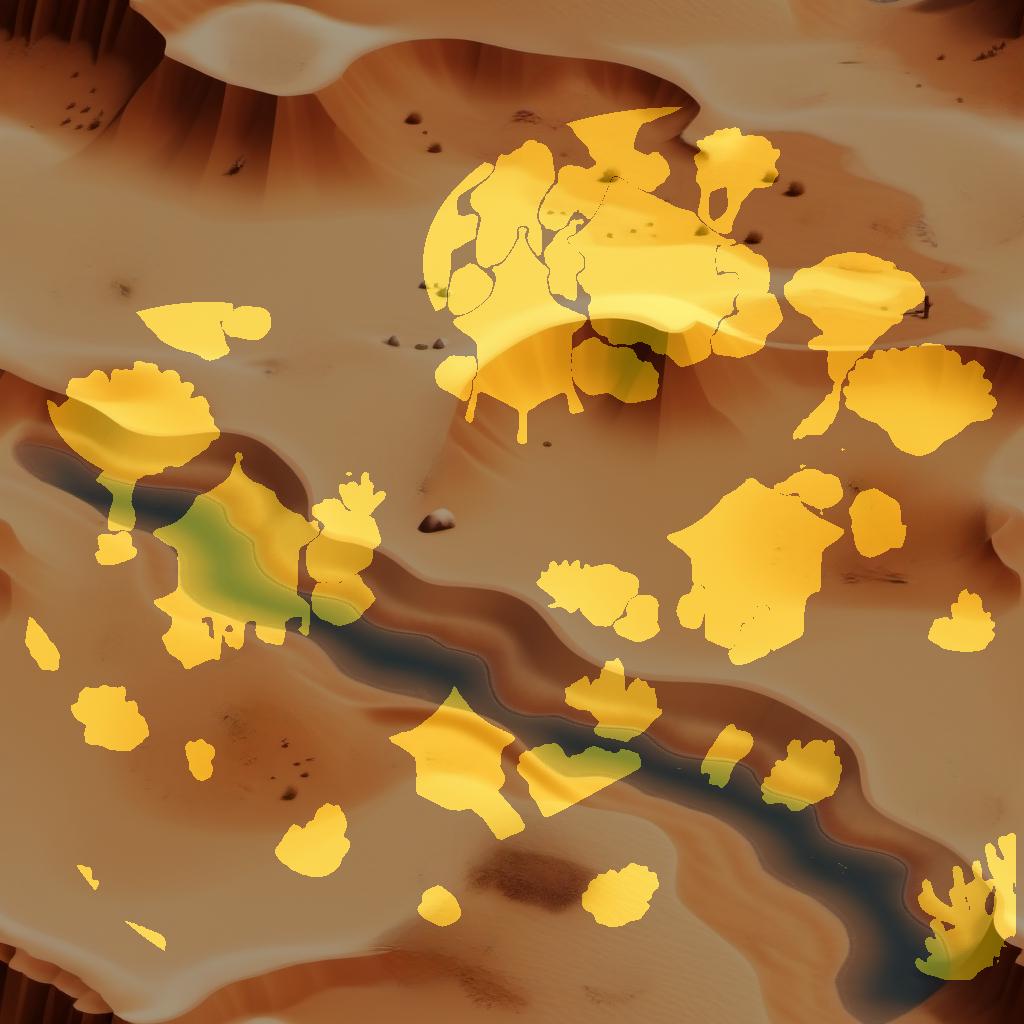}
    \includegraphics[width=0.31\linewidth]{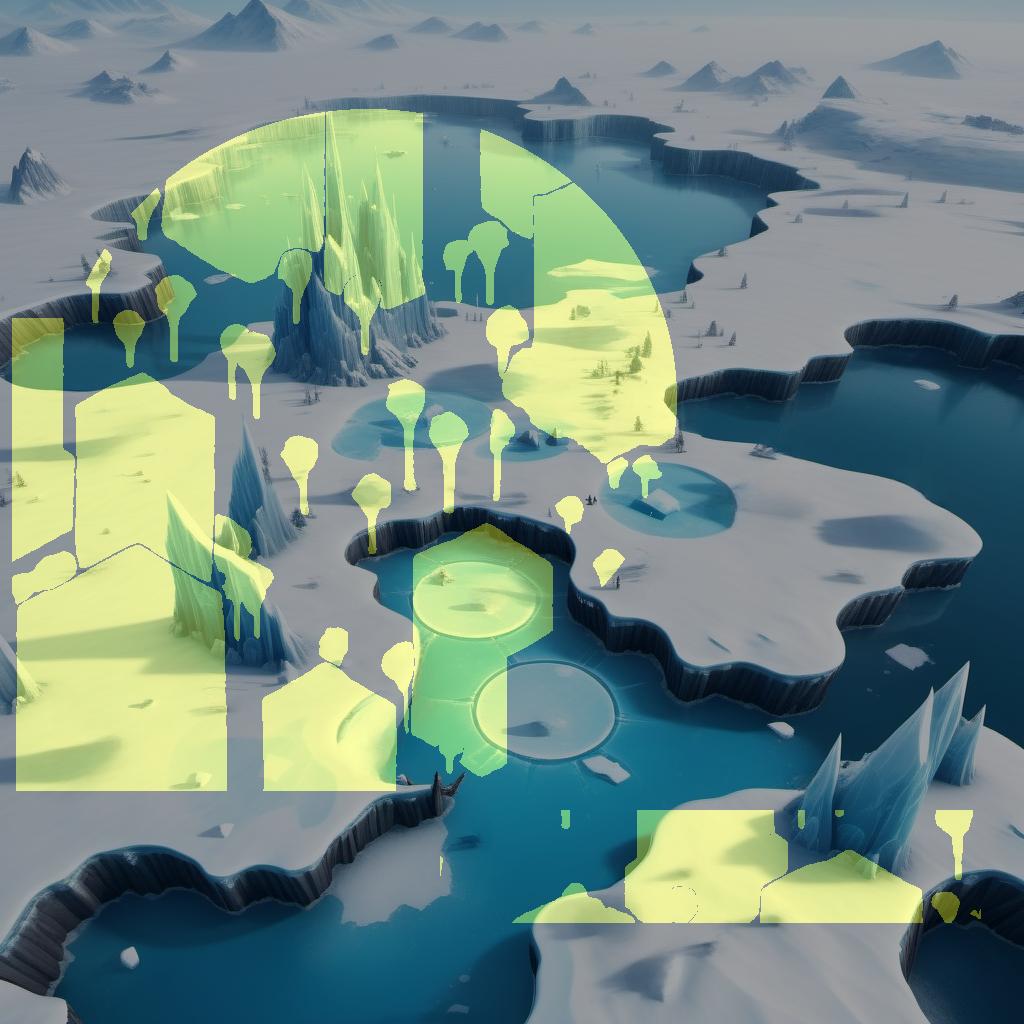}
    \end{subfigure}
  \caption{Examples of inpainting training data: empty map. }
  \label{fig:sup_inpaint_train_data_empty}
\end{figure}

\begin{figure}[h]
  \centering 
    \begin{subfigure}{1\linewidth}
    \includegraphics[width=0.31\linewidth]{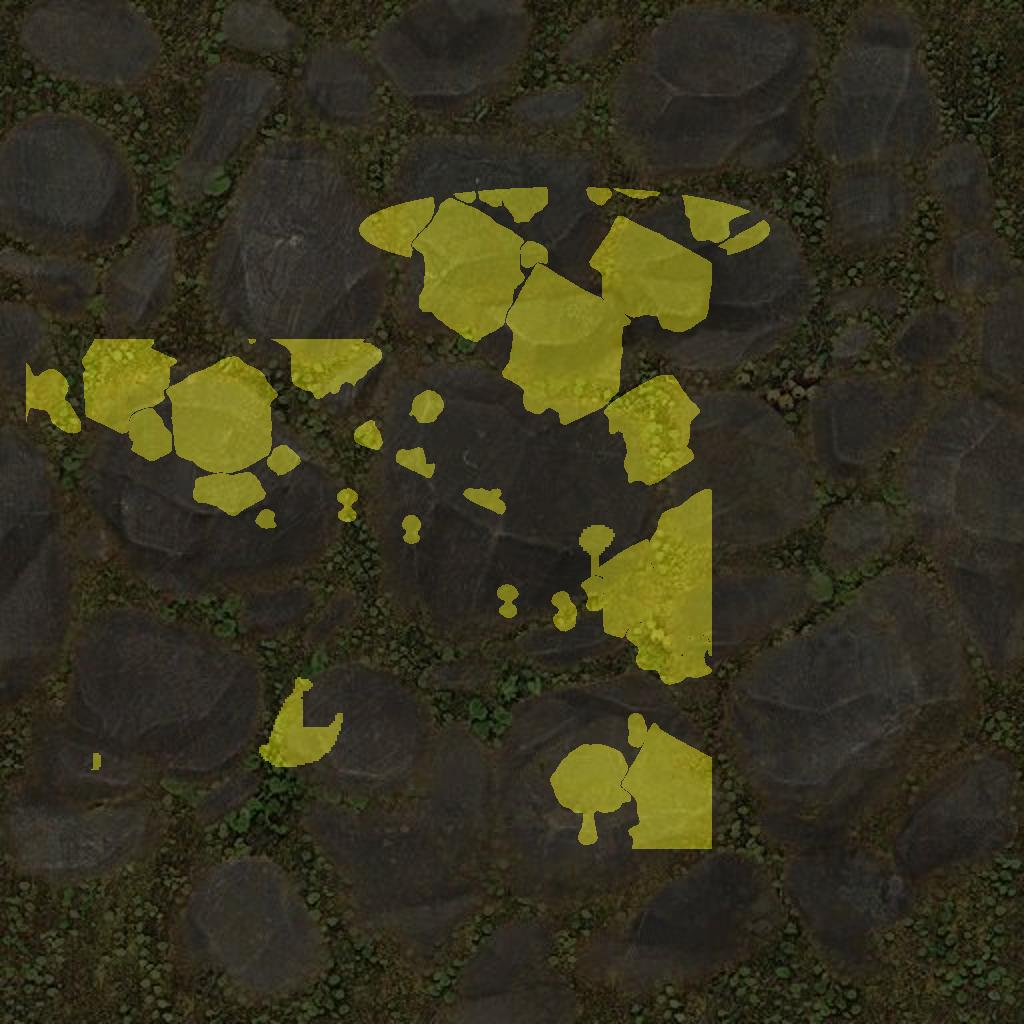} 
    \includegraphics[width=0.31\linewidth]{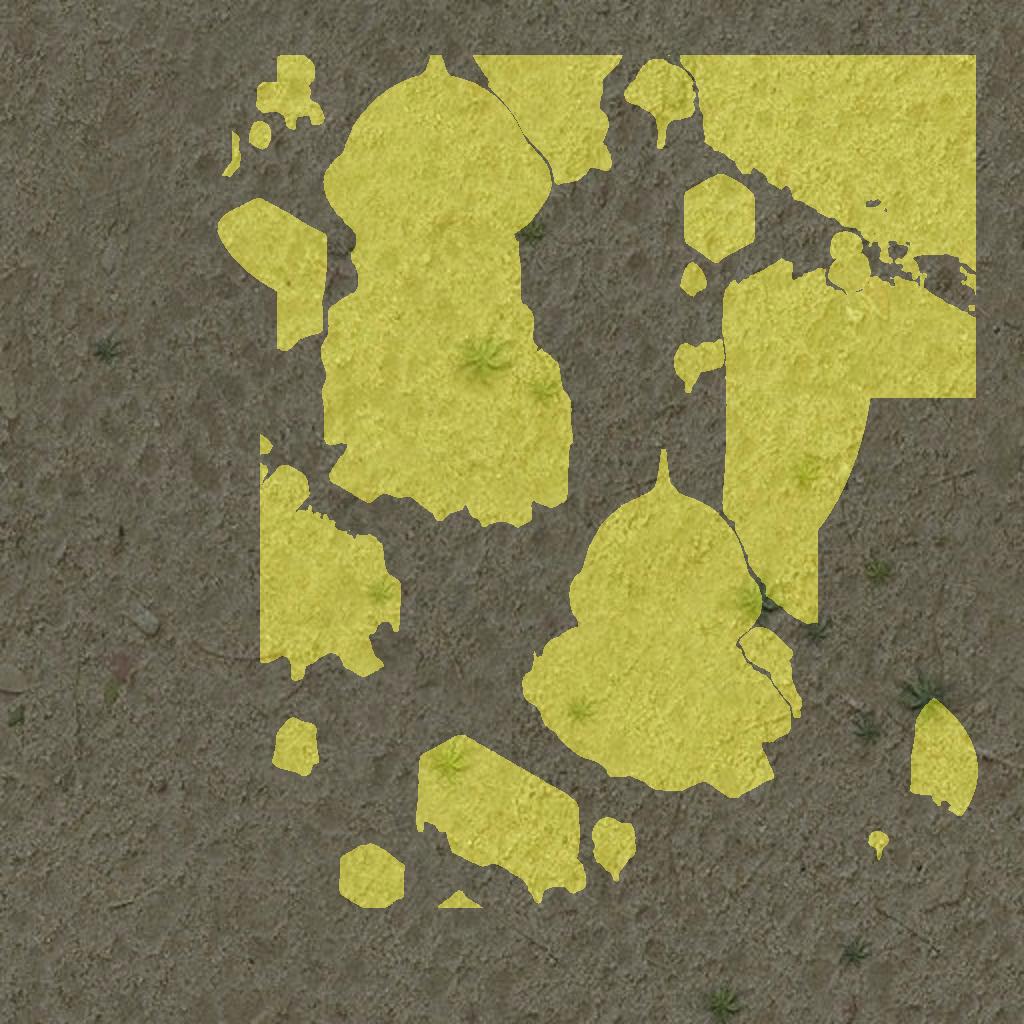}
    \includegraphics[width=0.31\linewidth]{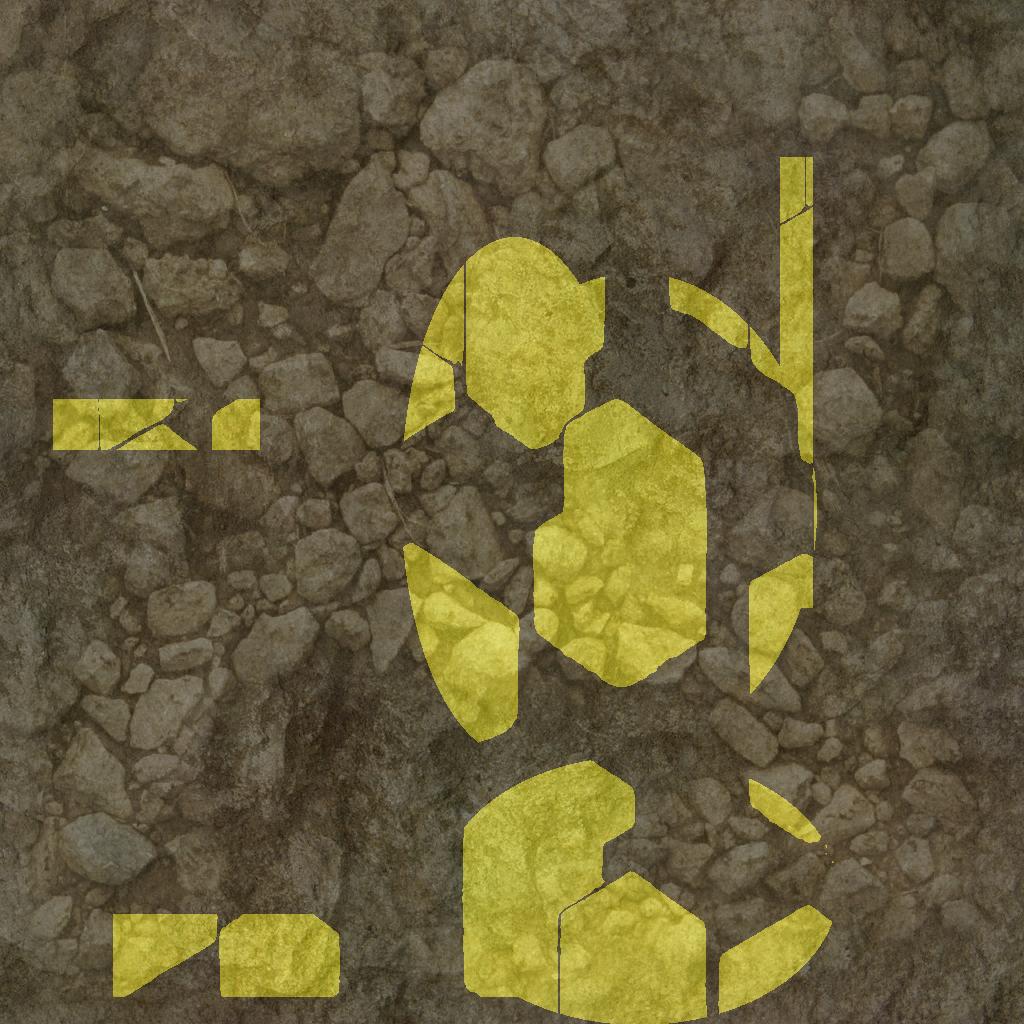}
    \end{subfigure}

    \begin{subfigure}{1\linewidth}
    \includegraphics[width=0.31\linewidth]{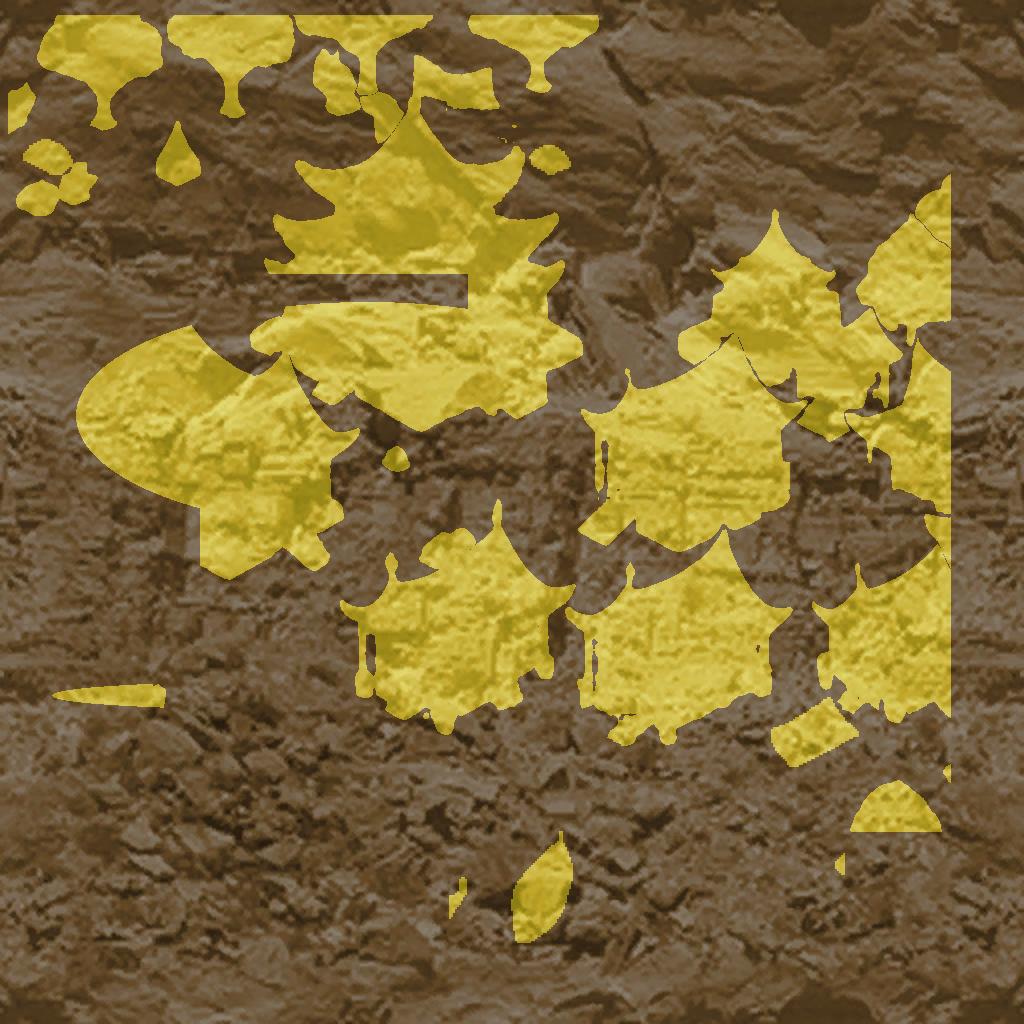} 
    \includegraphics[width=0.31\linewidth]{fig/inp_data/Inpaint_Empty_Train/Texture/s43_0_inpaint_masked.jpg}
    \includegraphics[width=0.31\linewidth]{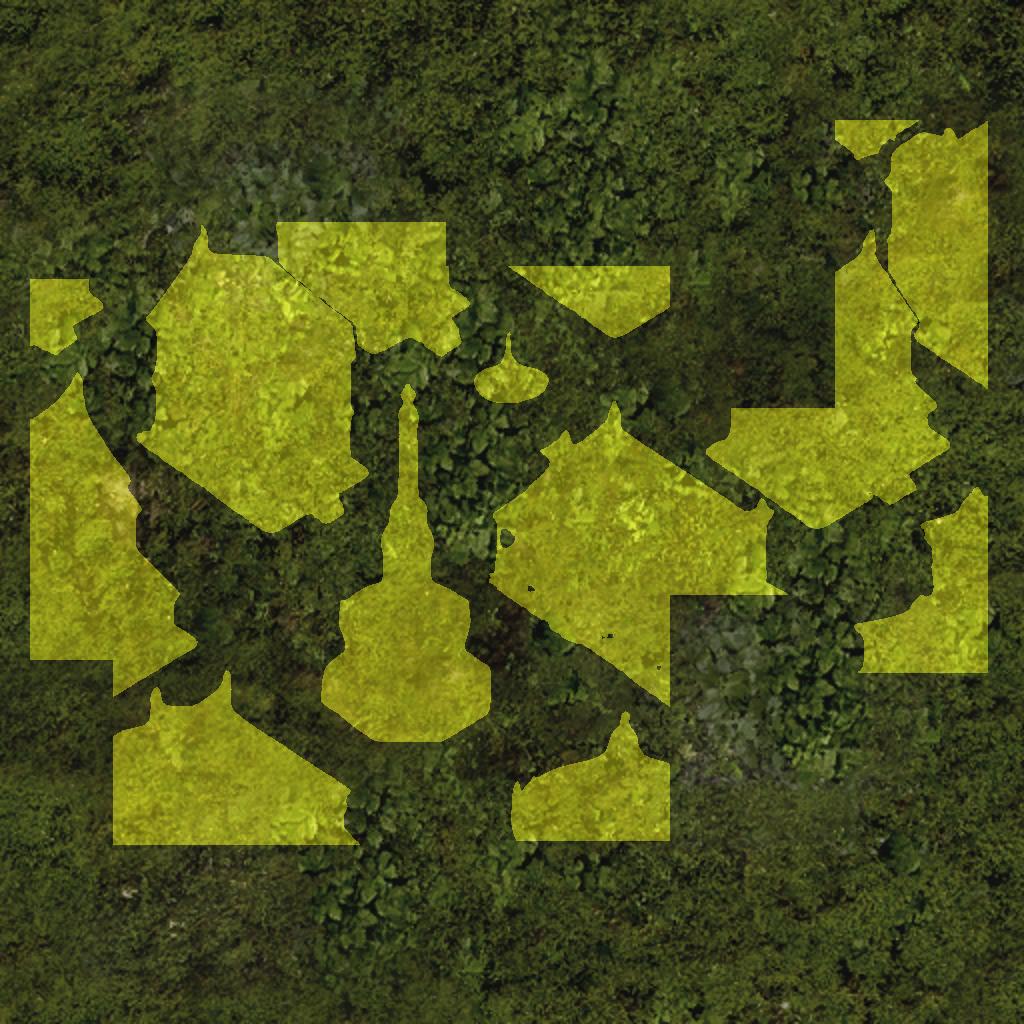}
    \end{subfigure}

  \caption{Examples of inpainting training data: texture images. }
  \label{fig:sup_inpaint_train_data_texture}
\end{figure}

\begin{figure}[h]
  \centering  
    \begin{subfigure}{1\linewidth}
    \includegraphics[width=0.31\linewidth]{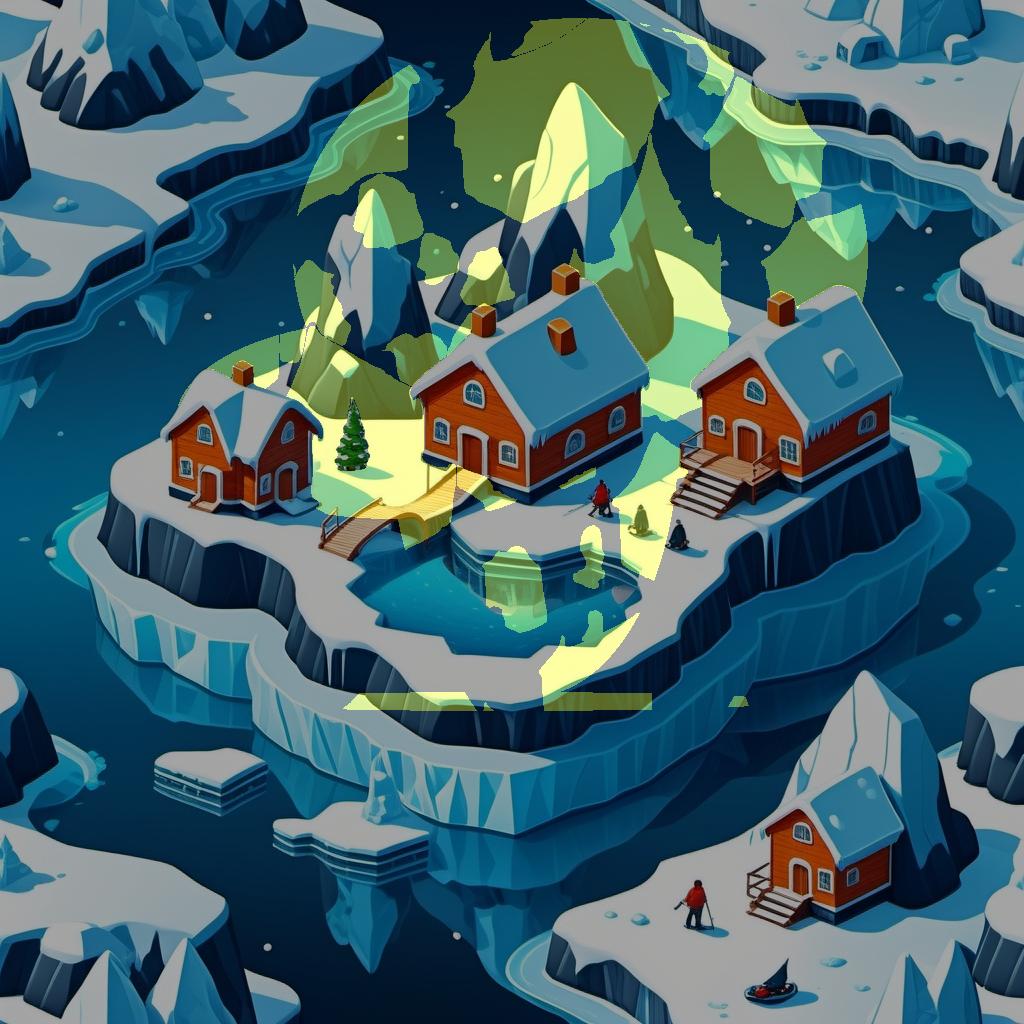} 
    \includegraphics[width=0.31\linewidth]{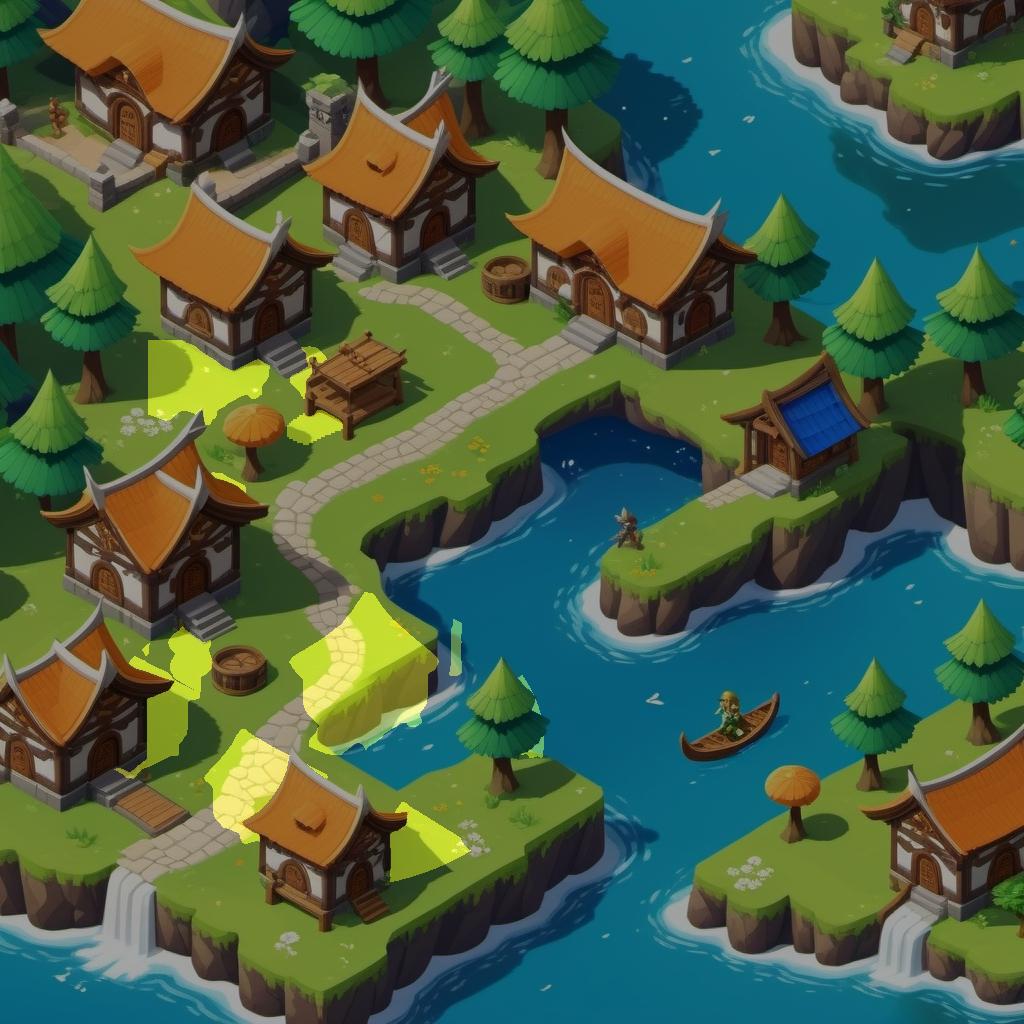}
    \includegraphics[width=0.31\linewidth]{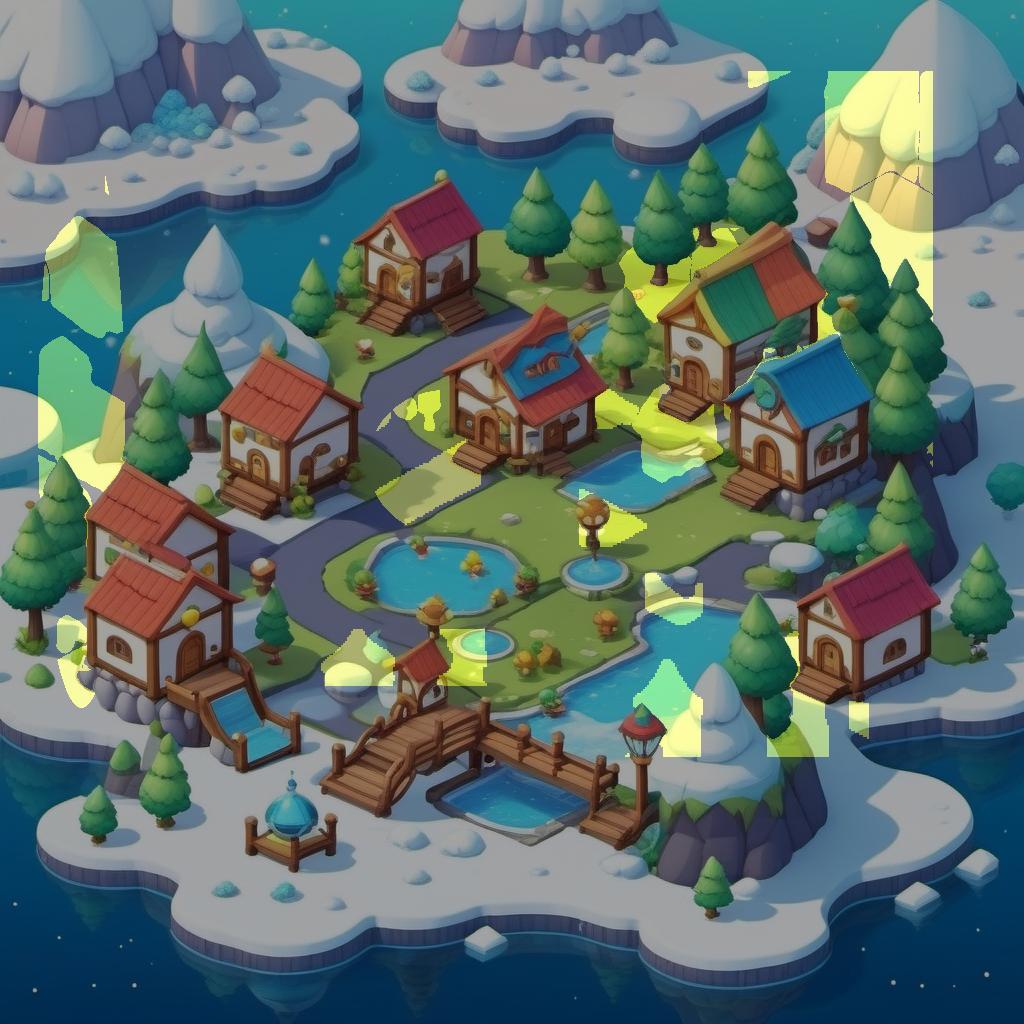}
    \end{subfigure}
     
    \begin{subfigure}{1\linewidth}
    \includegraphics[width=0.31\linewidth]{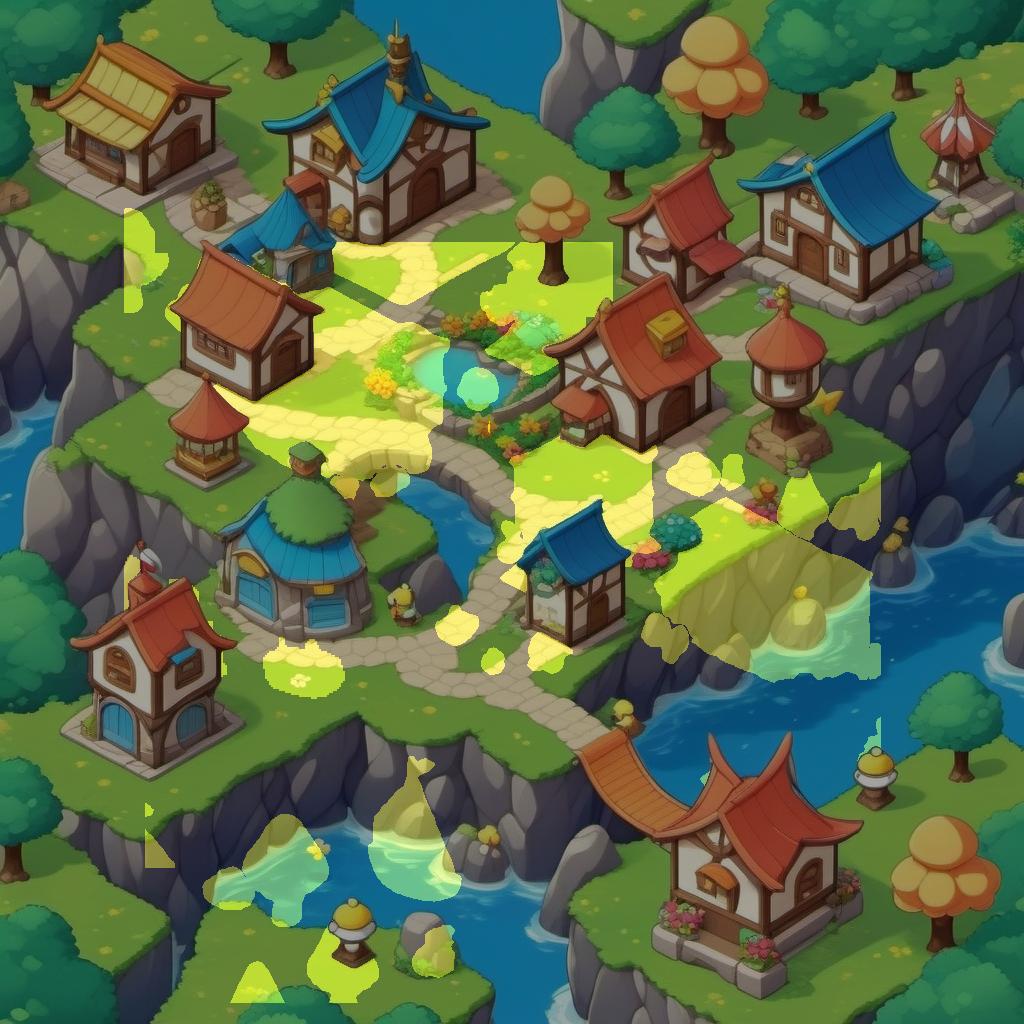} 
    \includegraphics[width=0.31\linewidth]{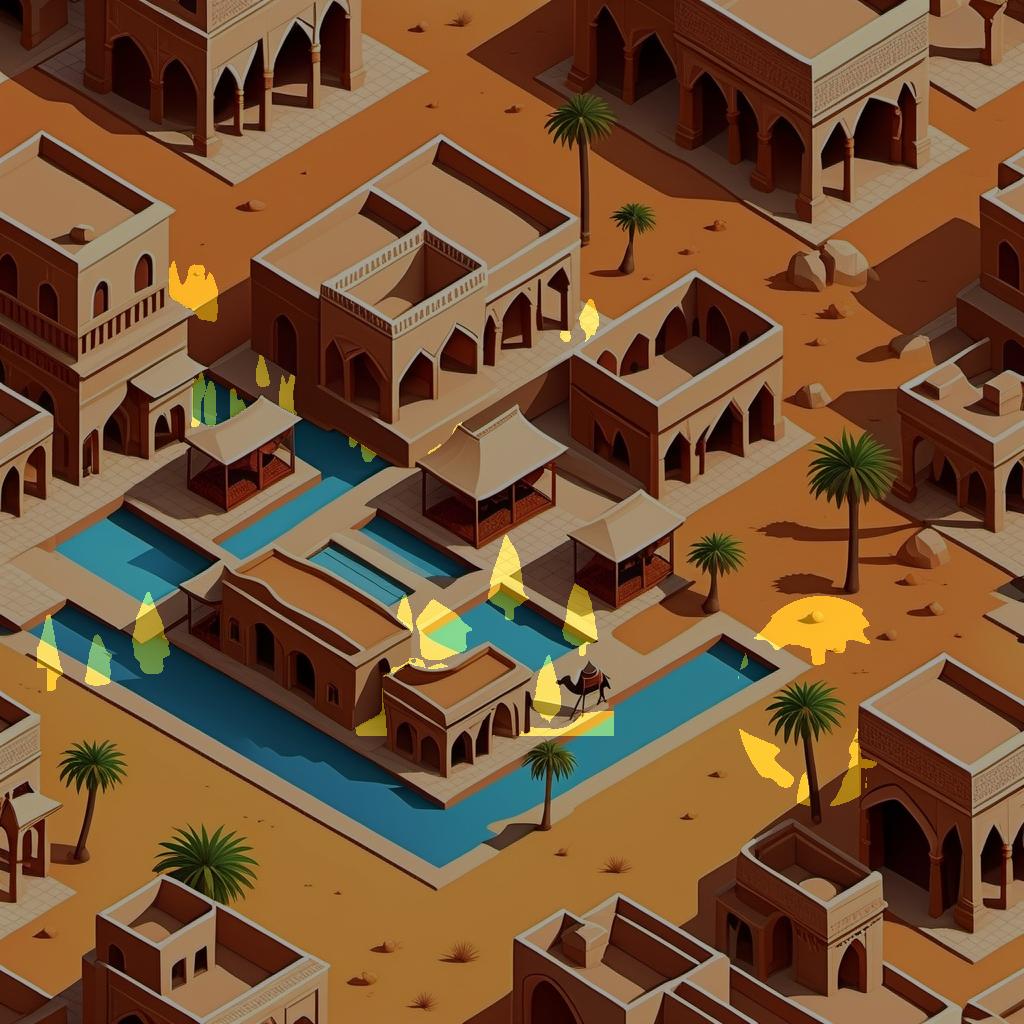}
    \includegraphics[width=0.31\linewidth]{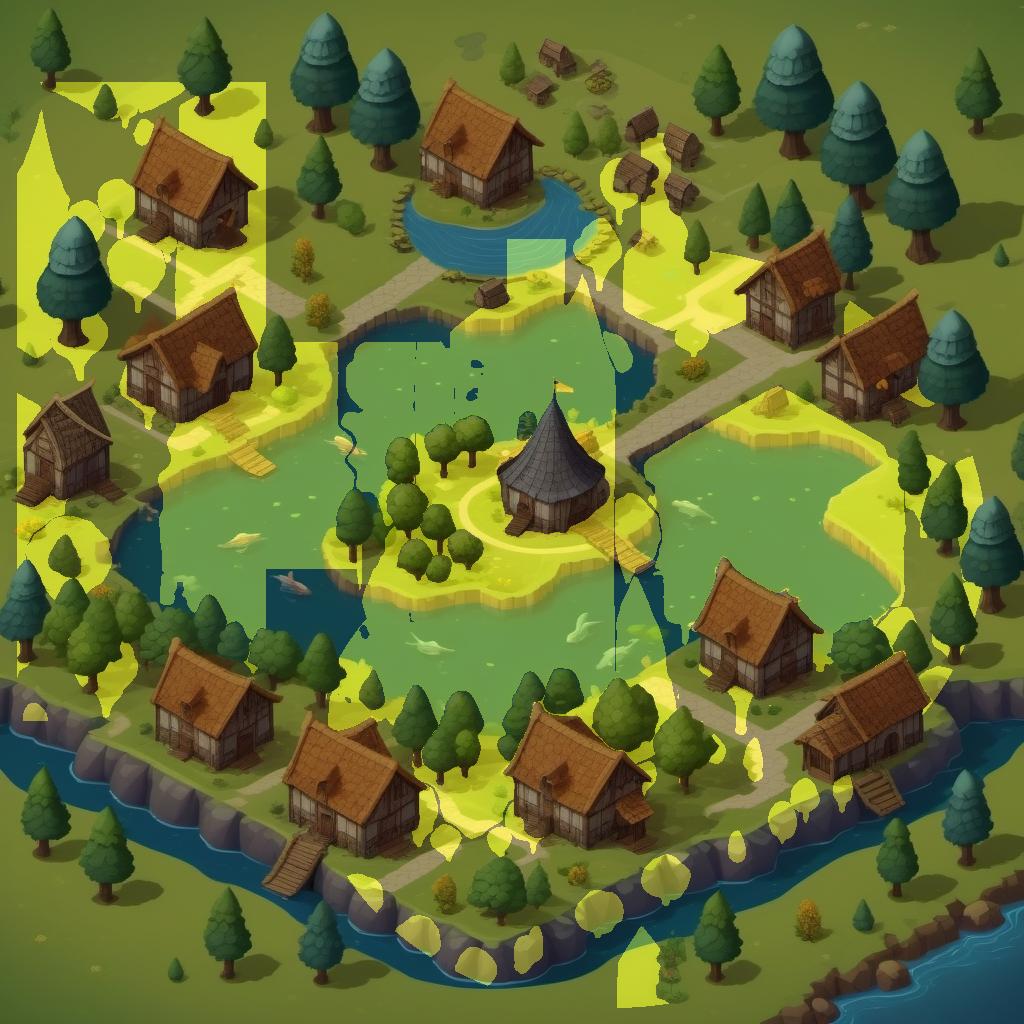}
    \end{subfigure}
  \caption{Examples of inpainting training data: full isometric images.}
  \label{fig:sup_inpaint_train_data_iso}
\end{figure}

To ensure a diverse range of masking scenarios and minimize the distribution discrepancy between training and inference masks, we utilize the intersection of random masks and pseudo-foreground masks for training the basemaps. These pseudo-foreground masks are randomly sampled from the foreground masks of the isometric dataset. Examples of these intersection results are shown in Fig~\ref{fig:sup_inpaint_train_data_empty},~\ref{fig:sup_inpaint_train_data_texture}, and~\ref{fig:sup_inpaint_train_data_iso}. For isometric images with foreground objects, only the background area can be masked and considered as inpainted ground truth. We therefore use the intersection of background masks and random masks as the training masks.

\subsection{Comparison with SDXL Inpainting}
\begin{figure}[h]
    \centering
    \begin{subfigure}{1\linewidth}
        \includegraphics[width=0.31\linewidth]{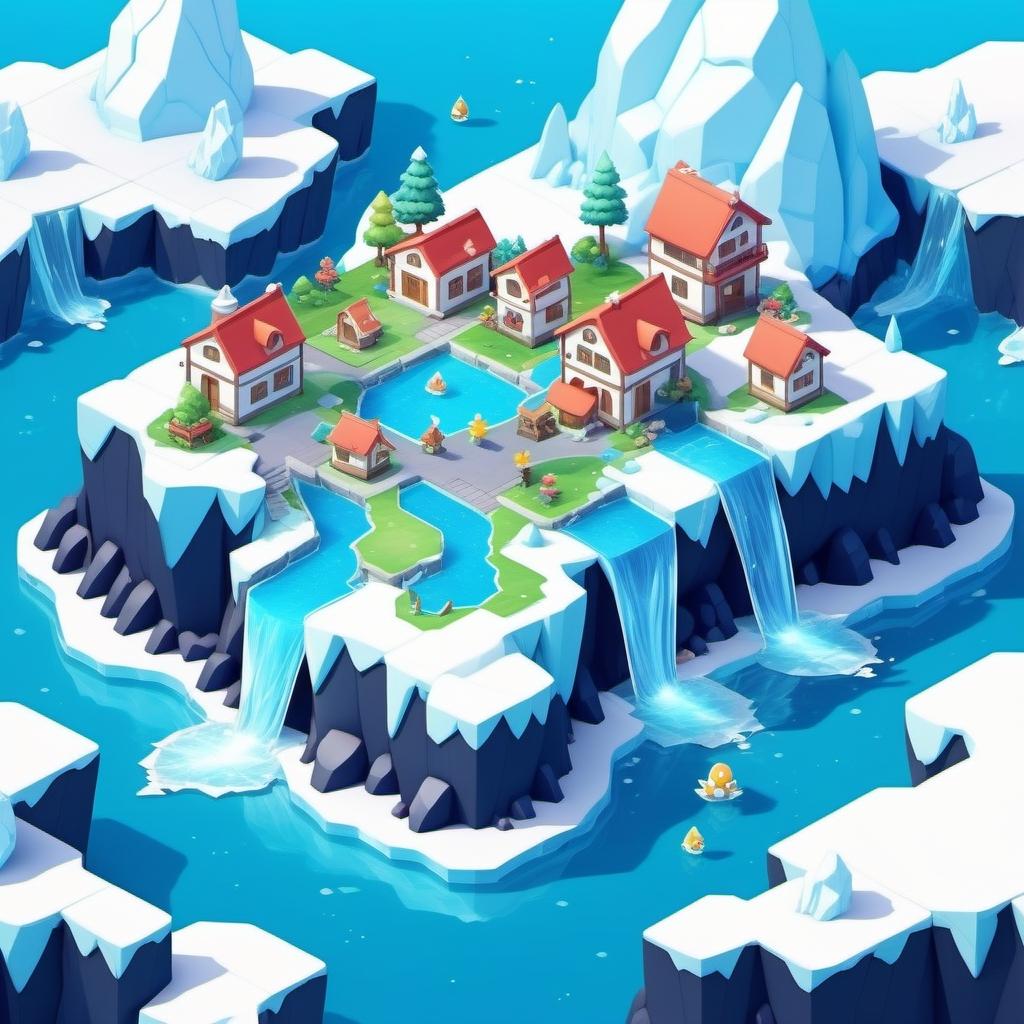} 
        \includegraphics[width=0.31\linewidth]{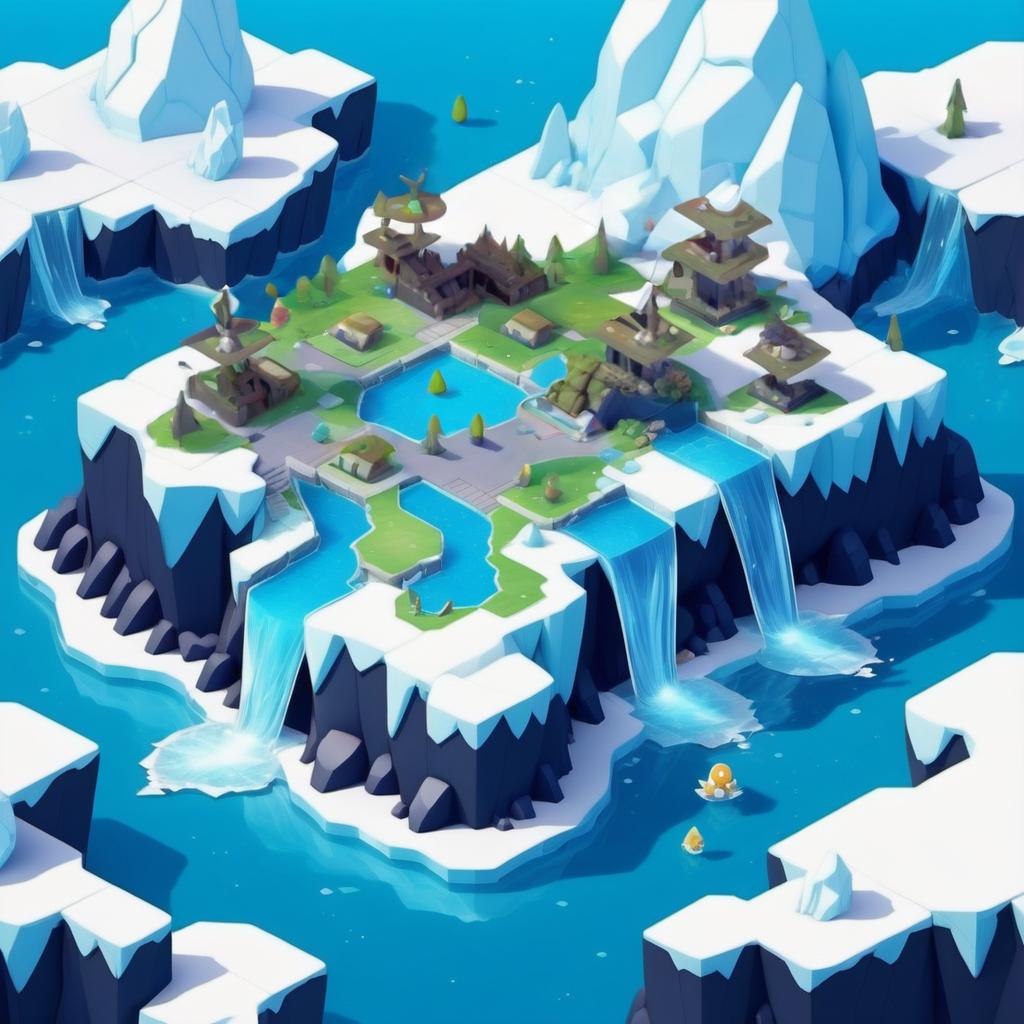} 
        \includegraphics[width=0.31\linewidth]{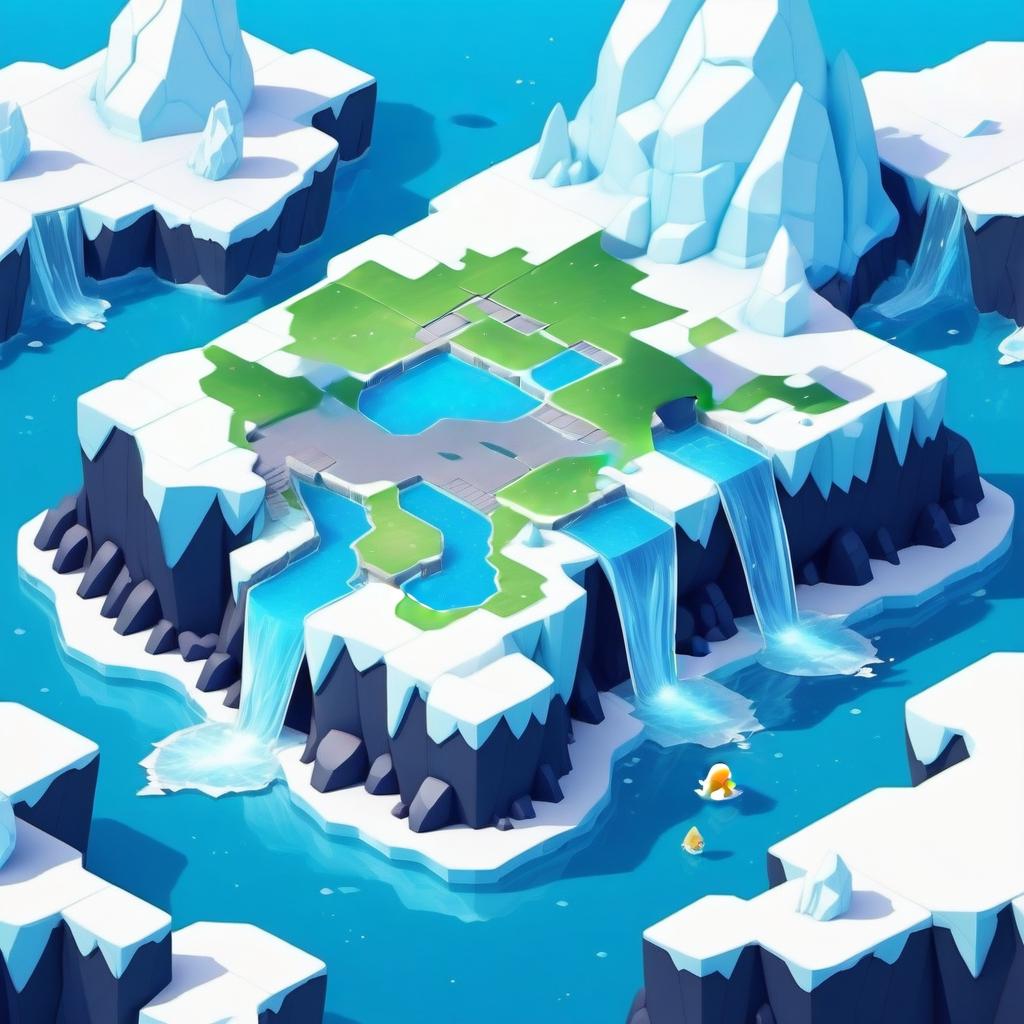} 
    \end{subfigure}

    \begin{subfigure}{1\linewidth}
        \includegraphics[width=0.31\linewidth]{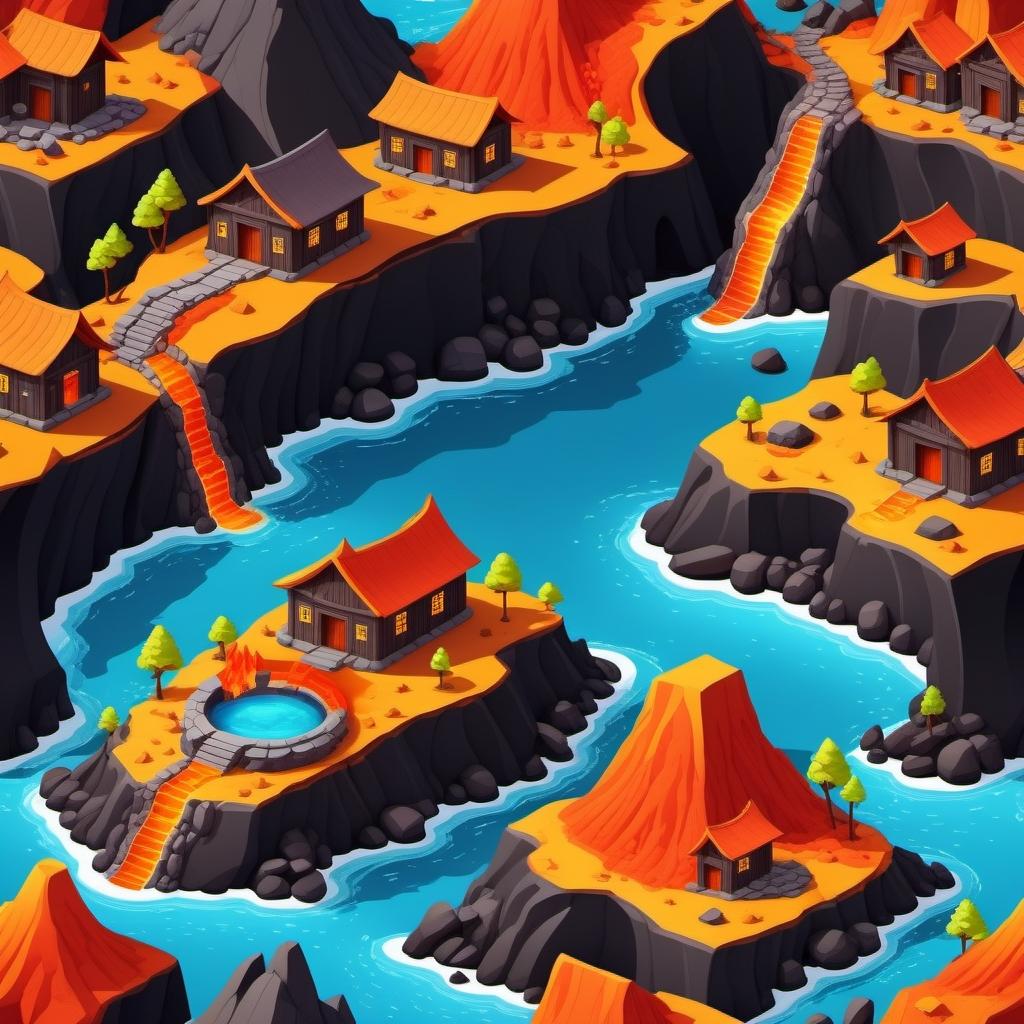} 
        \includegraphics[width=0.31\linewidth]{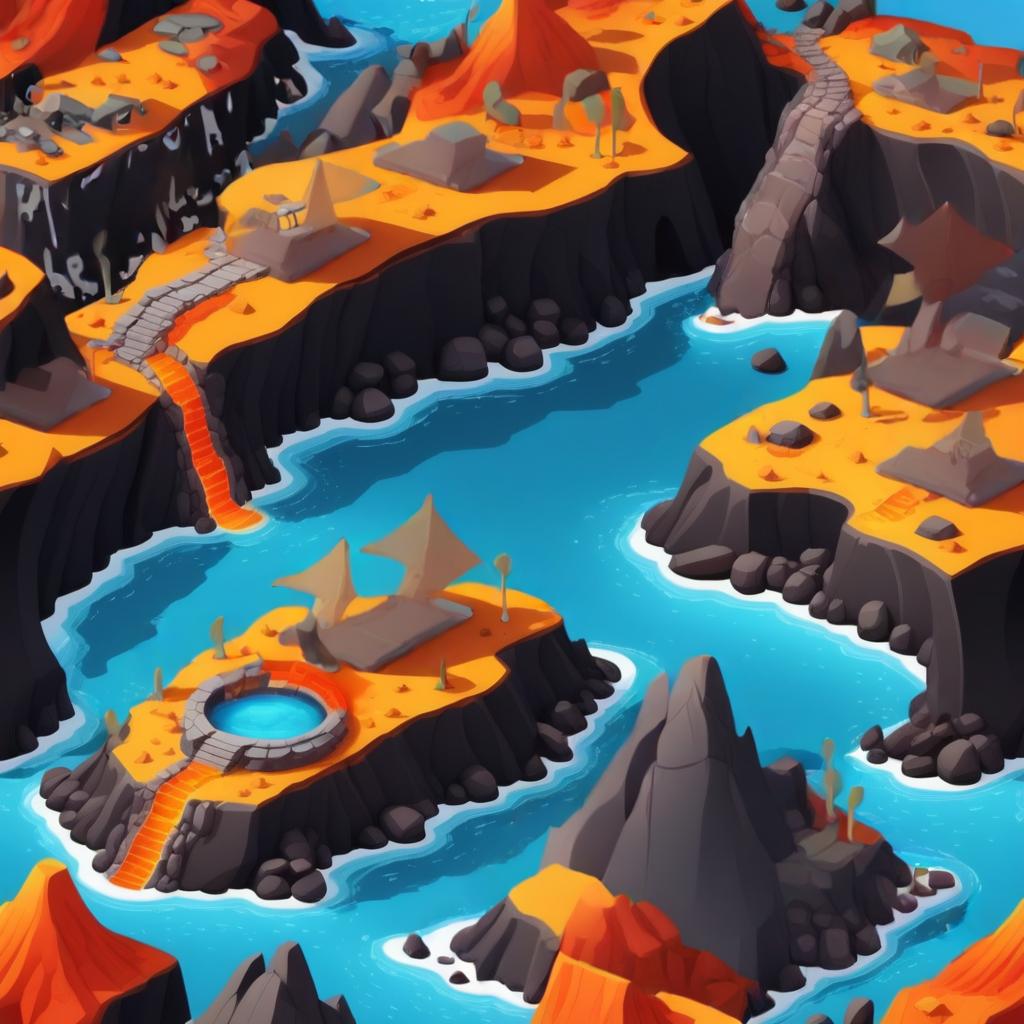} 
        \includegraphics[width=0.31\linewidth]{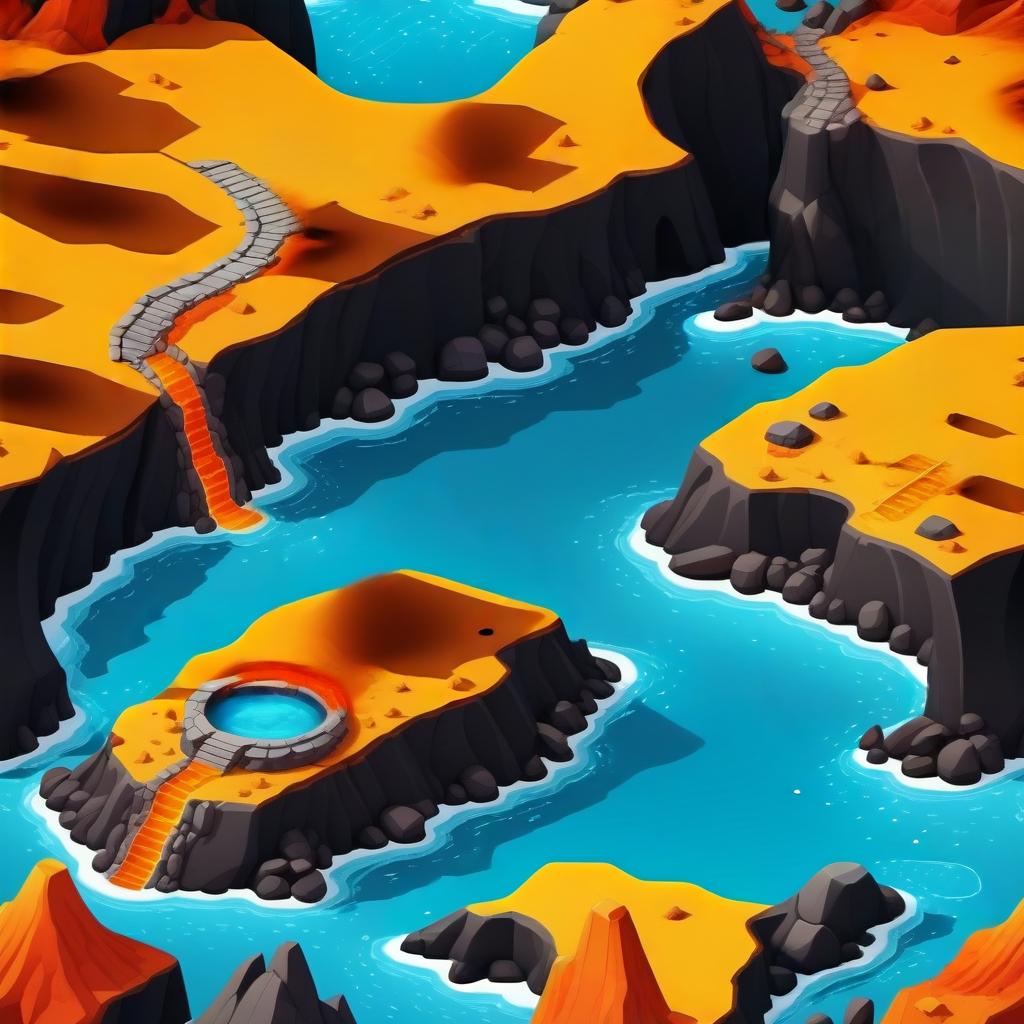} 
    \end{subfigure}

    \begin{subfigure}{1\linewidth}
        \includegraphics[width=0.31\linewidth]{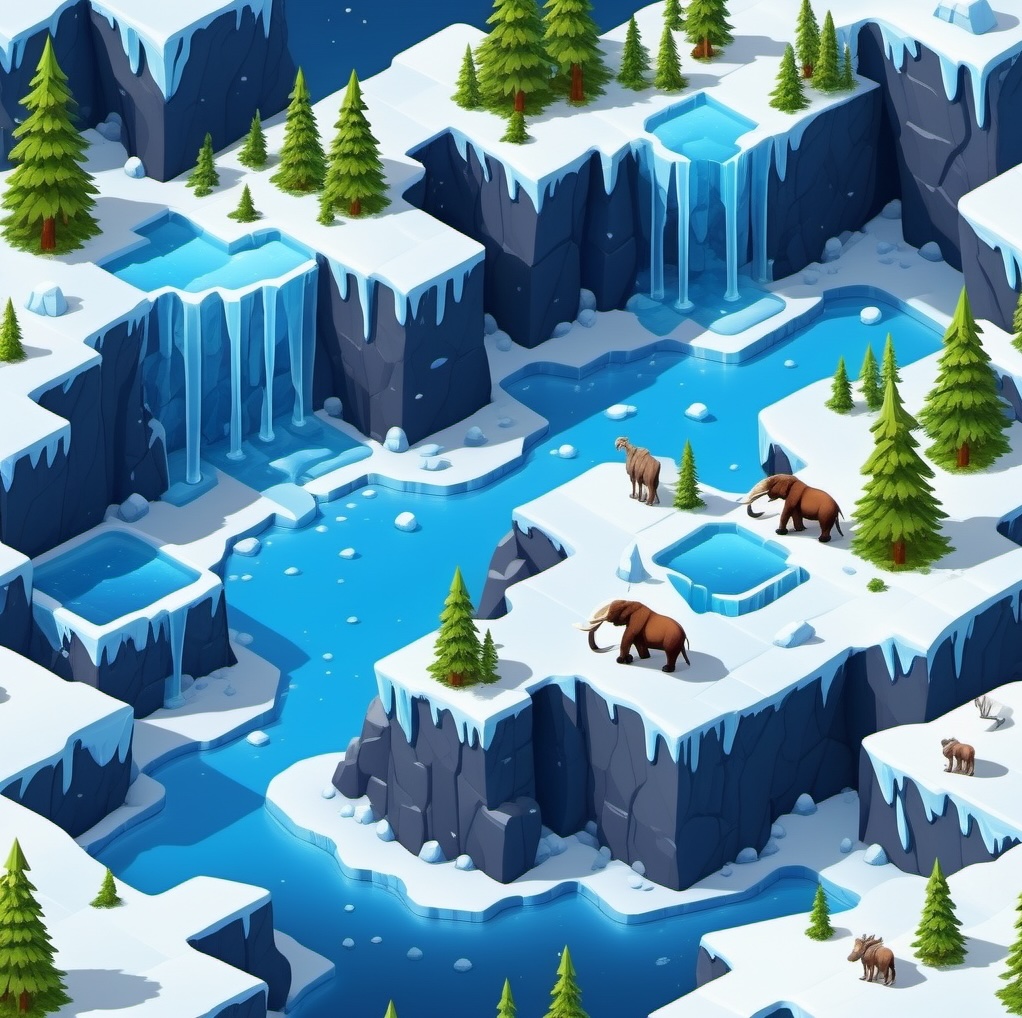} 
        \includegraphics[width=0.31\linewidth]{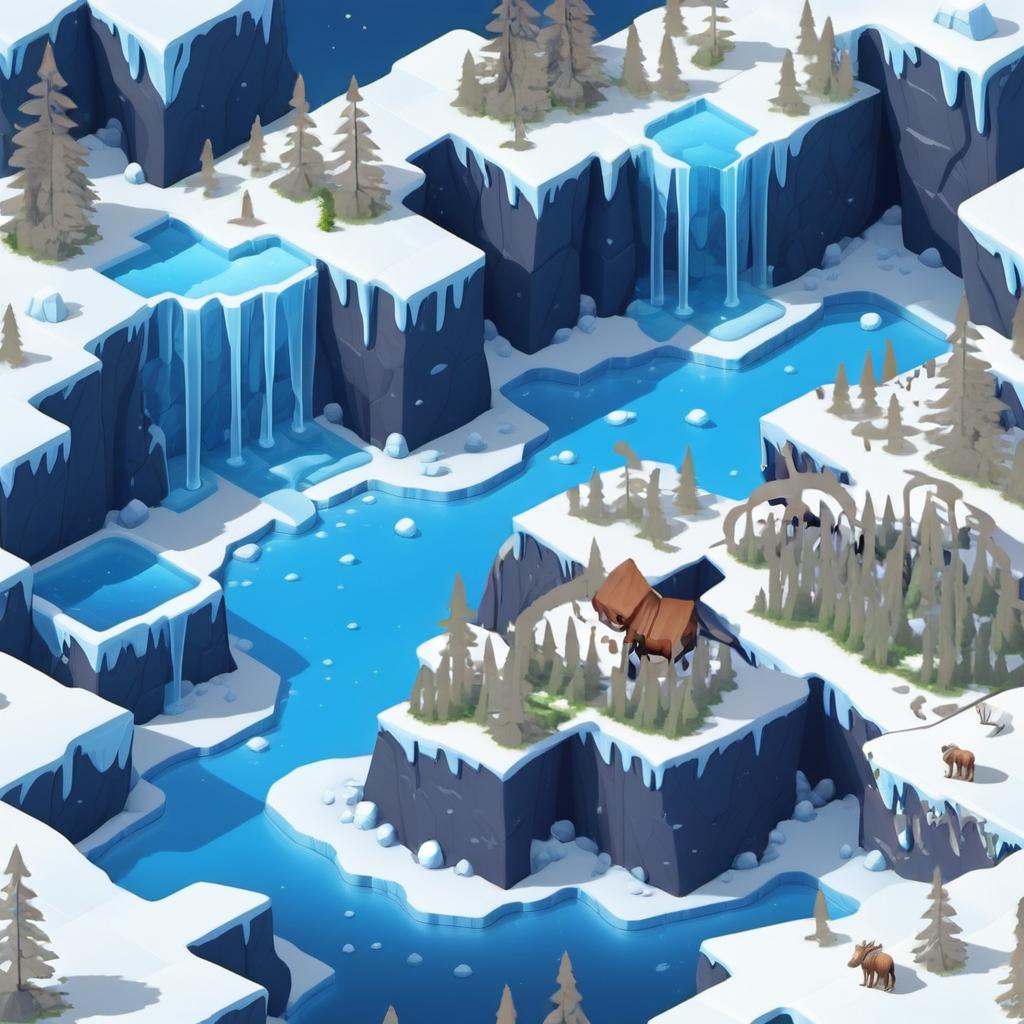} 
        \includegraphics[width=0.31\linewidth]{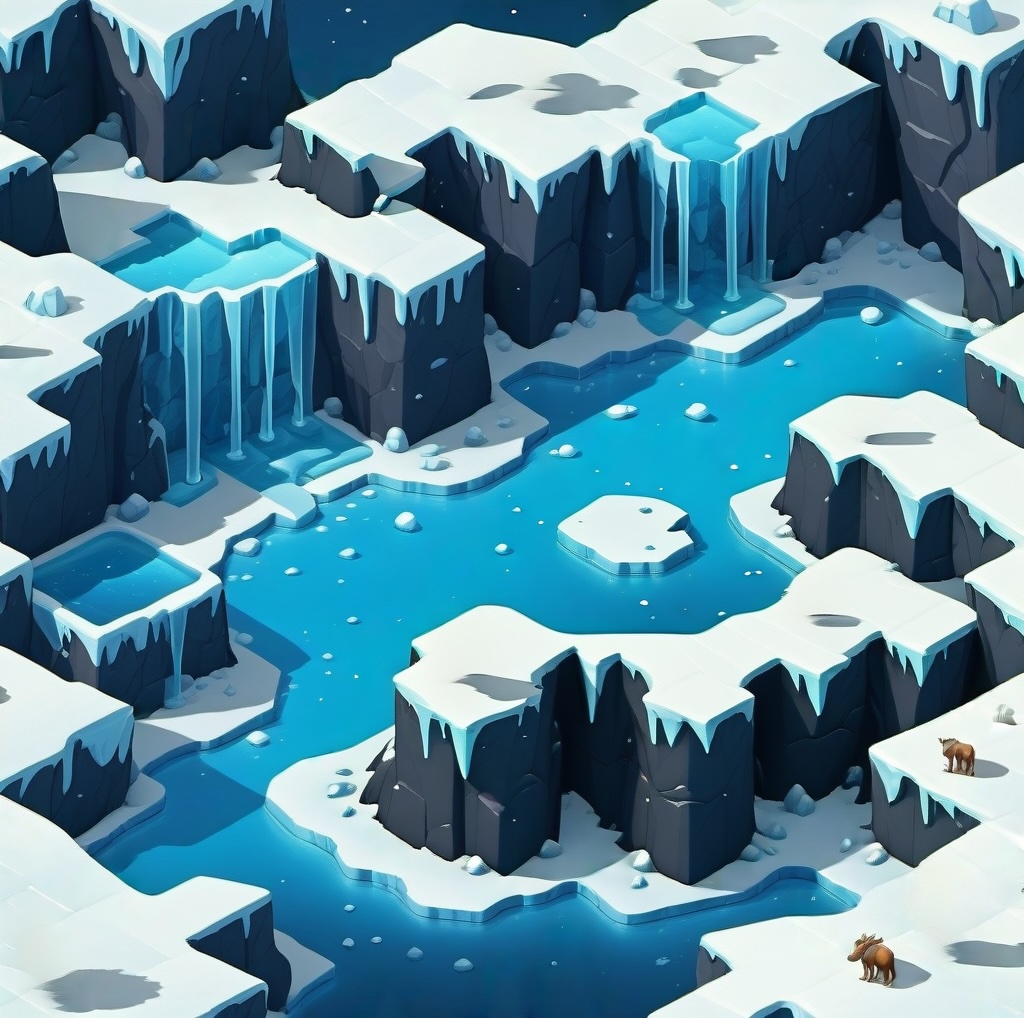} 
    \end{subfigure} 
    
    \caption{Comparison of basemap inpainting between ours (right) and SDXL-Inpaint (middle) on isometric test dataset (left). From the above results we can see our method produces much cleaner empty basemaps of the terrain.} 
    \label{fig:inpaint_result_supp}
\end{figure}

Fig.~\ref{fig:inpaint_result_supp} compares several inpainting results between our proposed method and SDXL\_Inpainting.

\subsection{2D Image Generation Results}
Fig. ~\ref{fig:con_inp_sup2} demonstrates a variety of supplementary examples of isometric images generated by ControlNet from texts and sketches, along with the results of basemap inpainting.

\begin{figure}[t!]
    \centering
    \begin{subfigure}{1\linewidth}
    \includegraphics[width=0.31\linewidth]{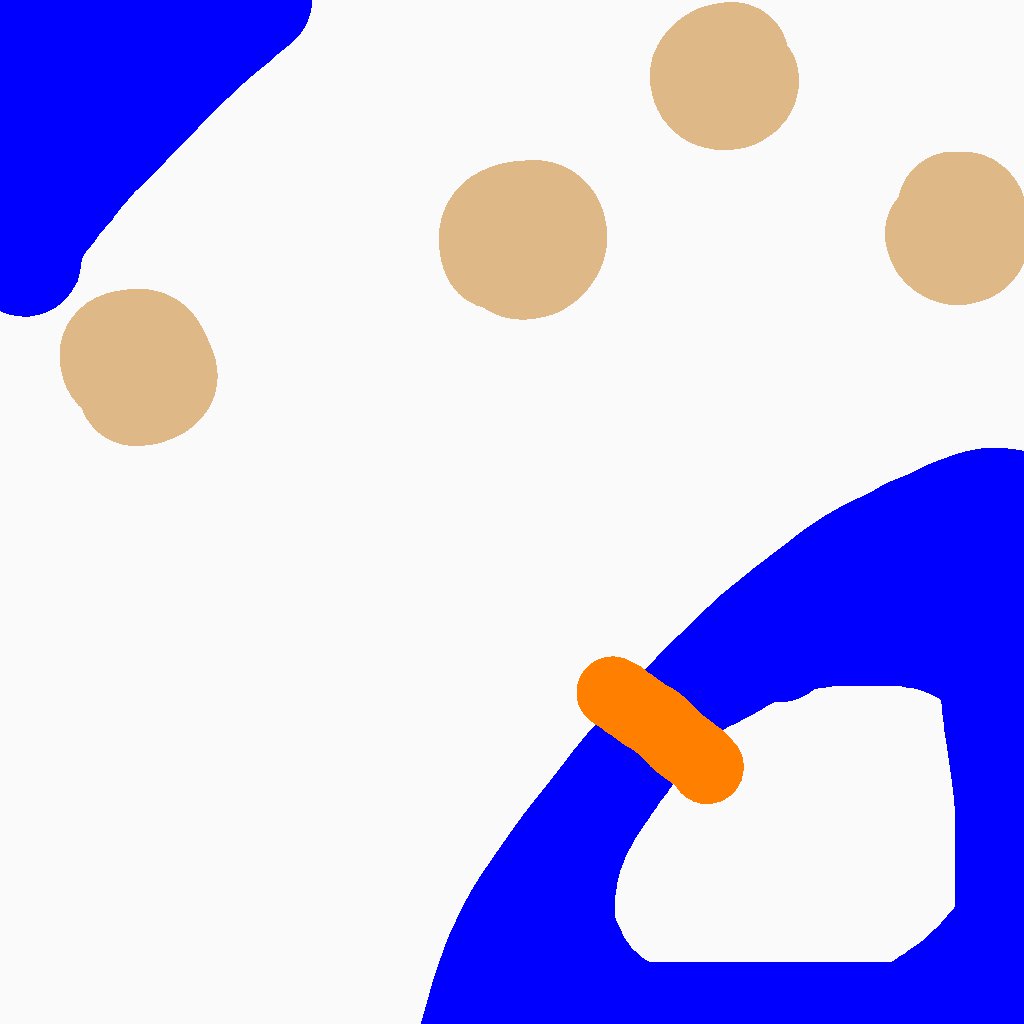}
    \includegraphics[width=0.31\linewidth]{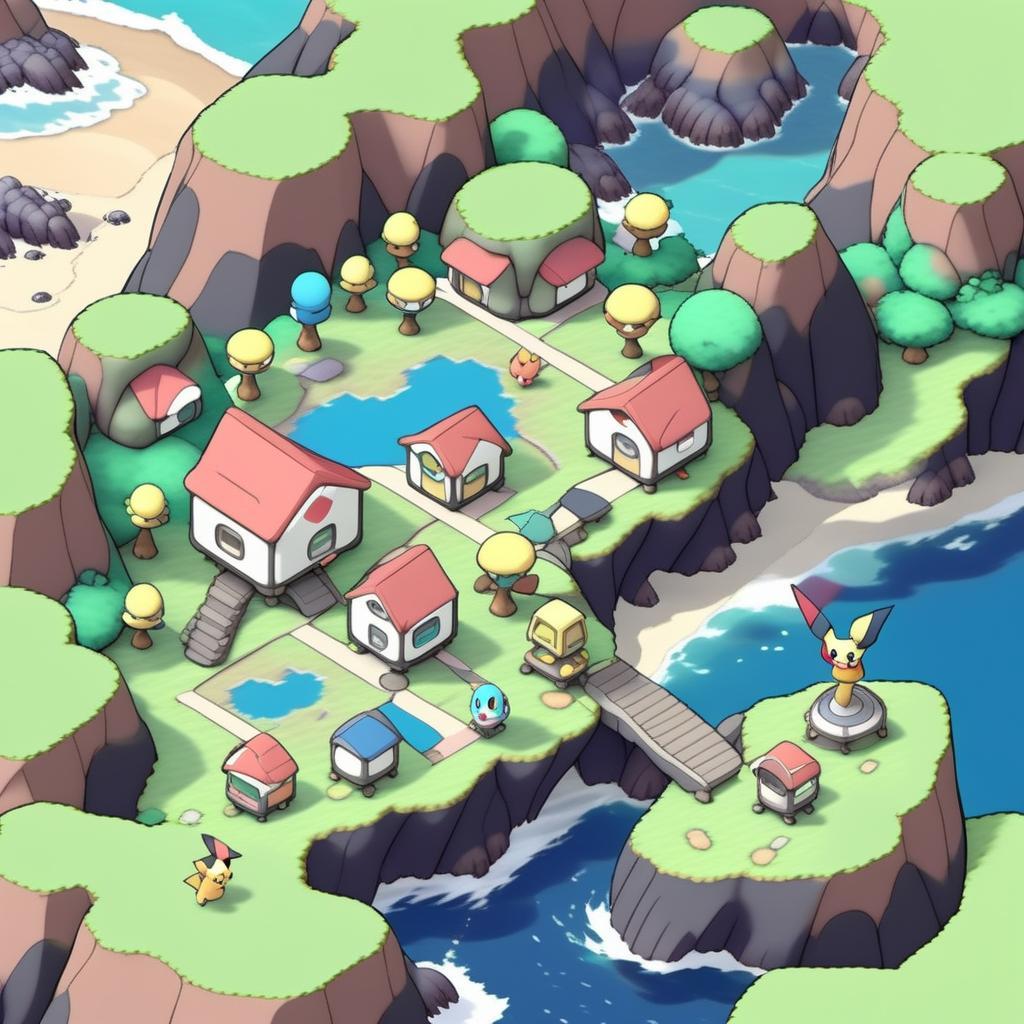} 
    \includegraphics[width=0.31\linewidth]{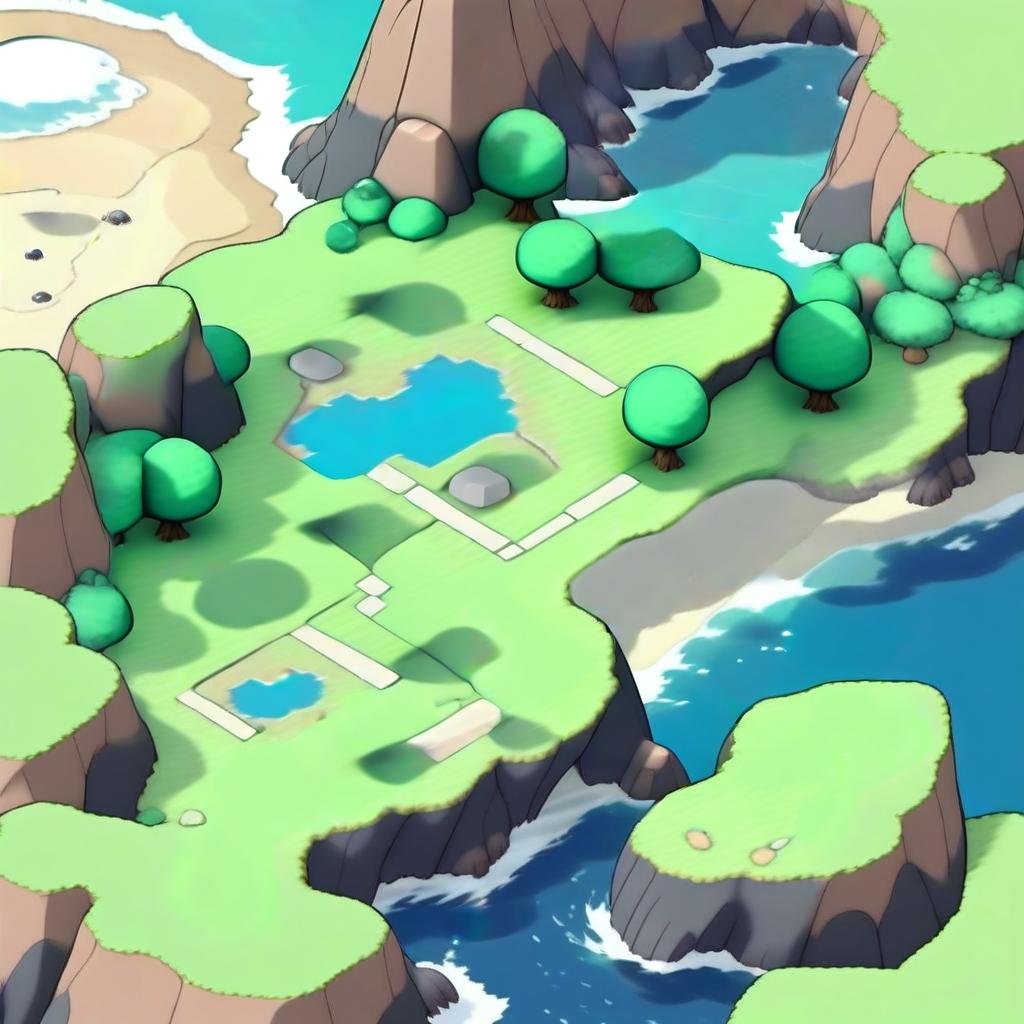} 
    \caption{An Pokemon-style isometric town around a craggy coastline}
    \label{fig:Pokemon_coastline}
    \end{subfigure}  

    \begin{subfigure}{1\linewidth}
    \includegraphics[width=0.31\linewidth]{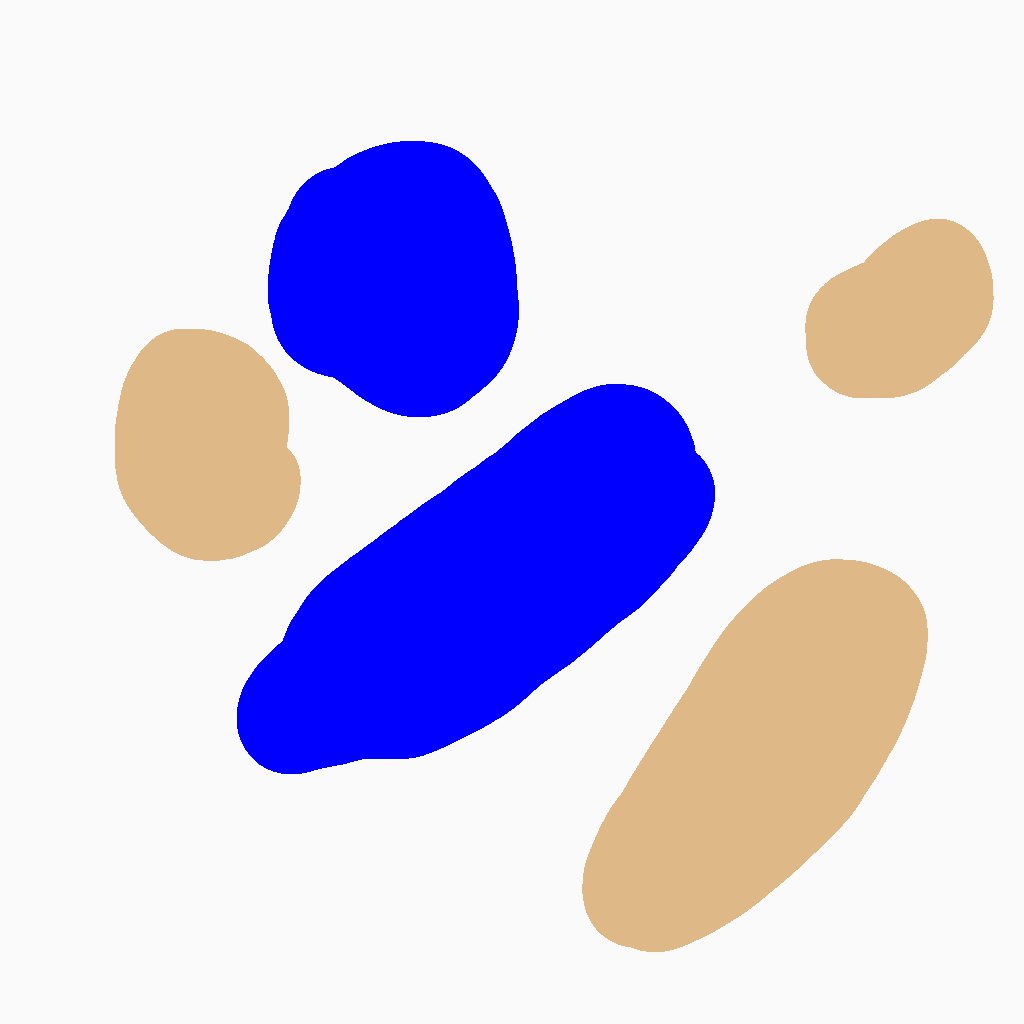}
    \includegraphics[width=0.31\linewidth]{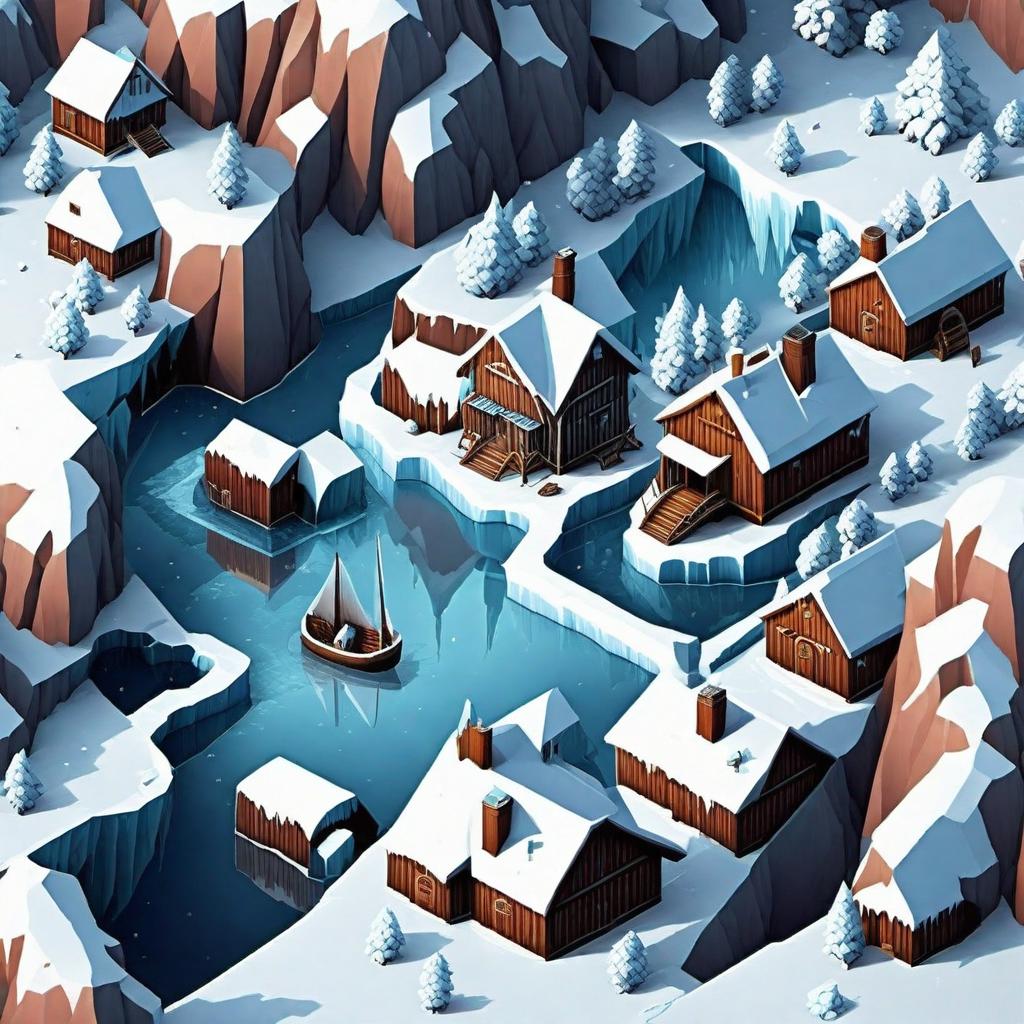} 
    \includegraphics[width=0.31\linewidth]{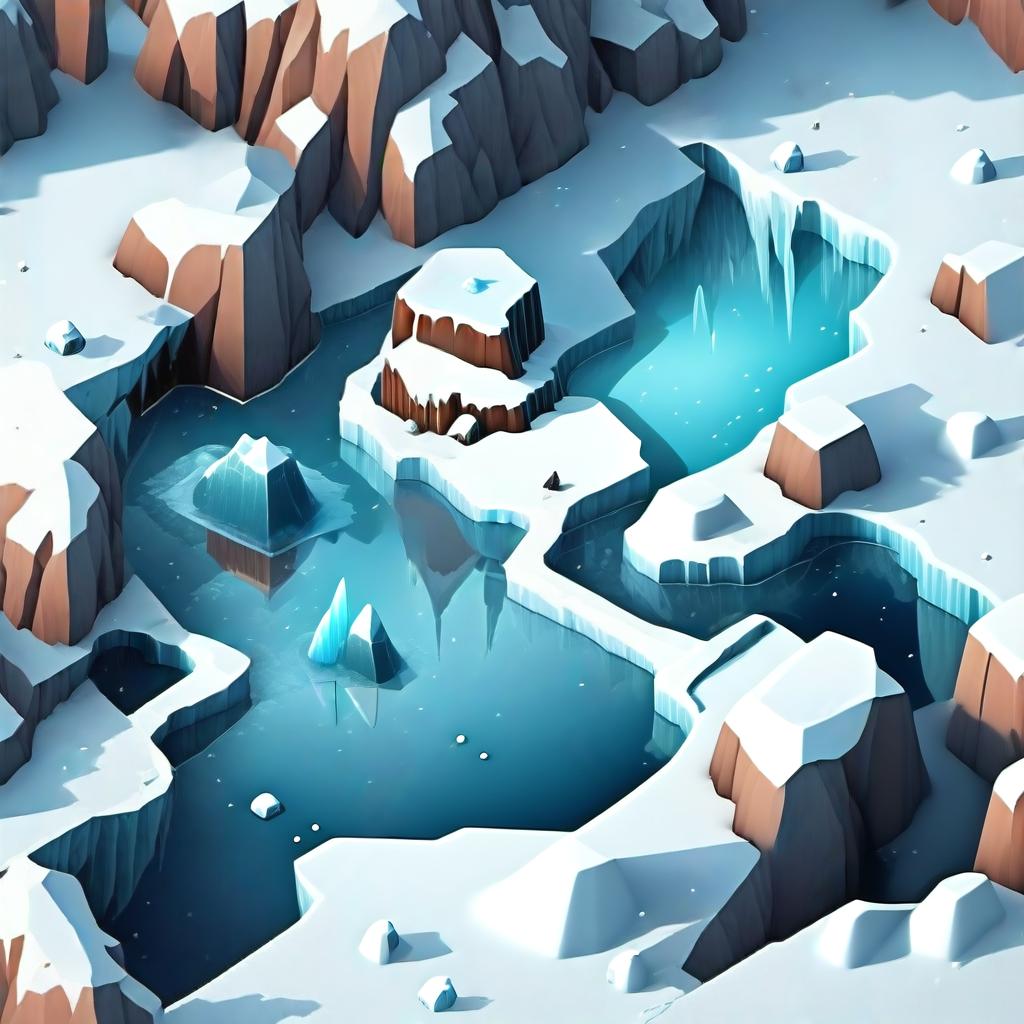} 
    \caption{A beautiful isometric world of ice and snow.}
    \end{subfigure}   
    \begin{subfigure}{1\linewidth}
    \includegraphics[width=0.31\linewidth]{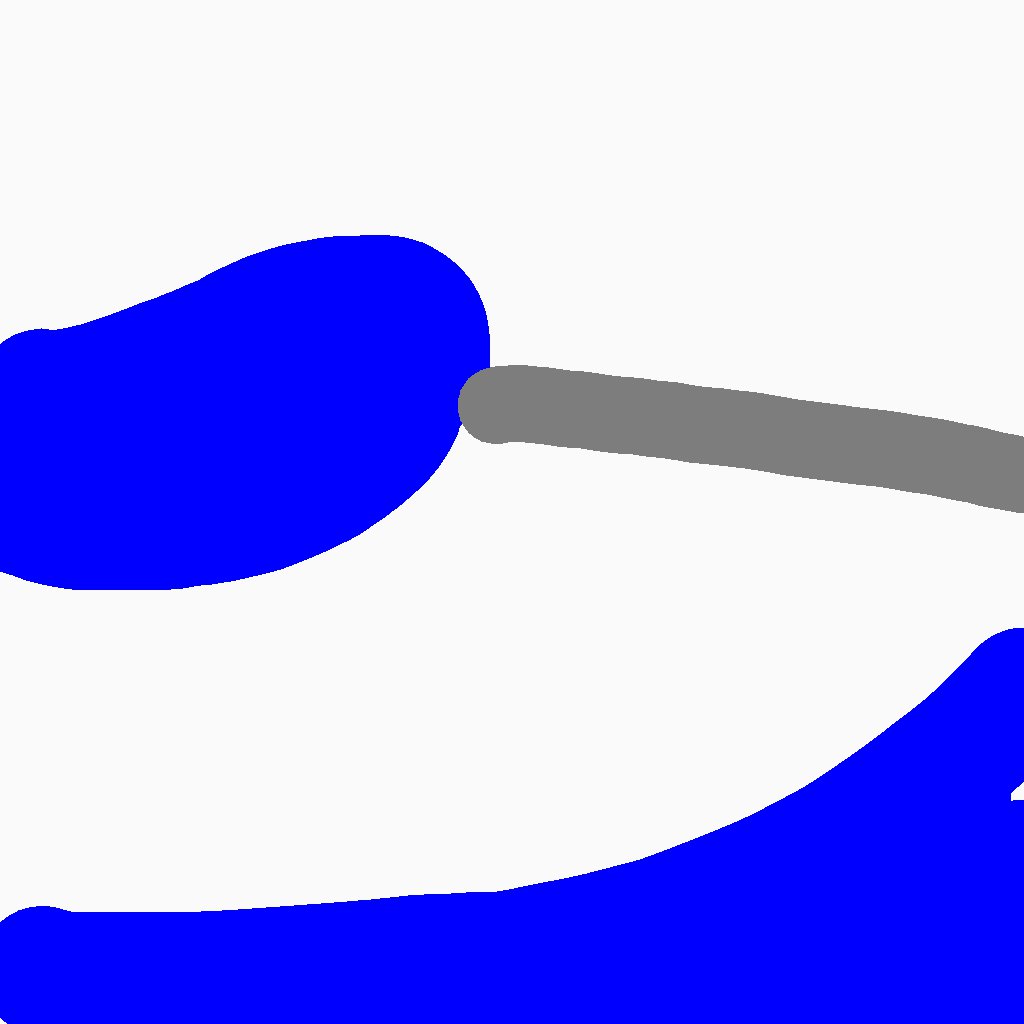}
    \includegraphics[width=0.31\linewidth]{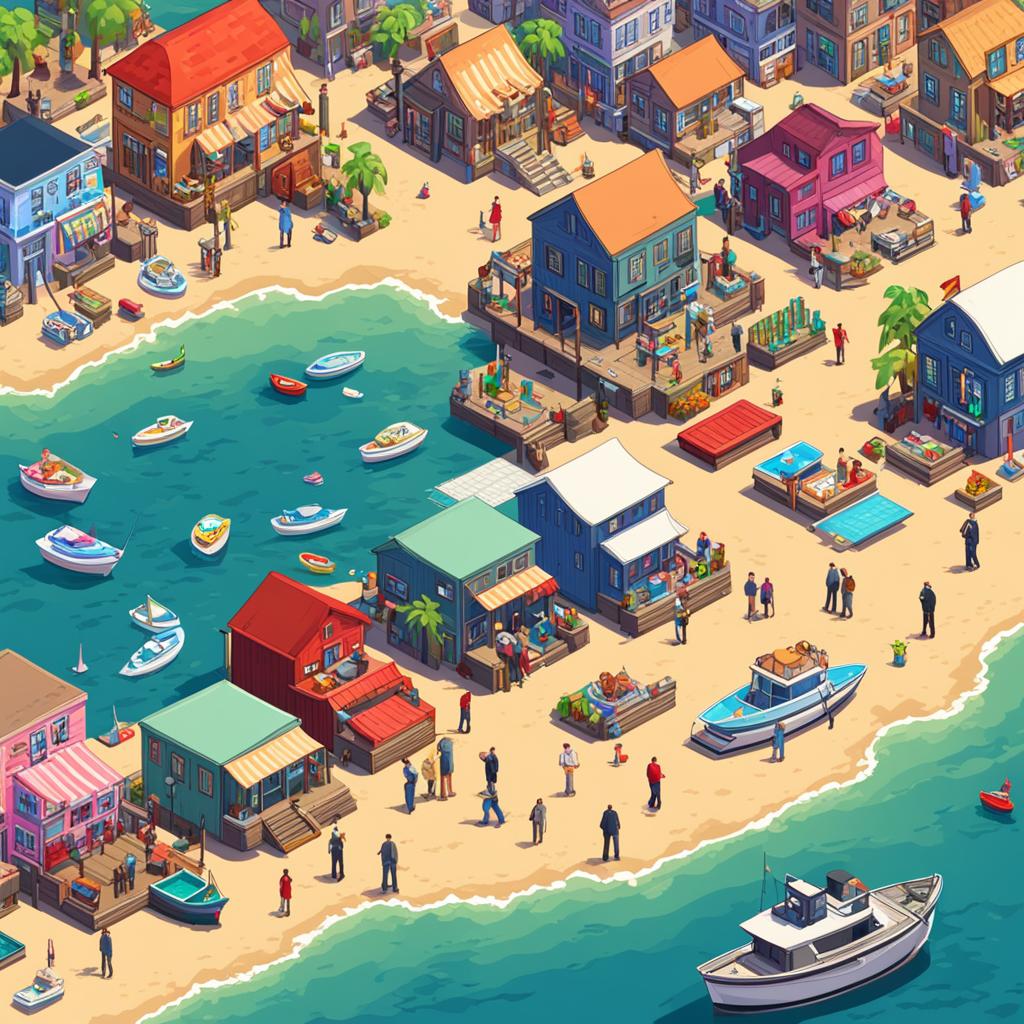} 
    \includegraphics[width=0.31\linewidth]{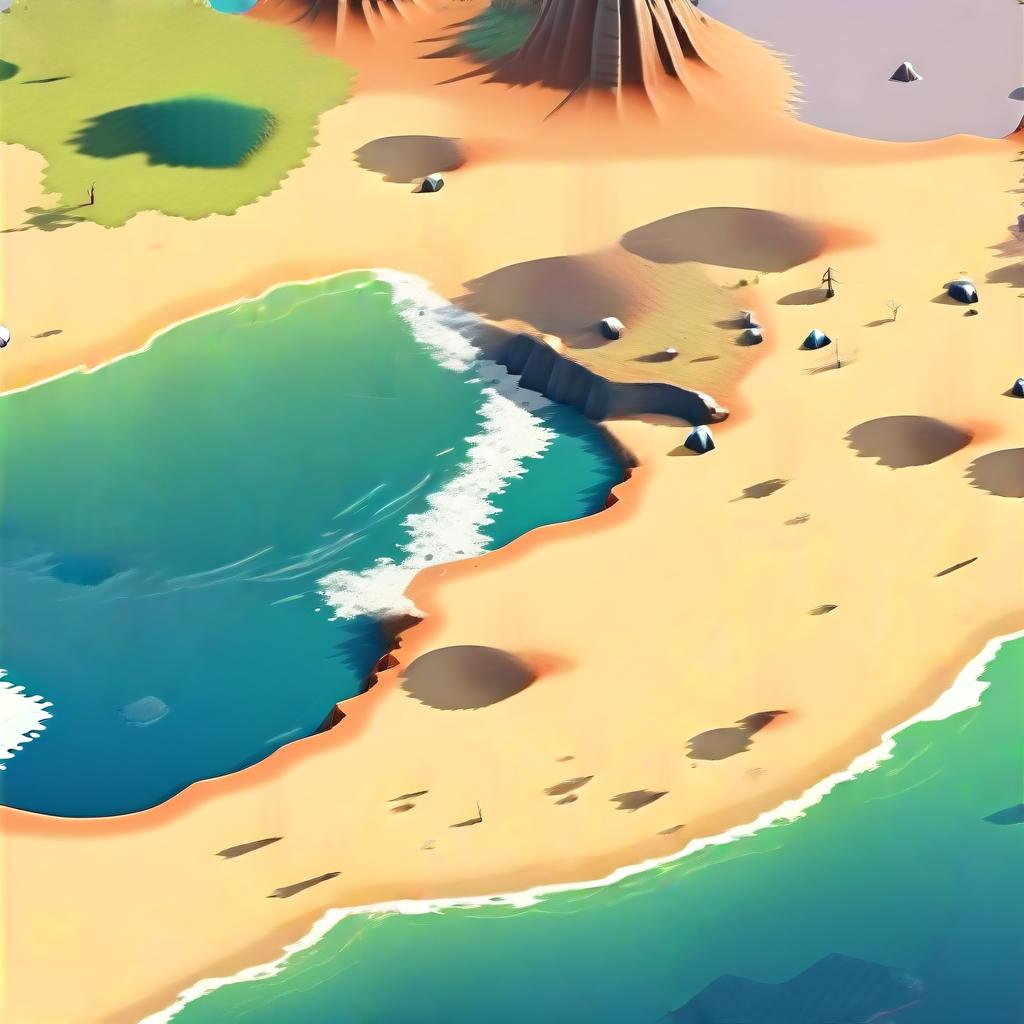 } 
    \caption{A GTA-style coastal town with charming seafront, colorful buildings, and fishing boats dotting the harbor.}
    \end{subfigure}  

    \begin{subfigure}{1\linewidth}
    \includegraphics[width=0.31\linewidth]{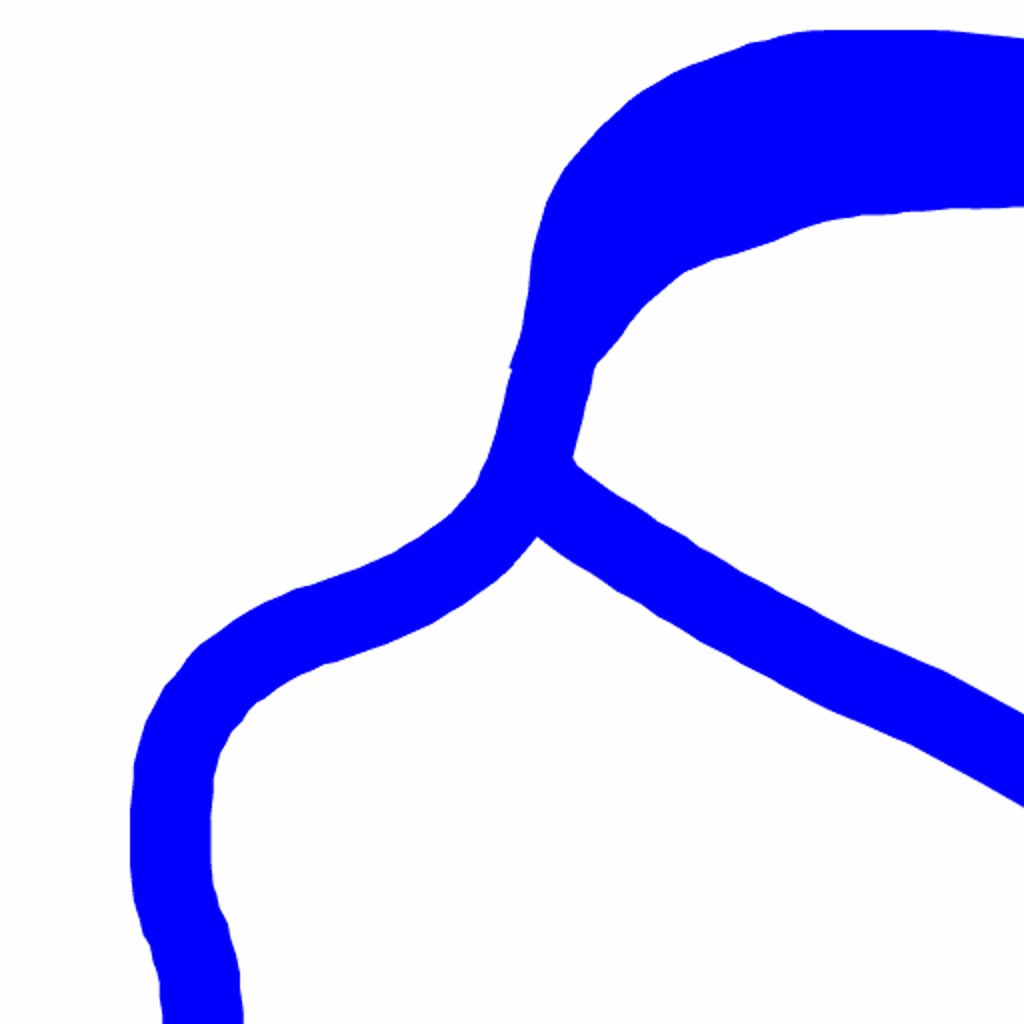}
    \includegraphics[width=0.31\linewidth]{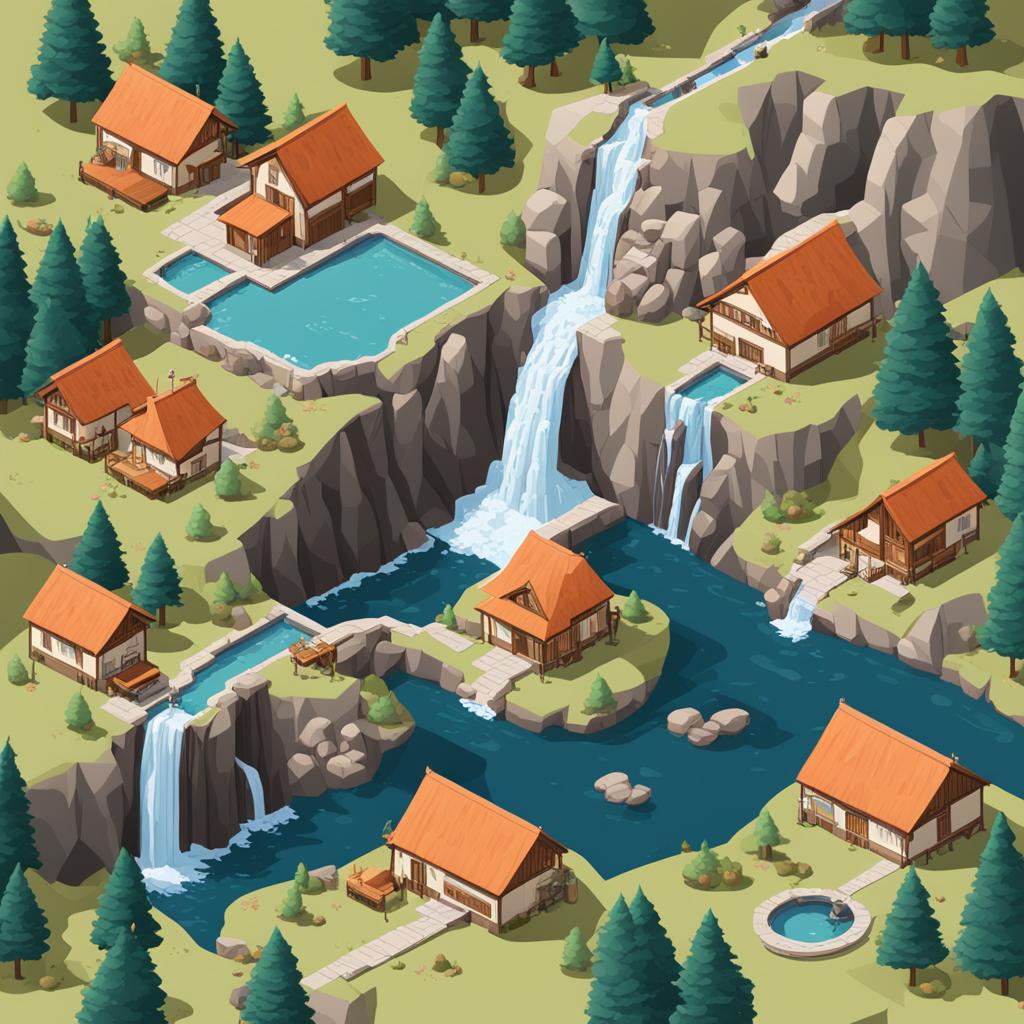} 
    \includegraphics[width=0.31\linewidth]{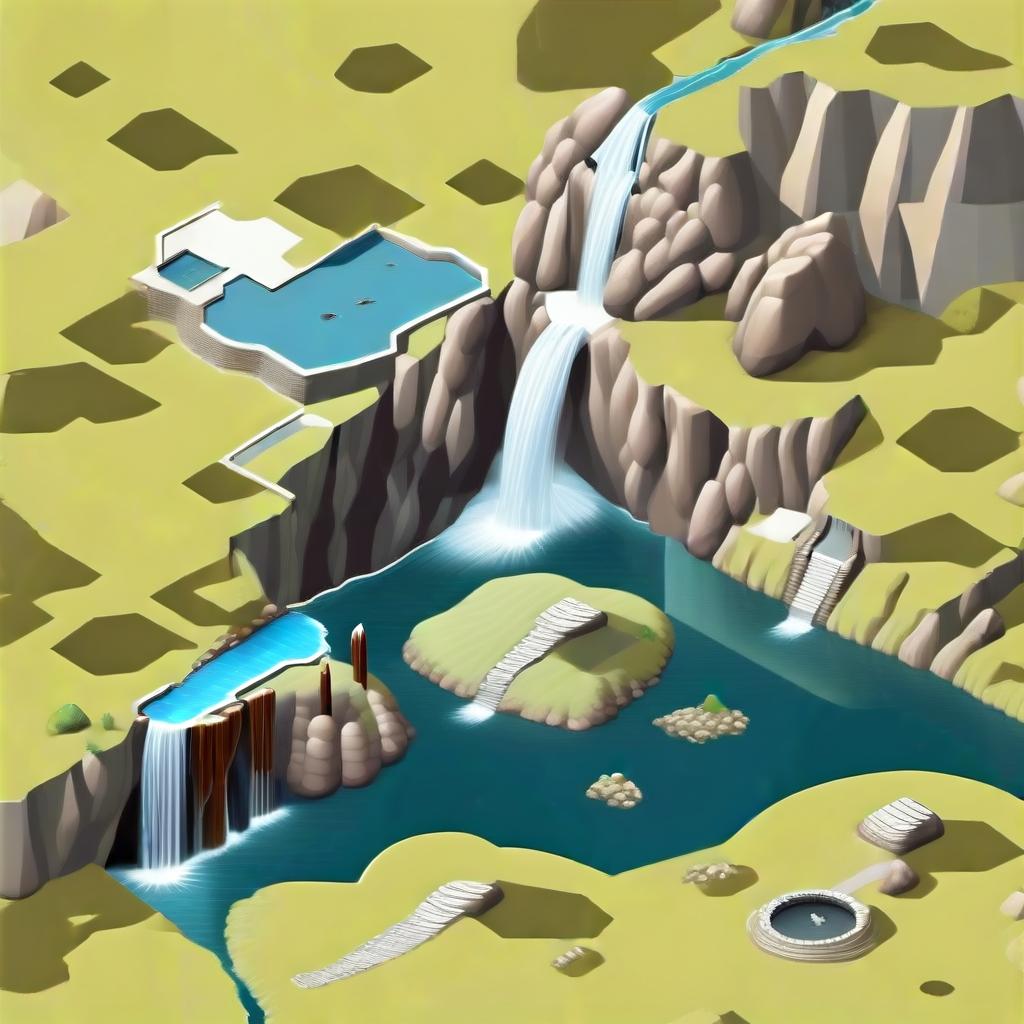} 
    \caption{The image depicts an isometric view of a mountainous landscape with a river, several houses, and a waterfall.}
    \label{fig:mountainous_waterfall}
    \end{subfigure}  

    \caption{More results of isometric image generation and basemap inpainting.}
    \label{fig:con_inp_sup2}
\end{figure}
\end{document}